\documentclass{article} 
\usepackage{iclr2027_conference,times} \iclrfinalcopy

\usepackage{amsmath,amssymb,amsthm,mathtools} \usepackage{booktabs}
\usepackage{multirow} \usepackage{graphicx} \usepackage{microtype} \usepackage{xcolor} \usepackage{placeins} \usepackage{needspace} \usepackage{float}
\extrafloats{200}

\usepackage{listings}
\definecolor{lstkw}{HTML}{8250DF}    
\definecolor{lstapi}{HTML}{0F766E}   
\definecolor{lststr}{HTML}{A6312C}   
\definecolor{lstcom}{HTML}{6B7280}   
\lstdefinestyle{refimpl}{%
language=Python, basicstyle=\ttfamily\footnotesize, keywordstyle=\bfseries\color{lstkw}, morekeywords={None,True,False}, emph={dyadic_radii,log_jitter,fd_amplitude,spectrum,aggregate,margin_spectra,margin_eval_fn,sample_dirs}, emphstyle=\bfseries\color{lstapi}, commentstyle=\itshape\color{lstcom}, stringstyle=\color{lststr}, backgroundcolor=\color{black!3}, columns=fullflexible, keepspaces=true, showstringspaces=false,
breaklines=true, breakatwhitespace=true, breakindent=2em, breakautoindent=true, frame=tb, framesep=5pt, captionpos=b, aboveskip=1.2\baselineskip, belowskip=0.8\baselineskip, } 

\usepackage{titletoc}

\titlecontents{lsection}[2.2em] {\addvspace{0.7em}} {\bfseries\contentslabel{2.2em}} {\bfseries\hspace*{-2.2em}} {\bfseries\hfill\contentspage} \titlecontents{lsubsection}[5.1em] {\addvspace{0.15em}} {\contentslabel{2.9em}} {\hspace*{-2.9em}} {\titlerule*[0.7pc]{.}\contentspage}

\usepackage[colorlinks=true,linkcolor=blue,citecolor=blue,urlcolor=blue]{hyperref} \usepackage{url}

\renewcommand{\eqref}[1]{Eq.~\ref{#1}}

\newsavebox{\tablebox} \newcommand{\fitwidth}[1]{%
\sbox{\tablebox}{#1}%
\ifdim\wd\tablebox>\linewidth \resizebox{\linewidth}{!}{\usebox{\tablebox}}%
\else \usebox{\tablebox}%
\fi}

\newtheorem{theorem}{Theorem} \newtheorem{proposition}{Proposition} \newtheorem{lemma}{Lemma} \newtheorem{corollary}{Corollary} \newtheorem{definition}{Definition} \newtheorem{remark}{Remark} \newtheorem{assumption}{Assumption}

\newcommand{\E}{\mathbb{E}} \newcommand{\Prob}{\mathbb{P}} \newcommand{\R}{\mathbb{R}}
\DeclarePairedDelimiter{\norm}{\lVert}{\rVert} \DeclarePairedDelimiter{\abs}{\lvert}{\rvert} \newcommand{\dskl}{\mathsf{d}_{\mathrm{SKL}}} \newcommand{\Lip}{\mathrm{Lip}}

\title{Scalar Smoothness Is One Point on a Besov Spectrum}

\author{Bum Jun Kim\thanks{Corresponding author}\\ The University of Tokyo\\ {\small\texttt{bumjun.kim@weblab.t.u-tokyo.ac.jp}} \And Anastasis Kratsios\\ McMaster University and Vector Institute\\ {\small\texttt{kratsioa@mcmaster.ca}}} \hypersetup{pdftitle={Scalar Smoothness Is One Point on a Besov Spectrum},pdfauthor={Bum Jun Kim and Anastasis Kratsios}} 
\begin{document} \maketitle
\lhead{}

\begin{abstract} Studies of neural network generalization and robustness often summarize smoothness with a single number, such as a Lipschitz constant or a Hessian curvature statistic. However, Lipschitz constants summarize global worst-case sensitivity and Hessian statistics summarize local curvature; neither captures how model responses vary with the size, direction, and location of perturbations encountered in real-world deployment. To measure smoothness under these perturbations, we introduce a finite-difference roughness spectrum that profiles model responses across these conditions. Our roughness spectrum generalizes scalar smoothness measures, bringing Lipschitz sensitivity and directional Hessian curvature into a common framework across perturbation scales. We connect this spectrum to classical Besov theory, which describes how a function's variation changes with scale. Our theoretical analysis also proves that a larger output shift can cause less task damage than a smaller one, and that local curvature alone can miss the blocks most vulnerable to finite perturbations. In image classification, Roughness Source Control (RSC) suppresses margin roughness along noise directions and improves noise robustness in a comparison of 80 checkpoints across four model--dataset combinations. In post-training quantization, we identify blocks that need higher precision by measuring task-loss increases from perturbations matched to the size and direction of each block's quantization error. Our mixed-precision allocation achieves lower perplexity than the Hessian-based HAWQ-V2 baseline under the same GPTQ weights and bit budgets in all 14 settings across seven held-out large language models. \end{abstract}

\section{Introduction} \label{sec:intro}

Smoothness underlies many accounts of neural-network generalization and robustness. Flat-minimum and sharpness-aware methods study the loss landscape \citep{hochreiter1997flat,keskar2017large,foret2021sharpness}; Lipschitz and spectral constraints target adversarial stability \citep{cisse2017parseval,tsuzuku2018lipschitz}; and Hessian statistics summarize local curvature \citep{ghorbani2019investigation,cohen2021gradient}. These lines of work summarize model smoothness with a single sensitivity or curvature score.

However, conventional scalar summaries do not show how sensitivity varies across the perturbations a model will encounter in deployment. Lipschitz bounds capture worst-case sensitivity, while Hessian statistics describe curvature under infinitesimal changes. Deployment perturbations act at finite scales and have source-specific structure: Image corruptions vary in severity and frequency content \citep{hendrycks2019benchmarking}, and quantization introduces block-specific weight errors set by the quantizer and bit width \citep{frantar2022gptq,dettmers2022llmint8}. Using Lipschitz constants and Hessian statistics leaves a gap between measured smoothness and model behavior under finite, structured deployment perturbations. To close this gap, smoothness assessment must account for the perturbation's scale, direction, and representation, meaning the quantity being perturbed and the perturbation site, such as input pixels, block weights, or intermediate features. The target task determines which changes a model should respond to: A classifier should tolerate sensor noise while remaining sensitive to changes that alter the class.

\Needspace{4\baselineskip} To measure smoothness under these deployment perturbations, we first choose a response that reflects how the perturbations affect the target task. Examples include the classification margin and prediction loss. We then profile this response across perturbation sizes and direction families. The resulting roughness spectrum shows how model smoothness varies with scale, direction, and representation. Selecting the point or scale band that matches a deployment perturbation gives a direct measure of smoothness under that scenario's actual conditions.

\begin{figure}[t!] \centering \includegraphics[width=\linewidth]{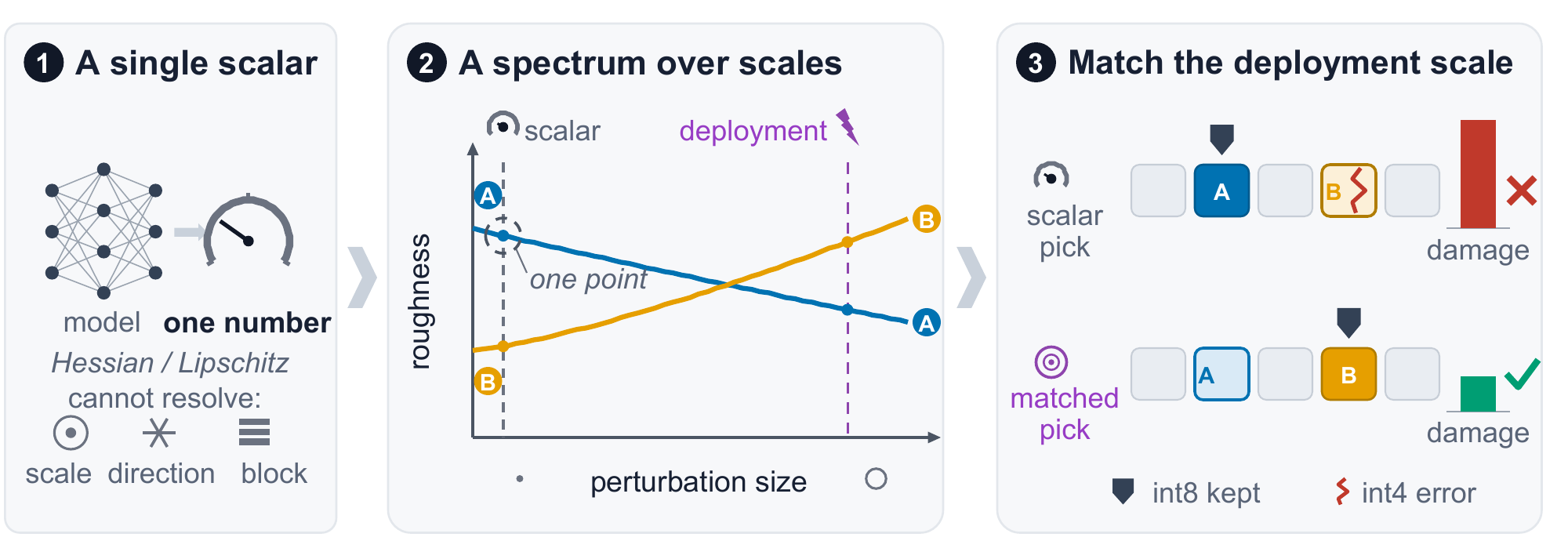} \caption{Overview of the roughness spectrum. A single scalar does not describe which scales, directions, or blocks carry sensitivity. A roughness spectrum measures smoothness at the perturbation scale a model will encounter. The block ranking at that scale can reverse the ordering suggested by a scalar summary. In quantization, we allocate the limited high-precision budget to blocks whose low-precision errors cause greater task damage.} \label{fig:roughness-overview} \end{figure}

We use this view in two applications. In vision, we measure and control margin roughness along nuisance directions over selected scale bands. In post-training quantization (PTQ), we measure each block's sensitivity through the increase in prediction loss caused by the block's actual quantization error, then allocate higher precision to the more sensitive blocks. Figure~\ref{fig:roughness-overview} illustrates how matching the measurement to the deployment perturbation guides a precision-allocation decision.

Our roughness spectrum is motivated by classical Besov theory, which describes how a function's variation changes with scale \citep{triebel1983theory,devore1993constructive}. We compute the spectrum with finite differences, comparing model responses at points a finite distance apart using forward passes alone. These differences measure variation at the chosen perturbation size. This construction connects familiar Lipschitz and Hessian diagnostics to finite perturbations, while the spectrum's multiscale population forms characterize Besov regularity.

Our contributions are: \begin{enumerate} \item We define finite-difference roughness spectra as a common smoothness diagnostic for perturbations to inputs, weights, and intermediate features. We connect the spectra's population forms to Besov regularity and show how scalar summaries can hide finite-scale variation (Sec.~\ref{sec:method}, App.~\ref{app:theory}). \item We prove theoretical limits of using output drift or local curvature to predict task damage: A larger output shift can cause less task damage than a smaller one, and local curvature alone can miss the blocks most vulnerable to finite perturbations (Sec.~\ref{sec:rsc}, App.~\ref{app:theory}). \item We design Roughness Source Control (RSC) for vision and a mixed-precision allocation rule for language models by matching spectrum measurements to the deployment perturbations in each application. RSC penalizes nuisance-direction margin roughness, while the allocation rule assigns higher precision to blocks whose quantization causes greater task damage (Sec.~\ref{sec:rsc}). \item Our evaluations demonstrate that RSC improves noise robustness in four vision pools and that damage-based allocation achieves lower perplexity than HAWQ-V2 in all 14 settings across seven held-out large language models (LLMs) under the same GPTQ weights and bit budgets (Sec.~\ref{sec:results}). \end{enumerate}

\section{Finite-Difference Roughness Spectra} \label{sec:method}

To construct the roughness spectrum, we first select the perturbation site: the input, block weights, or intermediate features. We then choose which response to track, such as the classification margin, output distribution, or task loss. Task loss measures prediction error against ground-truth targets. Probing this response across perturbation sizes and direction families gives the spectrum. To assess a particular deployment perturbation, we read the spectrum at the matching scale and direction family, or over the relevant scale band.

Let $z\in\mathcal{Z}$ denote the input, block weights, or intermediate features before perturbation, and let $g:\mathcal{Z}\to\mathcal{Y}$ map $z$ to the response being measured. For example, $g$ may map an input to the corresponding classification margin, a block's residual stream to the downstream model output, or parameters to the task loss. A first difference records how much the response changes over one step; a second difference compares that change over two adjacent steps. Both finite differences are evaluated at finite step sizes. For a perturbation $h\in\mathcal{Z}$, which changes $z$ to $z+h$, define the $m$-th-order forward finite difference \begin{equation} \begin{gathered} \Delta_h^m g(z) \coloneqq \sum_{a=0}^{m} (-1)^{m-a}\binom{m}{a} g(z + a h), \\ \Delta_h^1 g(z) = g(z+h)-g(z),\quad \Delta_h^2 g(z) = g(z+2h)-2g(z+h)+g(z). \end{gathered} \label{eq:fd} \end{equation}

We use these differences to see how rough the response is at small and large perturbation scales. The next definition gives one roughness value per scale and combines the values over a chosen scale band into one score.

\begin{definition}[Finite-difference roughness spectrum] \label{def:spectrum} Fix a data-induced base-point distribution $\mu$ on $\mathcal{Z}$, a family $\mathcal{H}_j(z)$ of unit-direction laws, and an independent log-uniform radius jitter $\xi\in[2^{-1/2},2^{1/2}]$.  On the dyadic grid $r_j \coloneqq r_0 2^{-j}$, $j\in J$, put $h_j\coloneqq r_j\xi u$ for $u\sim\mathcal H_j(z)$, and fix an integer $m\ge1$ and exponents $1\le p<\infty$, $0<q\le\infty$, $s>0$. The scale-$j$ amplitude, the scale-normalized spectrum, and the aggregate over $J$ are \begin{equation} \begin{gathered} A_j(g) \coloneqq (\E_{z\sim\mu}\E_{u\sim\mathcal{H}_j(z),\xi} \norm{\Delta^m_{h_j} g(z)}^p)^{1/p}, \qquad B_j(g) \coloneqq r_j^{-s} A_j(g), \\ R^{(m)}_{s,p,q}(g) \coloneqq (\textstyle\sum_{j\in J} B_j(g)^q)^{1/q}, \quad\text{read as a supremum when } q=\infty. \end{gathered} \label{eq:spectrum} \end{equation} \end{definition} The expectation in $A_j(g)$ averages the $p$-th power of the finite-difference magnitude over base points, directions, and radius jitter. Taking the $p$-th root gives $A_j(g)$, the response's roughness amplitude at radius $r_j$. A hat, as in $\widehat A_j(g)$, denotes the estimate computed from sampled base points and perturbations.

\paragraph{Matching the spectrum to the deployment perturbation.} The deployment perturbation determines which point or scale band to use. For a single scale and direction law, write \begin{equation} S_\star(g) \coloneqq B_{j^*}(g), \qquad (j^*,\mathcal{H}^*)\text{ fixed by the deployment perturbation}, \label{eq:readoff} \end{equation} with $\mathcal{H}^*$ the direction law entering $B_{j^*}$ through \eqref{eq:spectrum}. For a known deployment perturbation, we measure smoothness at the matching perturbation scale and along the matching directions. Vision RSC controls selected nuisance families over a finite band of scales (Sec.~\ref{sec:rsc}); PTQ uses the analogous task-damage measurement at the realized quantization error (Sec.~\ref{sec:exp-quant}).

The next remark connects these finite differences to Lipschitz sensitivity and directional Hessian curvature.

\begin{remark}[Classical diagnostics as endpoint readouts] \label{rem:scalar-endpoints} Fix a scalar function $g$ on an open domain $\mathcal Z\subseteq\R^d$, with $H_g(z)\coloneqq\nabla^2g(z)$. For $g\in C^2$, a unit direction $u$, and without radius jitter, \begin{equation} \frac{\Delta^1_{ru}g(z)}{r}\longrightarrow\nabla g(z)^\top u, \qquad \frac{\Delta^2_{ru}g(z)}{r^2}\longrightarrow u^\top H_g(z)u \quad(r\to0). \label{eq:scalar-limits} \end{equation} The global Lipschitz constant is a worst-case first-order readout, \begin{equation} \Lip(g)=\sup_{r>0}\sup_{\norm{u}_2=1}\sup_{z,z+ru\in\mathcal Z} \frac{\abs{\Delta^1_{ru}g(z)}}{r}. \label{eq:lipschitz-readoff} \end{equation} This worst-case readout extends the finite-$p$ average in Definition~\ref{def:spectrum} to a supremum over base points and directions; $q=\infty$ alone only maximizes over scales. The second-order $p=2$ endpoint gives root-mean-square directional Hessian curvature. For isotropic directions, the mean squared directional curvature at a fixed $z$ is $[(\operatorname{tr}H_g)^2+2\norm{H_g}_F^2]/[d(d+2)]$; the signed directional mean instead gives $\operatorname{tr}H_g/d$. These derivative readouts use $s=m$, while the Besov equivalence below applies to $0<s<m$. App.~\ref{app:scalar-endpoints} gives the population limits, including jitter and the regularity needed to exchange averaging with $r\to0$. \end{remark}

\paragraph{Canonical Besov connection.} The next theorem shows how the roughness spectrum in Definition~\ref{def:spectrum} connects to classical Besov smoothness.

\begin{assumption}[Sampling regularity] \label{asm:ambient} (A1) The base-point law $\mu$ has a density $\rho$ with respect to Lebesgue measure on the torus $\mathbb{T}^d\coloneqq(\R/\mathbb{Z})^d$ with $0<c_\mu\le\rho\le C_\mu<\infty$. (A2) The direction family is isotropic at every scale, $u\sim\mathrm{Unif}(S^{d-1})$, with the per-sample log-uniform radius jitter of Definition~\ref{def:spectrum}, so that $\norm{h}=r_j\xi$ with $\xi\in[2^{-1/2},2^{1/2}]$. (A3) $1\le p<\infty$, $0<q\le\infty$, $0<s<m$, and $r_0\le 1/(4m)$. \end{assumption}

\begin{theorem}[Ambient equivalence of the canonical population aggregate] \label{thm:besov-equiv} Let $R^{(m),\mathbb{N}}_{s,p,q}(g)$ be the aggregate of Definition~\ref{def:spectrum} over the full dyadic grid $J=\{0,1,2,\dots\}$. Under Assumption~\ref{asm:ambient}, there are constants $0<c\le C<\infty$ depending only on $(d,m,s,p,q,c_\mu,C_\mu,r_0)$, in particular not on $g$, such that for every $g\in L^p(\mathbb{T}^d)$, \begin{equation} c(\norm{g}_{L^p} + R^{(m),\mathbb{N}}_{s,p,q}(g)) \le \norm{g}_{B^s_{p,q}(\mathbb{T}^d)} \le C(\norm{g}_{L^p} + R^{(m),\mathbb{N}}_{s,p,q}(g)), \label{eq:ambient} \end{equation} where $\norm{g}_{B^s_{p,q}(\mathbb{T}^d)}$ denotes a standard periodic Besov quasi-norm, which is a norm when $q\ge1$. \end{theorem}

Under these conditions, adding $\norm{g}_{L^p}$ to the aggregate over the full dyadic grid gives the Besov (quasi-)norm up to fixed multiplicative factors independent of $g$. The proof is in App.~\ref{app:proof-besov}. Appendix~\ref{sec:ambient} develops finite-band sampling and truncation bounds; Apps.~\ref{app:directional-besov} and~\ref{app:uniform-besov} give mixed-direction, coordinate, and uniform extensions. The Lipschitz and Hessian endpoint readouts of Remark~\ref{rem:scalar-endpoints} lie outside the $s<m$ equivalence. Estimator validation against a closed-form Littlewood--Paley seminorm and empirical reliability checks appear in Apps.~\ref{sec:exp-besovequiv} and~\ref{sec:exp-reliability}, respectively.

\paragraph{Perturbation sites and measured responses.} We specify the perturbed quantity, measured response, and direction family for each probe below, with full details in App.~\ref{app:measurement}. Comparisons between models fix these choices and the probe settings $(\mu,\mathcal{H},J,s,p,q,m)$. We measure the raw amplitudes $A_j$ first and apply the scale weights $r_j^{-s}$ during analysis.

For vision, let $f_\theta(x)_k$ denote the pre-softmax score (logit) for class $k$ and $y$ the ground-truth label, held fixed under perturbation. The classification margin $m_\theta(x)\coloneqq f_\theta(x)_y-\max_{k\ne y}f_\theta(x)_k$ measures the true class's lead over the strongest competing class. The margin spectrum takes $g=m_\theta$ on test inputs, with directions drawn from an isotropic Gaussian, from high-frequency and low-frequency band-limited Gaussians cut out by radial fast Fourier transform (FFT) masks, and from label-preserving augmentation displacements, at $m=2$, $p=2$, and eight dyadic scales.

\begin{figure}[t!] \centering \includegraphics[width=\linewidth]{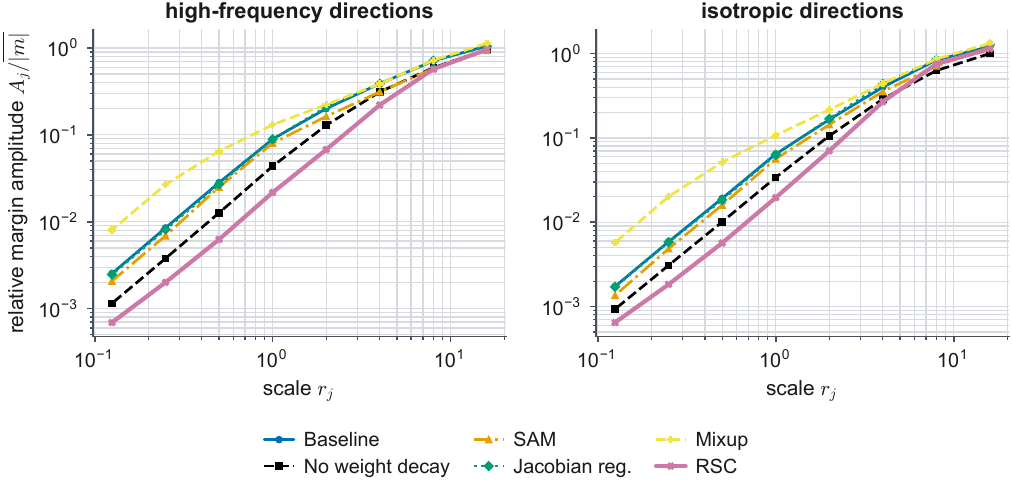} \caption{Relative second-difference amplitude of the classification margin, $A_j/\overline{\abs{m}}$, across probe radii for six CIFAR-10 recipes, measured along high-frequency (left) and isotropic (right) directions. Here $A_j$ is the root-mean-square second difference of the margin and $\overline{\abs{m}}=\E_x\abs{m_\theta(x)}$ is the margin's mean absolute clean value. Lower curves indicate smaller relative second differences. The pool spans only $1.2$ to $1.3\times$ at the coarsest radius but separates by $13.3\times$ and $11.0\times$, respectively, at $r_j=1/4$. RSC gives the lowest fine-scale amplitudes. SAM denotes sharpness-aware minimization.} \label{fig:vision-spectra} \end{figure}

Figure~\ref{fig:vision-spectra} plots the relative second-difference amplitudes of the CIFAR-10 models' classification margins at each probe radius, before aggregation across scales, for high-frequency and isotropic directions. The models have similar relative amplitudes at the coarsest radius, but the fine-scale amplitudes differ by more than a factor of ten. The model ranking by relative amplitude also changes with the selected radius.

For language models, the output-Kullback--Leibler (KL) spectrum measures how much the next-token probability distribution changes when an embedding or residual stream is perturbed. For a token sequence $x$, let $z$ denote the representation being perturbed and $h$ the perturbation. We measure $A^{\mathrm{KL}}_j \coloneqq ( \E \dskl(p_\theta(\cdot\mid z), p_\theta(\cdot\mid z+h))^{p} )^{1/p}$ using the symmetrized square-root KL $\dskl(P,Q) \coloneqq [(D_{\mathrm{KL}}(P\Vert Q) + D_{\mathrm{KL}}(Q\Vert P))/2]^{1/2}$, with radii relative to per-token representation norms. This output-KL curve compares two distributions at each scale ($m=1$) using a divergence and is reported separately from vector-valued finite-difference spectra.

Quantization-noise roughness is defined per layer block $\ell$ and bit width $b$. Let $Q_{\ell,b}$ be the per-channel symmetric quantizer applied to block $\ell$ only. The corresponding roughness is \begin{equation} B^{\mathrm{quant}}_{\ell,b} \coloneqq (\E_{x} \dskl(p_\theta(\cdot\mid x), p_{Q_{\ell,b}(\theta)}(\cdot\mid x))^{2})^{1/2}, \label{eq:bquant} \end{equation} a source-specific score at the realized weight error $\delta_{\ell,b}\coloneqq Q_{\ell,b}(\theta)-\theta$, supported on block $\ell$. The weight error's direction and magnitude depend jointly on the block weights, quantizer, and bit width. The measured response must also match the task. Drift amplitude measures how far the output distribution moves. For next-token prediction, the task loss is the negative log-probability assigned to the correct token: Lower probability means higher loss. To measure the task effect, we use the same forward passes to compute a damage score: the average increase in next-token negative log-likelihood (NLL) on calibration data when only block $\ell$ is quantized, \begin{equation} D^{\mathrm{quant}}_{\ell,b} \coloneqq \E_{x}[\mathrm{NLL}_{Q_{\ell,b}(\theta)}(x) - \mathrm{NLL}_{\theta}(x)], \label{eq:dquant} \end{equation} the first difference of the NLL functional along the same quantization-noise perturbation. A larger positive score means that quantizing the block hurts next-token prediction more. Both scores are computed from the same perturbation to a block's weights: Drift measures output-distribution movement, while damage measures the weight perturbation's effect on the task. For example, quantizing one block may mainly redistribute probability among incorrect tokens, giving large drift but little task damage. Quantizing another may shift the distribution less overall but lower the correct token's probability more, causing greater damage. Drift then ranks the first block as more sensitive, while damage ranks the second.

At a fixed block, quantizer, and bit width, each PTQ score measures the response to that block's actual quantization error. The residual-stream sweep in Sec.~\ref{sec:exp-quant} shows how block rankings change as each block's quantization-induced error is scaled. For damage-based allocation, we use only the task-damage measurement at the actual quantization scale. These objects share a scale--direction--representation indexing principle, with the response functional specified for each probe; only the margin and vector-valued constructions are literal instances of Definition~\ref{def:spectrum} (App.~\ref{app:measurement}). Each construction locates changes in model behavior caused by finite perturbations.

\section{RSC and Mixed-Precision Allocation} \label{sec:control} \label{sec:rsc}

RSC aims to keep classification stable under small, label-preserving input perturbations. RSC penalizes nonlinear changes in the margin, which measures the true class's lead over the strongest competing class, along nuisance directions at selected finite scales. Concretely, we probe the true-class margin $g=m_\theta$ along nuisance directions and penalize the difference between the margin's average slopes over two adjacent intervals. For a unit direction $u$ and $\phi(t)\coloneqq m_\theta(x+tu)$, we obtain the slope difference \begin{align*} \frac{\Delta^2_{ru}m_\theta(x)}{r} =\frac{\phi(2r)-\phi(r)}{r} -\frac{\phi(r)-\phi(0)}{r}. \end{align*} The second difference, with $m=2$, cancels affine variation, while division by the step length gives slope units, corresponding to $s=1$. Taking the second moment, with $p=2$, over training examples and nuisance directions and averaging uniformly over radii gives a squared $q=2$ roughness aggregate, normalized by the number of scales. Specializing the finite-difference construction in Definition~\ref{def:spectrum} to vision training gives the RSC objective \begin{equation} \mathcal{L}_{\mathrm{RSC}} \coloneqq \E_{j,x}\E_{u\sim\mathcal{H}^{\mathrm{RSC}}_j} (\Delta^2_{r_j u}m_\theta(x)/r_j)^2, \quad \mathcal{L}_{\mathrm{train}} =\mathcal{L}_{\mathrm{task}}+\lambda_{\mathrm{RSC}}\mathcal{L}_{\mathrm{RSC}}. \label{eq:rsc} \end{equation} Each step independently samples a radius $r_j\in\{4,2,1,0.5\}$ uniformly and a direction family uniformly from isotropic and high-frequency, with unit-norm directions (App.~\ref{app:measurement}).

For evaluation, we use $s=1.5$ and divide the roughness aggregate by the mean absolute margin on clean inputs (App.~\ref{app:measurement}). We measure how adding the RSC penalty to the training loss changes this relative roughness and classification accuracy under noise (Sec.~\ref{sec:exp-vision}). Appendix~\ref{app:theory-dyadic} develops the link between margin roughness and prediction stability.

For PTQ, we use a separate finite-scale procedure to choose which blocks receive higher precision. We measure the per-block functional scores (Eqs.~\ref{eq:bquant} and~\ref{eq:dquant}) on a calibration set. With a budget of $K$ high-precision blocks, the additive rule assigns higher precision to the $K$ blocks with the largest individual scores. The greedy rule starts with all blocks at low precision and upgrades one block at a time. At each step, the greedy rule evaluates every possible single-block upgrade on the current mixed-precision model and chooses the candidate with the lowest score for the resulting model, until $K$ blocks have been upgraded (Sec.~\ref{sec:exp-quant}, App.~\ref{app:alloc-extra}). Concretely, the implemented drift-greedy rule re-measures mean symmetric KL, the square of the drift amplitude in \eqref{eq:bquant}, while damage-greedy re-measures the signed NLL increase in \eqref{eq:dquant}. Appendix~\ref{app:theory-alignment} explains the relation between these greedy objectives and the reported block scores.

Our theoretical analysis shows why we measure task damage at the quantization scale: Output drift can misrank block damage, and local curvature can miss large finite-scale damage. Let $q_x$ be the target distribution, and let $P_x$ and $Q_{\ell,x}$ be the full-precision and single-block-quantized outputs. For forward-KL drift $G_\ell\coloneqq\E_x D_{\mathrm{KL}}(P_x\Vert Q_{\ell,x})$, task damage satisfies \begin{align*} D_\ell-G_\ell &=\E_x\sum_y(q_x(y)-P_x(y)) \log\frac{P_x(y)}{Q_{\ell,x}(y)}. \end{align*} Output drift alone can misrank task damage because the drift score does not use the task's target distribution (Theorem~\ref{thm:drift-damage}). Averaging forward and reverse KL does not prevent incorrect block rankings: A block with larger symmetric KL can still cause less task damage (Proposition~\ref{prop:alignment-inversion}). Existing methods such as HAWQ-V2 combine local Hessian information with quantization-error magnitude \citep{dong2020hawqv2}. We construct additive block-damage functions that have the same value, slope, and curvature at the full-precision model but different damage at the quantization scale. For these functions, any deterministic rule that allocates precision using only that local information can incur arbitrarily more damage than the optimal allocation (Theorem~\ref{thm:alloc-regret}).

\section{Experiments} \label{sec:results}

\paragraph{Setup.} We first measure roughness spectra and block-sensitivity scores on clean or calibration data, then compare model performance under the same perturbation protocols. The vision suite contains 80 checkpoints in four pools: ResNet-18 on CIFAR-10 and CIFAR-100, WRN-28-10 on CIFAR-10, and ViT-S/4 on CIFAR-10. We evaluate classification accuracy on CIFAR-C, after round-to-nearest (RTN) weight PTQ, and under Gaussian input noise, reporting clean accuracy alongside accuracy under each perturbation. Accuracy under CIFAR-C noise is averaged over the benchmark's noise family. The primary robustness test uses additive Gaussian input noise at $\sigma=0.05$. The primary diagnostic is high-frequency margin roughness $\tilde R^{\mathrm{hf}}_{1.5}$.

The language suite contains Pythia-160M to 6.9B \citep{biderman2023pythia}, a preregistered Pythia-12B check, seven held-out LLMs, Qwen2.5 \citep{yang2024qwen25}, and Mamba \citep{gu2023mamba}. The default calibration budget is 64 sequences; experiment-specific budgets are reported with the corresponding protocols. Scores are evaluated against the per-block $\Delta$NLL caused by int4 quantization. For allocation, we compare against an AWQ-inspired rule that ranks blocks by activation salience $\times$ weight mean squared error (MSE) \citep{lin2024awq}, and against HAWQ-V2 \citep{dong2019hawq,dong2020hawqv2}. All allocation rules use the same quantized weights and bit budgets. Full recipes, seeds, baselines, and protocols are in Apps.~\ref{app:setup-details} and~\ref{app:measurement}.

\subsection{Vision pools: RSC improves noise robustness} \label{sec:exp-vision}

\paragraph{RSC improves accuracy under input noise.} Across the three convolutional pools, RSC substantially improves classification accuracy under noise. RSC matches the baseline's clean accuracy on CIFAR-10 and WRN-28-10, with a $0.6$-point clean-accuracy cost on CIFAR-100. On CIFAR-10, accuracy under CIFAR-C noise is $71.2\%$, against $63.4\%$ for SAM and $62.3\%$ for the Jacobian penalty, while Gaussian-noise accuracy is $82.5\%$ and clean accuracy is $95.30\pm0.07$ against the baseline's $95.31\pm0.03$. RSC consistently improves on the baseline in the CIFAR-100 and WRN-28-10 pools, by $8.6$ to $10.2$ points under CIFAR-C noise and $17.5$ to $17.9$ points under Gaussian noise. App.~\ref{app:vision-pools} gives the complete tables and auxiliary arms; Figure~\ref{fig:vision-int} in the same appendix summarizes the two ResNet pools. Matched-wall-clock and BatchNorm-recalibration controls are reported in App.~\ref{app:vision-pools}.

\paragraph{RSC reduces the reported roughness in convolutional models.} The margin spectrum measures responses to input perturbations, while the Hessian baseline measures training-loss curvature with respect to model parameters (App.~\ref{app:measurement}). RSC reduces high-frequency margin roughness by about $3\times$: $\tilde R^{\mathrm{hf}}_{1.5}$ falls from $0.18$ to $0.06$ on CIFAR-10 and from $0.48$ to $0.13$ on CIFAR-100, while the uniform Jacobian penalty changes neither the diagnostic nor the failures. However, the dominant Hessian eigenvalue rises on CIFAR-100 from $12.39\pm2.85$ to $23.61\pm5.17$ while noise robustness improves, which shows that a larger Hessian eigenvalue does not imply worse noise robustness in this comparison. On ViT-S/4, the same intervention increases accuracy by $6.5$ points under CIFAR-C noise, $8.4$ under Gaussian noise, and $1.1$ on clean data without retuning. RSC's accuracy gains extend to ViT-S/4 even though the reported roughness remains at $0.06$ (Table~\ref{tab:vision-vit}).

\suppressfloats[t] 
\subsection{Roughness-guided quantization of pretrained LLMs} \label{sec:exp-quant}

Mixed-precision quantization assigns a limited high-precision budget to model blocks. HAWQ-V2 selects blocks using local Hessian sensitivity weighted by quantization error \citep{dong2020hawqv2}. Our damage-based rule quantizes one block at a time, measures the increase in next-token loss on calibration data, and assigns higher precision to the blocks with the largest increases in next-token loss. This damage measurement captures each block's sensitivity at the size, direction, and location of the block's actual quantization error. We then compare the resulting allocations using the same quantized weights and bit budget.

We evaluate drift $B^{\mathrm{quant}}_{\ell,4}$ and task-aligned damage $D^{\mathrm{quant}}_{\ell,4}$ on Pythia-160M to 6.9B. Table~\ref{tab:pythia} (App.~\ref{app:alloc-extra}) reports the complete allocation comparison at int4 and int8. Ground-truth block damage is measured on 131,072 paired evaluation tokens per block. In all ten Pythia model--budget settings, one of our allocation rules achieves the lowest or tied-lowest perplexity. The best rule varies by setting: additive drift, additive damage, or interaction-aware greedy. Committing to one rule, additive damage in every setting, removes this choice: Paired bootstrap intervals for its perplexity gap to HAWQ-V2, computed over the same 128 evaluation windows, exclude zero in all ten settings, with damage ahead in nine and behind in one, Pythia-6.9B at $K=L/4$ (Table~\ref{tab:allocci}). App.~\ref{app:alloc-extra} also reports additivity audits.

\Needspace{9\baselineskip} \paragraph{Probing the effect of perturbation scale.} To test whether smaller perturbations identify the same vulnerable blocks as the full quantization error, we vary each block's residual-stream error magnitude while keeping the error direction and injection site fixed. For each block $i$, we capture the residual-stream output error $e_i$ induced by quantizing that block to int4 and trace the downstream teacher-forced damage \begin{align*} D_i(r) &\coloneqq \E_t[\mathrm{NLL}_t(\mathrm{res}_i + re_i) - \mathrm{NLL}_t(\mathrm{res}_i)] \end{align*} over dyadic multipliers of the fixed error $e_i$, from $r=1$ down to $r=2^{-4}$. At $r=1$, we apply each block's full realized error, exactly reproducing the output and task damage of single-block int4 quantization on the same calibration windows. Damage on the independent 131,072-token evaluation set can differ from these calibration values.

The damage profile $D_i(r)$ is a signed first-difference curve that shows how each block's effect on next-token loss changes as the quantization error grows. Let $G_i(r)\coloneqq\E_t[\mathrm{NLL}_t(\mathrm{res}_i+re_i)]$. When $G_i$ is $C^2$ near zero, $D_i(r)/r\to G_i'(0)$ and the Hessian enters the quadratic term (App.~\ref{sec:exp-spectrum}). Small probes describe the local loss response; $r=1$ measures the effect of the full quantization error. The Hessian baselines measure loss curvature with respect to weights, whereas $D_i(r)$ measures loss changes from residual-stream perturbations. The comparison differs in perturbation site as well as scale.

\paragraph{Calibration damage predicts held-out loss increases.} The calibration damage score has the highest rank correlation with independently measured evaluation-set $\Delta$NLL on every tested Pythia model: $0.98$, $0.95$, $0.99$, $0.92$, and $0.99$ from 160M to 6.9B (Table~\ref{tab:pythia-rc}, App.~\ref{app:alloc-extra}). The originally reported single-draw drift score tracks damage through 1.4B but reverses to $-0.09$ at 2.8B and $-0.45$ at 6.9B. Four fresh calibration draws show the same pattern: The pooled estimates are $-0.06$ at 2.8B and $-0.45$ at 6.9B, with a $95\%$ confidence interval (CI) of $[-0.46,-0.43]$ at 6.9B, while inter-draw rank agreement remains $0.98$ to $0.99$. The reversal persists across independent calibration draws.

\paragraph{The quantization-scale probe best predicts per-block int4 loss increases.} The rank correlation between $D_i(r)$ and true per-block int4 damage is highest at the full quantization error ($r=1$) and decreases as the probe radius shrinks to $r=2^{-4}$. Averaged over seven models, the correlation falls from $0.92$ at $r=1$ to $0.53$ at $r=2^{-4}$, while the parameter-space loss-Hessian trace and HAWQ-V2 baselines sit at $0.59$ and $0.70$ (Table~\ref{tab:spectrum-corr}). On three held-out modern deployment LLMs, the quantization-scale correlation averages $0.87$, compared with $0.15$ for the loss-Hessian trace and $0.22$ for HAWQ-V2. The finite-scale measurement identifies the blocks most vulnerable to int4 quantization even on models where Hessian-based scores correlate only weakly with the true block damage ranking. App.~\ref{sec:exp-spectrum} gives the numerical-precision and sampling details, along with a held-out allocation example in Figure~\ref{fig:qual-scale-crossing}.

\begin{table}[t!] \centering \caption{Spearman correlation between fp32 residual-stream damage $D_i(r)$ and independently measured per-block int4 $\Delta$NLL across dyadic probe scales. The next-token loss increase measured at the full quantization error ($r=1$) predicts block damage better than smaller probes or the Hessian baselines on every model shown. The local-curvature baselines are the loss-Hessian trace and HAWQ-V2. Here $\nu$ is the median of $\log_2(D_i(1)/D_i(1/2))$ over blocks with positive damage at both scales. The lower group tests transfer to three held-out LLMs without retuning.} \label{tab:spectrum-corr} \small \begin{tabular}{l c ccccc c cc} \toprule & & \multicolumn{5}{c}{$\rho$ at dyadic probe scale $r$} & & \multicolumn{2}{c}{local-curvature baselines} \\ \cmidrule(lr){3-7}\cmidrule(lr){9-10} Model & $L$ & $1$ & $1/2$ & $1/4$ & $1/8$ & $1/16$ & $\nu$ & $\rho_{\mathrm{htr}}$ & $\rho_{\mathrm{HAWQ}}$ \\ \midrule Pythia-160M & 12 & 0.94 & 0.90 & 0.87 & 0.76 & 0.58 & 2.1 & 0.43 & 0.41 \\ Pythia-410M & 24 & 0.89 & 0.88 & 0.87 & 0.85 & 0.73 & 2.0 & 0.67 & 0.69 \\ Pythia-1.4B & 24 & 0.98 & 0.94 & 0.83 & 0.56 & 0.40 & 2.0 & 0.60 & 0.76 \\ Pythia-2.8B & 32 & 0.87 & 0.77 & 0.65 & 0.53 & 0.50 & 2.0 & 0.37 & 0.64 \\ Pythia-6.9B & 32 & 0.97 & 0.96 & 0.88 & 0.80 & 0.73 & 1.9 & 0.46 & 0.83 \\ Qwen2.5-1.5B & 28 & 0.94 & 0.92 & 0.87 & 0.74 & 0.47 & 1.9 & 0.83 & 0.84 \\ Qwen2.5-7B & 28 & 0.83 & 0.75 & 0.49 & 0.34 & 0.28 & 2.2 & 0.76 & 0.72 \\ \midrule \multicolumn{10}{l}{Held-out LLMs}\\ Phi-3.5-mini-instruct & 32 & 0.86 & 0.78 & 0.58 & 0.37 & 0.26 & 2.0 & 0.17 & 0.25 \\ Granite-3.1-8B-Instruct & 40 & 0.87 & 0.82 & 0.78 & 0.74 & 0.71 & 1.6 & 0.15 & 0.22 \\ Hermes-3-Llama-3.1-8B & 32 & 0.88 & 0.76 & 0.61 & 0.58 & 0.58 & 1.9 & 0.15 & 0.20 \\ \bottomrule \end{tabular} \end{table}

\paragraph{Preregistered evaluation on Pythia-12B.} We introduced the calibration damage score after observing the drift reversal. Before producing the Pythia-12B quantization and diagnostic outputs, we preregistered five predictions about drift, damage, prediction flips, and allocation quality, together with numerical thresholds, then ran the same pipeline. All five predictions met the preregistered thresholds (Table~\ref{tab:pythia-held}): Drift's rank correlation with held-out block damage was $-0.221$, below the preset $+0.10$ ceiling; the damage score's correlation was $+0.973$, above the $+0.85$ floor; and damage-based allocation achieved the lowest perplexity among the preregistered rules at both audited budgets. App.~\ref{app:prereg} gives the protocol and an additional HAWQ-V2 comparison. Depth-resolved and calibration-budget checks are reported in Apps.~\ref{app:depth} and~\ref{app:hawqbudget}.

\subsection{Held-out generalization} \label{sec:exp-modllm} We test generalization to seven instruction-tuned LLMs from six vendors using the same GPTQ allocation procedure, without model-specific retuning. Table~\ref{tab:heldout-gptq} compares perplexity after damage-based and HAWQ-V2 allocation under the same GPTQ weights and bit budgets. Damage-based allocation yields lower perplexity in all 14 model--budget settings and also improves over uniform GPTQ-int4. Under RTN, one fixed drift-greedy rule yields lower perplexity than HAWQ-V2 in 27 of the 28 held-out model--budget settings from $K=L/2$ to $L/16$, and the advantage concentrates where the int8 budget is tight (Table~\ref{tab:tight}): At $K=2$ on Mistral-NeMo, HAWQ-V2 collapses to $16.6$ perplexity, barely improving on uniform int4 at $17.0$, while the greedy rule holds at $8.3$. App.~\ref{app:modllm-extra} reports all baselines, per-block rankings, pooled oracle-set agreement, and a Hermes-3 case analysis.

\begin{table}[t!] \centering \caption{Held-out GPTQ mixed-precision allocation. WikiText-103 perplexity is lower-is-better; $K$ of $L$ blocks use int8 and the rest int4. Damage and HAWQ-V2 use the same GPTQ weights and bit budget; bold marks the lower value in each head-to-head. Choosing blocks by calibration damage gives lower perplexity than HAWQ-V2 on all seven models at both budgets.} \label{tab:heldout-gptq} \small \begin{tabular}{l cc cc} \toprule & \multicolumn{2}{c}{$K=L/4$} & \multicolumn{2}{c}{$K=L/2$} \\ \cmidrule(lr){2-3}\cmidrule(lr){4-5} Model & HAWQ-V2 & damage & HAWQ-V2 & damage \\ \midrule Phi-3.5-mini (3.8B) & 7.70 & \textbf{7.53} & 7.45 & \textbf{7.15} \\ OLMo-2-7B & 12.01 & \textbf{11.40} & 11.02 & \textbf{10.49} \\ Granite-3.1-8B & 6.22 & \textbf{6.11} & 6.04 & \textbf{5.95} \\ Hermes-3 & 9.78 & \textbf{9.50} & 9.26 & \textbf{8.99} \\ Mistral-NeMo (12B) & 7.31 & \textbf{7.14} & 7.08 & \textbf{6.86} \\ Qwen2.5-14B & 7.13 & \textbf{7.10} & 6.93 & \textbf{6.87} \\ Qwen2.5-32B & 6.78 & \textbf{6.59} & 6.56 & \textbf{6.38} \\ \bottomrule \end{tabular} \end{table}

\paragraph{Additional transfer settings.} \label{sec:exp-generality} App.~\ref{app:contests} extends the comparison to Qwen2.5 and Mamba and repeats the allocation evaluation under GPTQ, AWQ, and a controlled SpQR-style format. The appendix also reports an ImageNet-scale ResNet-50 transfer and five zero-shot language-model benchmarks (Apps.~\ref{sec:exp-inq} and~\ref{sec:exp-downstream}). These results test transfer across architectures, quantizers, and evaluation tasks.

\section{Discussion and Conclusion} \label{sec:discussion}

Our finite-difference spectrum measures model smoothness across perturbation scales, directions, and locations. The spectrum's population forms connect to Besov regularity. The construction includes classical scalar diagnostics as special cases: Worst-case slopes give Lipschitz sensitivity, and small-step limits of second-order difference quotients give directional Hessian curvature. Selecting the point on this spectrum that matches a deployment perturbation's scale, direction, and location gives a smoothness measure for that scenario.

Our experiments demonstrate two applications. In image classification models, RSC regularizes margin roughness along nuisance directions and improves noise robustness across the four model--dataset pools. In PTQ of LLMs, we treat quantization error as a finite perturbation to block weights. We allocate higher precision to blocks whose quantization causes larger increases in next-token prediction loss. This allocation achieves lower perplexity than HAWQ-V2 on all seven held-out LLMs under the same GPTQ weights and precision budgets.

The drift--damage analysis shows why the choice between measuring output drift and prediction loss matters: A larger change in the output distribution can cause a smaller increase in prediction loss. Our practical recommendation is to identify the expected deployment perturbation, choose a response that captures the perturbation's effect on the task, and evaluate the spectrum at the matching scale or scale band and along the matching directions.

\phantomsection \addcontentsline{toc}{section}{AI use statement} \section*{AI use statement}

Generative AI tools were used as assistive tools for writing and editing, mathematical proofreading and consistency checking, and code development and debugging. Any AI-assisted outputs incorporated into this work were reviewed and validated by the authors as appropriate. The authors take full responsibility for the final content of this work, including any text, claims, or artifacts produced with the assistance of generative AI.

\phantomsection \addcontentsline{toc}{section}{Reproducibility statement} \section*{Reproducibility statement}

Apps.~\ref{app:setup-details} through~\ref{app:code} provide complete recipes, hyperparameters, seeds, public models and data, baselines and budgets, infrastructure, protocols, and code; Apps.~\ref{sec:exp-besovequiv} and~\ref{sec:exp-reliability} validate the estimator. Assumptions and theoretical statements appear in Sec.~\ref{sec:method} and App.~\ref{app:theory}, with proofs in App.~\ref{app:proofs}; App.~\ref{app:prereg} records the preregistration.

\phantomsection \addcontentsline{toc}{section}{Acknowledgements} \section*{Acknowledgements} We thank Taiji Suzuki for insightful discussions.

\bibliographystyle{iclr2027_conference} \bibliography{iclr2027_conference}

\appendix

\clearpage

\phantomsection \addcontentsline{toc}{section}{Appendix Table of Contents} \section*{Appendix Table of Contents} \startcontents[apx] \printcontents[apx]{l}{1}{\setcounter{tocdepth}{2}}

\clearpage

\section{Notation} \label{app:notation}

\begin{table}[t!] \centering \caption{List of notation. The symbols used throughout the paper, grouped by where each is introduced: the roughness spectrum of Definition~\ref{def:spectrum}, the vision instantiation of the spectrum (Sec.~\ref{sec:rsc}), the language-model drift and damage scores (Eqs.~\ref{eq:bquant} and~\ref{eq:dquant}), and the precision-allocation theory (App.~\ref{app:theory}).} \label{tab:notation} \small \begin{tabular}{ll} \toprule Symbol & Meaning \\ \midrule \multicolumn{2}{l}{Roughness spectrum} \\ $g$ & Probed function: a margin, an output law, or a loss \\ $z$, $\mu$ & Base point in $\mathcal{Z}$ and data-induced base-point law \\ $m$, $\Delta^m_h g(z)$ & Order and $m$-th forward difference of $g$ at step $h$ \\ $\mathcal{H}_j(z)$, $u$ & Unit-direction law at scale $j$ and a draw from that law \\ $r_j \coloneqq r_0 2^{-j}$, $J$ & Dyadic probe radius and the finite grid of scales, $j\in J$ \\ $\xi$, $h_j$ & Radius jitter and the probe step $h_j \coloneqq r_j\xi u$ \\ $A_j(g)$ & Scale-$j$ amplitude: $p$-th moment of $\Delta^m_{h_j}g$ \\ $U_j^{(m)}(g;\Omega)$ & Uniform difference amplitude on the domain $\Omega$ \\ $B_j(g)$ & Scale-normalized spectrum $r_j^{-s}A_j(g)$ \\ $R^{(m)}_{s,p,q}(g)$ & Aggregate of $B_j$ over the scale grid $J$ \\ $S_\star(g)$, $j^*$, $\mathcal{H}^*$ & Single-scale read-off \eqref{eq:readoff}, read-off scale, and direction family \\ $s$, $p$, $q$ & Smoothness exponent, moment order, aggregation order \\ $B^s_{p,q}$, $\dot B^s_{p,q}$ & Besov (quasi-)norm and homogeneous Besov seminorm \\ \midrule \multicolumn{2}{l}{Vision instantiation} \\ $\theta$, $f_\theta$ & Model parameters and network logits \\ $m_\theta(x)$ & Margin: true-class logit minus the best rival \\ $\hat y(x)$, $\gamma$ & Predicted label and the margin threshold \\ $\tilde R^{\mathrm{hf}}_{1.5}$ & Margin-normalized high-frequency roughness at $s=1.5$ \\ $\widetilde\Lip$ & Margin-normalized empirical (sampled) Lipschitz constant \\ $\lambda_{\mathrm{dom}}$ & Signed dominant-magnitude Hessian eigenvalue \\ $\mathcal{L}_{\mathrm{RSC}}$, $\lambda_{\mathrm{RSC}}$ & RSC margin-roughness penalty and penalty weight \\ $\mathcal{L}_{\mathrm{cross}}$, $\lambda_{\mathrm{cross}}$ & Auxiliary crossing penalty and weight (App.~\ref{app:crossing-penalty}) \\ \midrule \multicolumn{2}{l}{Language-model drift and damage scores} \\ $\mathrm{NLL}_\theta(x)$ & Next-token NLL at $\theta$ \\ $P_x$, $Q_{\ell,x}$ & Output law at $x$: full precision, block $\ell$ quantized \\ $q_x$ & Task target law; a point mass on the observed token \\ $\dskl(P,Q)$ & Symmetrized square-root KL divergence \\ $\ell$, $L$, $b$ & Block index, number of blocks, bit width \\ $Q_{\ell,b}$ & Per-channel symmetric quantizer on block $\ell$ \\ $B^{\mathrm{quant}}_{\ell,b}$ & Label-free drift score of block $\ell$ at $b$ bits \\ $D^{\mathrm{quant}}_{\ell,b}$, $D_\ell$ & Task damage of block $\ell$ alone: the block's $\Delta$NLL \\ $G_\ell$ & Label-free forward-KL self-damage of block $\ell$ \\ $C_{\mathrm{tgt}}$, $V_\ell$ & Target mismatch and log-ratio spread in the bound \\ \midrule \multicolumn{2}{l}{Precision-allocation theory} \\ $K$, $k\coloneqq L-K$ & Blocks at high precision and the rest at low precision \\ $S$, $\widehat S$, $S^\star$ & Low-precision set: generic, selected, oracle \\ $\delta_\ell$, $\delta_S$ & Quantization weight residual of block $\ell$, of set $S$ \\ $\Delta(\delta)$, $\mathcal{D}(S)$ & Damage of a perturbation, of a configuration \\ $\mathcal{C}_{\mathrm{SKL}}(S)$ & Configuration-level drift criterion (drift-greedy) \\ $C_{\mathrm{out}}$ & Mean symmetric KL at the final output in the depth probe \\ $I_{\ell\ell'}$, $\Lambda_k$ & Pairwise and worst-case mixed-curvature envelopes \\ $\varepsilon$ & Calibration error of the estimates $\widehat D_\ell$ \\ $H$, $\mathrm{HAWQ}_\ell$ & Loss Hessian and the HAWQ-V2 sensitivity score \\ $\phi_\ell(t)$, $\mathcal{P}_{L,c,M}$ & Block loss profile; the jet-matched instance class \\ $M$ & Per-block damage scale bounding that class \\ $\mathrm{Reg}(S)$ & Allocation regret against the oracle set \\ $D_i(r)$, $e_i$ & Residual-stream damage curve; block $i$'s int4 error \\ $\rho_s$ & Spearman rank correlation between two scores \\ \bottomrule \end{tabular} \end{table}

Table~\ref{tab:notation} collects the mathematical symbols used in the paper.

\section{Additional scope and evidence details} \label{app:background} \label{app:contributions} \label{app:caveats} \label{app:evidence-grades} \label{app:findings-summary}

\paragraph{Scope.} We study CIFAR-scale training, ImageNet-scale PTQ, selected language-model families, and the RTN, GPTQ, AWQ, and controlled SpQR-style formats. Results for other datasets, architectures, training scales, and quantizers remain to be established.

\paragraph{Interpretation and evidence.} The canonical finite-$p$ aggregate, the mixed-direction aggregate, and the uniform aggregate have ambient Besov characterizations (Theorem~\ref{thm:besov-equiv}, Corollary~\ref{cor:mixed-besov}, Proposition~\ref{prop:uniform-besov}). Data-dependent margin spectra are moments of the same differences and are bounded by the uniform amplitudes. Output divergences and signed task-loss differences provide the corresponding task-specific diagnostics. The measurement evidence consists of closed-form validation, estimator reliability, and movement of the reported margin roughness under RSC in the convolutional pools. The allocation evidence is more direct: matched-format, matched-budget comparisons against HAWQ-V2, including a preregistered Pythia-12B evaluation and seven held-out LLMs. The allocation conclusions rest on downstream head-to-heads rather than on treating every deployed probe as an ambient norm estimator.

\paragraph{Empirical findings.} In vision, RSC reduces high-frequency margin roughness in convolutional models and improves noise robustness across four architecture--dataset pools, with clean costs explicitly reported. In language models, task alignment matters: Our observation on Pythia-6.9B is that output drift correlates negatively with realized int4 block damage at $\rho_s=-0.45$, whereas damage computed from the same forward passes correlates with realized damage at $0.99$. All five preregistered predictions also held on Pythia-12B. After fixing the allocation rule, task-aligned damage outperforms HAWQ-V2 under the same GPTQ weights and bit budgets in all 14 settings across seven held-out LLMs from six vendors. In a separate seven-model sweep along each block's fixed residual-error direction, mean Spearman correlation with int4 damage rises from $0.53$ at the smallest measured radius to $0.92$ at the quantization scale. The separate parameter-space Hessian-trace baseline attains $0.59$.

\paragraph{Additional limitations.} Label-free drift can separate from task damage, and additive scores can weaken when block interactions become important; the task-aligned and greedy variants address these cases empirically. The experiments do not establish that the spectrum universally dominates scalar diagnostics. Profiles matched to deployment perturbations complement Hessian and Lipschitz analyses, which remain appropriate for infinitesimal analysis and worst-case certification.

\subsection{Extended related work} \label{app:related}

Our construction is closest to four lines of work, but differs in the measured object. Sharpness, Lipschitz, and Jacobian methods provide scalar local, worst-case, or average sensitivity summaries \citep{foret2021sharpness,tsuzuku2018lipschitz,novak2018sensitivity}; frequency-resolved robustness studies input-space spectral structure \citep{yin2019fourier,rahaman2019spectral}. Besov approximation theory instead assumes a target belongs to a smoothness class and derives approximation rates \citep{suzuki2019adaptivity,suzuki2021deep}. In PTQ, HAWQ allocates precision using Hessian sensitivity, while task-loss-aware reconstruction methods optimize rounding or block reconstruction \citep{dong2020hawqv2,nagel2020adaround,li2021brecq}. We measure a trained model's response across perturbation scales, direction families, and representations, then use the response at the matching perturbation scale and direction for intervention or allocation. The following paragraphs develop these comparisons.

\paragraph{Flatness, sharpness, and generalization measures.} Flat-minima arguments date to \citet{hochreiter1997flat} and were revived by \citet{keskar2017large}; \citet{dinh2017sharp} showed parameter-space sharpness is reparameterization-sensitive, \citet{jiang2020fantastic} found sharpness measures fragile across settings, and \citet{andriushchenko2023modern} found even reparameterization-invariant adaptive sharpness only weakly related to generalization, consistent with our position that the useful information sits in scale- and direction-resolved structure that any scalar discards. \citet{petzka2021relative} define a relative flatness that is invariant under layerwise rescaling and provably tied to generalization. We study the complementary question of how responses vary across perturbation scales: Theorem~\ref{thm:kink} exhibits a family with identical local Hessians and different finite-scale responses. Input-space sensitivity summaries correlate with generalization empirically \citep{novak2018sensitivity}, and SAM \citep{foret2021sharpness} and SAM variants \citep{kwon2021asam} convert sharpness into a training objective. These methods motivate our comparison of scalar sensitivity summaries with scale-indexed response curves under the same corruption and quantization protocols. The same axis surfaces inside optimization: The curvature threshold that governs gradient descent is not a scale-free property of the loss but one the step size sets, with sharpness hovering at $2/\eta$ at the edge of stability \citep{cohen2021gradient}, so the governing quantity there is already indexed by a finite step rather than read as $r\to0$. We run no optimization experiments, and App.~\ref{app:discussion-extra} lists that setting as an extension rather than a result.

\paragraph{Lipschitz and Jacobian control.} Parseval networks \citep{cisse2017parseval}, spectral normalization \citep{miyato2018spectral}, certified Lipschitz bounds \citep{tsuzuku2018lipschitz}, and Jacobian penalties \citep{hoffman2019robust,drucker1992improving} regularize first-order sensitivity globally or on average; randomized smoothing \citep{cohen2019certified} certifies robustness at one chosen finite noise scale, effectively fixing a single point on the scale axis that we sweep. Such single-scale summaries are blind to kink density (Theorem~\ref{thm:kink}) and expensive to pin down: Exact computation of the Lipschitz constant is NP-hard already for two-layer networks, and the standard layerwise product bound can overestimate that constant by orders of magnitude \citep{virmaux2018lipschitz}, so the practical choice is between a loose certified upper bound and the sampled lower bound our vision tables report. These tools also answer a different question, bounding the worst case over an adversarially chosen perturbation set \citep{madry2018towards}, whereas our spectra average an absolute $p$-moment over a stated direction law at a stated scale and so describe the typical finite-scale response to the perturbations a deployed model actually meets, namely corruptions and quantization noise, and carry no worst-case guarantee. We read worst-case bounds and finite-scale spectra as complementary rather than competing; empirically the scalar sensitivity summaries are less predictive of finite-scale failures than the spectrum, and RSC's nuisance-direction control improves noise robustness with clean costs reported where uniform Jacobian regularization leaves the targeted roughness and the failure unmoved. As motivated in Sec.~\ref{sec:intro}, a single global or average sensitivity summary does not specify which direction families need control at the failure scale. A direction-indexed profile makes this choice explicit, allowing a regularizer to target selected nuisance families while the task objective continues to reward correct classification. The drift--damage dissociation of Sec.~\ref{sec:exp-quant} gives a related reason to retain task information in output space: Output movement can be large without causing comparable task damage.

A broader point applies to both lines of work. These constants are sufficient-condition devices: Each enters a theorem by bounding behavior uniformly over a set, so the guarantee inherits whatever slack the constant carries at the scale of interest. Two of the facts above are that slack made concrete: the layerwise product bound overestimating by orders of magnitude \citep{virmaux2018lipschitz} and reparameterization-invariant adaptive sharpness relating only weakly to what the measure is meant to predict \citep{andriushchenko2023modern}. Theorem~\ref{thm:kink} supplies the sharper version, fixing both constants exactly across a family whose finite-scale variation is unbounded, so no bound stated in those constants can separate the members. These bounds can remain valid yet provide limited discrimination at the perturbation scales of interest. This limited discrimination motivates measuring responses directly at those scales. Consistency losses under input perturbations \citep{zheng2016improving,hendrycks2020augmix} belong to the same family, being first-order roughness penalties at a fixed scale in input space; RSC penalizes second-order margin differences over selected scale bands and nuisance direction families. These differences vanish within affine regions and respond to changes of slope, including maximal-rival switches in the margin; on smooth architectures the second differences measure finite-scale curvature. The auxiliary activation-crossing surrogate is evaluated separately in App.~\ref{app:crossing-penalty}.

\paragraph{Besov spaces and neural networks.} Besov smoothness governs approximation rates of deep ReLU networks \citep{suzuki2019adaptivity,devore1993constructive}, and wavelet and finite-difference characterizations are classical \citep{triebel1983theory}. \citet{suzuki2021deep} study approximation and estimation for anisotropic Besov targets, affine transformations of those targets, and compositions of anisotropic functions. The anisotropic smoothness vector is defined relative to coordinates. Figure~1 of \citet{suzuki2021deep} illustrates how a representation can supply curved coordinates adapted to a data manifold, with different regularity along the resulting axes. This view makes the coordinate map part of the smoothness model. Our Fourier-subspace probes instantiate a fixed linear coordinate choice; App.~\ref{app:directional-besov} relates the Fourier-subspace probes to anisotropic exponents and gives a localized Besov characterization in manifold coordinates, together with difference bounds for compositional representations. Covariance eigenfunctions of Gaussian processes provide another choice of linear perturbation coordinates \citep{kanagawa2018gaussian}; Corollary~\ref{cor:gaussian-directions} relates the resulting normalized perturbations to directional Besov weights. The realized network response is the object measured here, and scale-resolved differences of the network response can also be used for intervention. Theorem~\ref{thm:besov-equiv} and Corollary~\ref{cor:mixed-besov} identify population Besov equivalences for isotropic and mixed direction laws. Proposition~\ref{prop:directional-energy} specializes the connection to the unjittered RSC energy, while Proposition~\ref{prop:uniform-besov} bounds data-dependent moments by classical $B^s_{\infty,q}$ control.

\paragraph{Robustness benchmarks and frequency-resolved analysis.} Corruption benchmarks quantify finite-scale input robustness \citep{hendrycks2019benchmarking}, and frequency-domain analyses connect corruption type to spectral sensitivity \citep{yin2019fourier}, motivating the Fourier direction families used in our vision probes. A separate frequency-resolved literature studies the function a network represents rather than the perturbations the network meets: \citet{rahaman2019spectral} show that ReLU networks fit low frequencies first and that the higher-frequency content of the learned function is both learned later and less robust to parameter perturbation. This spectral-bias analysis is the closest antecedent to the framing here, and the contrast is worth stating: Spectral bias resolves an input-space Fourier decomposition of the learned function over the course of training, whereas we resolve the response amplitude of an already trained model across perturbation scale, direction family, and representation, including perturbations to residual streams whose effects are measured through output distributions. These probes do not require a Fourier decomposition of the learned function.

\paragraph{PTQ and mixed-precision allocation.} PTQ methods such as GPTQ, AWQ, LLM.int8(), SmoothQuant, and OWQ \citep{frantar2022gptq,lin2024awq,dettmers2022llmint8,xiao2023smoothquant,lee2024owq} minimize weight or activation reconstruction error, keep the outlier dimensions in higher precision, or migrate quantization difficulty between the two; and QuaRot \citep{ashkboos2024quarot} removes outliers from the hidden state by computationally invariant rotations before any bit assignment is made. All act at the level of the format and are orthogonal to the axis we test, since our comparisons fix the format and vary only which blocks receive the larger budget; the rotation family is the least predictable of these quantization methods a priori, since that family globally flattens the outlier structure that the per-block damage keys on. Precision allocation typically uses Hessian, Fisher, or salience sensitivity proxies \citep{dong2019hawq,dong2020hawqv2,kim2024squeezellm}, with direct per-layer loss probing appearing as a sensitivity heuristic in systems work \citep{yao2022zeroquant}. Both branches descend from Optimal Brain Damage and Optimal Brain Surgeon \citep{lecun1990optimal,hassibi1993second}, which score a weight by the second-order loss increase induced by removing that weight: GPTQ's layerwise solver is the modern reconstruction-side form of that objective, and HAWQ's per-block trace is the allocation-side form of the same objective; these curvature-based scores form precisely the infinitesimal class Theorem~\ref{thm:alloc-regret} shows to be minimax-insufficient once the perturbation has a finite size. HAWQ-V3 \citep{yao2021hawqv3} pairs the criterion with an integer-only inference graph and an integer linear program over hardware-feasible bit assignments; we compare against the HAWQ-V2 sensitivity score rather than the V3 solver because the object under test is the per-block score, not the constraint solver that consumes the score, and any score, ours included, can be dropped into a V3-style program.

Closest in spirit to our drift--damage distinction is the task-loss-aware line begun by AdaRound \citep{nagel2020adaround}, which shows that rounding chosen to minimize weight error is measurably worse than rounding chosen against a second-order approximation of the task loss, and continued by BRECQ \citep{li2021brecq}, which reconstructs a block at a time precisely because layerwise objectives ignore cross-layer dependence. Our results are the allocation-axis analogue of both observations: Table~\ref{tab:pythia-rc} measures directly that weight-proxy and Hessian-proxy scores mis-rank which blocks break, and the additivity audits and interaction-aware greedy of Sec.~\ref{sec:rsc} answer at the allocation stage the cross-block coupling BRECQ handles at the reconstruction stage, with Proposition~\ref{prop:damage} and Theorem~\ref{thm:alloc-stability} making the size of that coupling explicit. The 4-bit precision range our comparisons allocate within is the one $k$-bit inference scaling laws single out \citep{dettmers2023case}, and the precision-aware scaling laws of \citet{kumar2025precision} are the population-level counterpart of the scale dependence we measure per block. Those laws describe total damage at a fixed bit width as models grow. Our measurements locate that damage within each model and assess how well local-curvature proxies predict the block ranking (Sec.~\ref{sec:exp-modllm}). We score blocks with functional probes of the quantization-noise perturbation, namely output-distribution drift (\eqref{eq:bquant}) and calibration damage (\eqref{eq:dquant}), compared directly against HAWQ-V2 sensitivity-based allocation on the same models and budgets; our contribution there is not the loss probe but the measured dissociation between drift and damage that emerges at 2.8B and 6.9B, the dissociation's depth structure, and the two-regime allocation picture that dissociation implies (Sec.~\ref{sec:exp-quant}). That dissociation is worth stating as a caution with an explicit scope. The caution does not bear on methods that minimize a local weight or activation reconstruction error, a different objective on a different object; the caution bears on the narrower family that scores or selects by how far the full output distribution travels from the full-precision output distribution, which is exactly the quantity we measure reversing against realized damage at the largest models. Where such an objective drives an allocation, the target-labeled counterpart of the drift objective costs the same forward passes. The extreme case of a single dominant block, seen on Hermes-3, a Llama-3.1-8B fine-tune, in Sec.~\ref{sec:exp-modllm}, connects to the massive-activation phenomenon documented by \citet{sun2024massive}: A few hidden coordinates in early layers carry outlier-magnitude values, and a scoring rule that reads only infinitesimal curvature has no reason to flag the block that hosts those coordinates.

\section{Supporting theoretical results} \label{app:theory}

This section gives the supporting theoretical results and finite-grid consequences used in the main text. The canonical Besov equivalence is stated in Sec.~\ref{sec:method}. The main proofs are collected in App.~\ref{app:proofs}; short proofs accompany the corresponding statements.

\paragraph{Scalar blind spots: kink density and affine-region boundaries.} \label{sec:theory-blindspot} Within a straight segment of a piecewise-affine response, the Hessian is zero; a finite step can still cross a change of slope. Second differences can register such crossings, so familiar scalar readouts need not determine the full profile. The following statements exhibit matched Lipschitz constants and almost-everywhere Hessians with different multiscale roughness. The proofs of both statements are in App.~\ref{app:proofs}; the smooth population limit of Remark~\ref{rem:scalar-endpoints} does not apply across these kinks.

\begin{theorem}[Kink density separates models of matched Lipschitz constant and Hessian] \label{thm:kink} Let $\tau(x)\coloneqq\operatorname{dist}(x,\mathbb Z)$ be the 1-periodic triangular wave and $f_N(x) \coloneqq \tau(Nx)/N$, $N\in\mathbb{N}$. Then $\Lip(f_N) = 1$ and $f_N'' = 0$ almost everywhere, both for every $N$, while for $1\le p<\infty$, $0<q\le\infty$, and $1 < s < 1+1/p$, the periodic homogeneous Besov (quasi-)seminorm satisfies $\norm{f_N}_{\dot B^s_{p,q}(\mathbb T)} \asymp N^{s-1}$, with comparison constants independent of $N$. At $p=2$, the critical endpoint satisfies $\norm{f_N}_{\dot B^{3/2}_{2,\infty}(\mathbb T)}\asymp N^{1/2}$, whereas $\norm{f_N}_{\dot B^{3/2}_{2,q}(\mathbb T)}=\infty$ for $q<\infty$. \end{theorem}

\begin{lemma}[Second differences at affine-region boundaries] \label{lem:boundary-local} Let $g$ be continuous and piecewise affine with affine-region boundary set $\mathcal{S}$. If $z, z+h, z+2h$ lie in one affine region, then $\Delta^2_h g(z) = 0$. If a single boundary is crossed transversally, let $u\coloneqq h/\norm h$ and let $\Gamma$ be the norm of the jump in the directional derivative $D_u g$ across that boundary. Then $\norm{\Delta^2_h g(z)} \le \Gamma \norm{h}$. \end{lemma}

For piecewise-affine $g$, second differences suppress within-region affine variation and detect changes of slope. Each crossed boundary contributes a directional slope jump weighted by the boundary's position in the stencil; contributions from multiple boundaries can cancel. With a continuous sampling density, the fine-scale $p=2$ energy is proportional to $r^3$ times the boundary integral of squared slope jumps, weighted by data density and the angular factor $\E_\nu|u\cdot n|^3$ (Proposition~\ref{prop:weighted-kinks}). Margin slope changes can also arise from a switch of the maximal rival class. The same stencil measures finite-scale curvature on smooth architectures, motivating the margin-based intervention of Sec.~\ref{sec:rsc}.

\paragraph{Finite-grid consequence of Theorem~\ref{thm:kink}.} The family in Theorem~\ref{thm:kink} is indistinguishable by Lipschitz constants and almost-everywhere Hessians but separated at rate $N^{s-1}$ by the $s>1$ roughness spectrum. We verify the finite-grid form of this law directly on the construction (App.~\ref{sec:exp-toy}, Figure~\ref{fig:toy}): Across $N\in\{1,\dots,64\}$, the empirical Lipschitz constant is $1.000$ throughout and the second derivative vanishes almost everywhere, while the measured fine-scale spectrum aggregate grows with a fitted exponent of $0.500$, matching the predicted $N^{s-1}=N^{0.5}$, and the scale-resolved curves additionally localize each $f_N$'s kink spacing at $r_j\approx1/N$, information no scalar summary retains.

\paragraph{Finite-scale damage and the precision-allocation problem.} Because \eqref{eq:dquant} is the finite-scale roughness of the task-loss functional, the precision-allocation rule of Sec.~\ref{sec:rsc} can be analyzed exactly, and two of the results below fix the operating range of the rules we deploy. Drift is not damage: Theorem~\ref{thm:drift-damage} makes the gap between task damage and label-free drift exact as a target mismatch, so a large, precisely estimated output movement can be less damaging than a smaller one aimed along the task direction, and only the target-labeled finite difference measures task damage directly.

Infinitesimal information does not suffice either: On an explicit additive class whose blocks share one second-order jet, every deterministic rule reading a block only through that jet, including HAWQ and the earlier Optimal Brain Damage approach, incurs minimax regret $\min(K,L-K)M$, unbounded in the per-block damage scale $M$ (Theorem~\ref{thm:alloc-regret}), against which Theorem~\ref{thm:alloc-stability} gives the additive rule a conditional near-oracle bound. Where cross-block curvature is non-negligible, the interaction-aware greedy of Sec.~\ref{sec:rsc} drops the additive model entirely by re-measuring realized damage along the greedy selection path, which is why we report the greedy rule alongside the additive rule.

\subsection{Classical scalar readouts and derivative limits} \label{app:scalar-endpoints}

Remark~\ref{rem:scalar-endpoints} connects the finite-difference primitive to familiar scalar diagnostics. This subsection makes the worst-case extension, the population limit, and the endpoint normalization explicit.

\paragraph{Worst-case first- and second-order sensitivity.} \eqref{eq:lipschitz-readoff} is the usual pairwise Lipschitz quotient with the displacement written as $h=ru$; the quotient requires no differentiability. On an open convex domain, integration along line segments and directional differentiation give $\Lip(g)=\sup_z\norm{\nabla g(z)}_2$ for $g\in C^1$. Likewise, for $g\in C^2$, $\Lip(\nabla g)=\sup_z\norm{H_g(z)}_{\mathrm{op}}$, the usual uniform gradient-smoothness constant. A supremum restricted to sampled base points or directions does not establish either global constant.

\paragraph{Population limits with radius jitter.} Consider the full grid $j\ge0$, $m\in\{1,2\}$, $g\in C^m$, and a fixed base-point law $\mu$ with a scale-independent direction law $\mathcal H(z)$. Suppose all sufficiently small probe stencils lie in $\mathcal Z$ and the scaled differences $r_j^{-m}\Delta^m_{r_j\xi u}g(z)$ are dominated in absolute value by a common $L^p$ function under the joint probe law of $(z,u,\xi)$, with $1\le p<\infty$. Bounded $m$-th derivatives on a neighborhood containing all sufficiently small probes are sufficient. Taylor expansion and dominated convergence then give \begin{equation} \begin{gathered} \lim_{j\to\infty}r_j^{s-m}B_j(g) =c_{m,p}(\E_{z,u}\abs{D_u^m g(z)}^p)^{1/p},\\ c_{m,p}\coloneqq(\E\xi^{mp})^{1/p},\qquad \E\xi^{mp}=\frac{2^{mp/2}-2^{-mp/2}}{mp\ln2}. \end{gathered} \label{eq:population-derivative-limit} \end{equation} Here $D_u g(z)=\nabla g(z)^\top u$ and $D_u^2g(z)=u^\top H_g(z)u$. Independence of the jitter produces the factor $c_{m,p}$; the explicit value of $c_{m,p}$ follows by integrating the density $1/(\xi\ln2)$ on $[2^{-1/2},2^{1/2}]$.

\paragraph{Recovered Hessian statistics.} For a fixed base point, let $H=H_g(z)$ and $u\sim\mathrm{Unif}(S^{d-1})$. The spherical second and fourth moments yield \begin{align} \E_u[u^\top Hu]&=\frac{\operatorname{tr}H}{d}, &\sup_{\norm{u}_2=1}\abs{u^\top Hu}&=\norm{H}_{\mathrm{op}}, \label{eq:hessian-readouts}\\ \E_u\abs{u^\top Hu}^2 &=\frac{(\operatorname{tr}H)^2+2\norm{H}_F^2}{d(d+2)}. \label{eq:hessian-rms} \end{align} The moment identities follow from $\E_u[uu^\top]=I/d$ and $\E_u[u_i u_j u_k u_l] =(\delta_{ij}\delta_{kl}+\delta_{ik}\delta_{jl}+\delta_{il}\delta_{jk})/[d(d+2)]$. Thus the $m=s=p=2$ endpoint in \eqref{eq:population-derivative-limit} is the root-mean-square directional curvature over both $z$ and $u$, multiplied by $c_{2,2}$. The trace uses a signed average of the same quadratic form; the absolute-moment spectrum does not retain that sign.

\paragraph{Endpoint scope.} \eqref{eq:population-derivative-limit} recovers derivative statistics at $s=m$; under the hypotheses of the population limit, $B_j\to0$ when $s<m$. The Besov equivalence of Theorem~\ref{thm:besov-equiv} concerns the multiscale aggregate in the strict range $0<s<m$, not these derivative endpoints. In particular, the worst-case first-order endpoint is not an identification of Lipschitz space with $B^1_{\infty,\infty}$, the classical Zygmund space \citep{triebel1983theory}. Almost-everywhere existence of a Hessian alone does not justify the population limit: The kink family of Theorem~\ref{thm:kink} has zero Hessian almost everywhere while the family's second differences register boundary crossings. The required smoothness and domination are what separate that example from the derivative limit.

\subsection{Canonical Besov connection and finite-band estimation} \label{sec:ambient}

The canonical population equivalence is stated in Theorem~\ref{thm:besov-equiv} of Sec.~\ref{sec:method}, under Assumption~\ref{asm:ambient}. For $p=q=2$, Proposition~\ref{prop:finite-band-estimation} bounds the sampling error in estimates of the squared roughness aggregate over a finite scale band. Proposition~\ref{prop:trunc} bounds the contribution of unmeasured finer scales under additional smoothness. The full proof of the population equivalence is in App.~\ref{app:proof-besov}.

The proof in App.~\ref{app:proofs} compares the sampling measures and difference aggregates before applying a classical characterization. Assumption (A1) exchanges $\mu$ for Lebesgue measure with lower and upper factors $c_\mu^{1/p}$ and $C_\mu^{1/p}$, equivalently a condition-number loss of $(C_\mu/c_\mu)^{1/p}$. The jittered isotropic probe law is, by an explicit change of variables, an annulus mean of $m$-th differences with a $1/t$ radial weight. A dyadic covering argument plus a discrete Hardy-type inequality converts annulus means into the classical ball means of differences, in both directions, at the level of the $s$-weighted $q$-aggregate. Ball means of differences characterize $B^s_{p,q}$ \citep[\S2.5.12]{triebel1983theory}; for the periodic case, see \citet[Ch.~3]{schmeisser1987topics}.

\begin{proposition}[Population estimation on a finite band] \label{prop:finite-band-estimation} Fix a scalar response $g$, a finite grid $J$, and the probe laws in Definition~\ref{def:spectrum}, with $p=q=2$. Let $Z_1,\ldots,Z_n$ be independent with law $\mu$. Conditional on $Z_i$, draw $K$ independent copies of the entire probe tuple $(h_j)_{j\in J}$; the samples for different $i$ are independent. Define \begin{equation} \begin{gathered} W_{ik}=\sum_{j\in J}r_j^{-2s} \abs{\Delta^m_{h_{ik,j}}g(Z_i)}^2,\qquad \widehat E_J=\frac1{nK}\sum_{i=1}^n\sum_{k=1}^K W_{ik},\\ E_J=(R^{(m),J}_{s,2,2}(g))^2,\qquad w(z)=\E[W_{ik}\mid Z_i=z]. \end{gathered} \label{eq:finite-band-estimator} \end{equation} If $W_{ik}$ has a finite second moment, then \begin{equation} \E\widehat E_J=E_J,\qquad \operatorname{Var}(\widehat E_J) =\frac{\operatorname{Var}(w(Z))}{n} +\frac{\E[\operatorname{Var}(W_{ik}\mid Z_i)]}{nK}. \label{eq:finite-band-variance} \end{equation} If $\abs{\Delta^m_{h_j}g(Z)}\le L_j$ almost surely, put $b_J=\sum_{j\in J}r_j^{-2s}L_j^2$. For $0<\delta<1$, with probability at least $1-\delta$, \begin{equation} \abs{\widehat E_J-E_J} \le \epsilon_n(\delta) \coloneqq b_J\sqrt{\frac{\log(2/\delta)}{2n}}. \label{eq:finite-band-concentration} \end{equation} \end{proposition}

\begin{proof} Conditional expectation gives the mean. Conditional independence gives $\operatorname{Var}(K^{-1}\sum_k W_{ik}\mid Z_i) =\operatorname{Var}(W_{ik}\mid Z_i)/K$, and the law of total variance gives \eqref{eq:finite-band-variance}. The $n$ independent averages $K^{-1}\sum_kW_{ik}$ lie in $[0,b_J]$, so Hoeffding's inequality gives \eqref{eq:finite-band-concentration}. \end{proof}

The first variance term records variation between base points; the second records variation between probes at a fixed point. For continuous responses on a compact probe domain, $L_j=U_j^{(m)}(g;\Omega)$ from Proposition~\ref{prop:uniform-besov} supplies the envelope. The result also applies to labeled margins by sampling $(X,Y)$ and using the same $g_Y$ throughout each stencil. Population Besov comparisons then apply to the class-conditional laws before averaging over labels.

\begin{proposition}[Fine-scale truncation bound] \label{prop:trunc} Let $R_J$ be the aggregate over the finite grid $J=\{0,\dots,j_1\}$ actually measured, and let $1\le q\le\infty$. Under Assumption~\ref{asm:ambient}, if $g\in B^{s'}_{p,\infty}(\mathbb{T}^d)$ for some $s'\in(s,m)$, then \begin{equation} 0\le R^{(m),\mathbb{N}}_{s,p,q}(g) - R_J(g) \le C\norm{g}_{B^{s'}_{p,\infty}} 2^{-j_1(s'-s)}, \label{eq:trunc} \end{equation} with $C=C(d,m,s,s',p,q,c_\mu,C_\mu,r_0)$: The omitted fine-scale tail decays geometrically in the grid depth at rate $s'-s$. \end{proposition}

For $p=q=2$, the envelope condition above and the hypotheses of Proposition~\ref{prop:trunc} give, on the event in \eqref{eq:finite-band-concentration}, \begin{equation} \abs{\sqrt{\widehat E_J}-R^{(m),\mathbb N}_{s,2,2}(g)} \le \sqrt{\epsilon_n(\delta)} +C\norm{g}_{B^{s'}_{2,\infty}}2^{-j_1(s'-s)} \label{eq:sampling-truncation} \end{equation} by $|\sqrt a-\sqrt b|\le\sqrt{|a-b|}$. This sampling-and-truncation bound estimates the difference component of the Besov characterization in Theorem~\ref{thm:besov-equiv}. Under (A1), the Besov characterization's additional $L^2$ term can equivalently use $\mu$, with squared norm estimated by $n^{-1}\sum_i|g(Z_i)|^2$.

App.~\ref{sec:exp-besovequiv} validates this canonical connection against a closed-form equivalent Littlewood--Paley seminorm: On lacunary series, which place one frequency per Littlewood--Paley block so that the chosen dyadic block seminorm is exact, the estimator's measured-to-exact ratio stays inside $[2.6,10.9]$ across all $108$ $(\alpha,\mathrm{window},s,p,q)$ cells while the exact (quasi-)seminorm sweeps a $482\times$ range, and a square-wave family reproduces both the predicted exponent and the leading constant to under one percent.

\paragraph{Regularized base measures.} On the torus, a fixed smoothing scale gives a density satisfying (A1) for any base law. Let $K_\varepsilon$ be the density of $\varepsilon\zeta$ modulo $\mathbb Z^d$, with $\zeta\sim\mathcal N(0,I_d)$ and $\varepsilon>0$, and set $\mu_\varepsilon=\mu*K_\varepsilon$. The density of $\mu_\varepsilon$ obeys \begin{equation} 0<\min_{\mathbb T^d}K_\varepsilon \le \rho_\varepsilon(z) =\int K_\varepsilon(z-x) d\mu(x) \le\max_{\mathbb T^d}K_\varepsilon<\infty. \label{eq:regularized-base-density} \end{equation} The bounds follow from positivity and continuity of the periodized Gaussian on the compact torus. Thus Theorem~\ref{thm:besov-equiv} and Corollary~\ref{cor:mixed-besov} apply to the same difference construction with base law $\mu_\varepsilon$, including when $\mu$ is empirical. Here $\varepsilon$ fixes the base-point smoothing scale independently of the probe radii $r_j$; the comparison constants depend on $\varepsilon$ and the dimension.

\begin{remark}[Homogeneous Besov counterpart] \label{rem:besov-anchor} For scalar $g$ on $\R^d$ with the base-point expectation in \eqref{eq:spectrum} replaced by Lebesgue integration, isotropic $\mathcal{H}_j$, $m>s$, and the full bilateral dyadic grid $J=\mathbb{Z}$, the resulting aggregate is equivalent to the classical finite-difference Besov (quasi-)seminorm $\dot B^s_{p,q}$ \citep{triebel1983theory}. These population characterizations and the uniform construction give Besov spectra; sampling the finite differences under task-relevant base-point and direction laws gives the empirical roughness spectra. \end{remark}

\subsection{Directional Besov energies and anisotropic coordinates} \label{app:directional-besov}

\paragraph{Subspace directions and anisotropic coordinates.} Gaussian perturbations connect direction families to covariance coordinates. In the Karhunen--Lo\`{e}ve representation of a Gaussian process, covariance eigenfunctions specify orthogonal modes, and eigenvalues give the variances of the modes' independent Gaussian coefficients \citep{kanagawa2018gaussian}. On an image grid these modes are vectors, giving the covariance eigendecomposition used in principal component analysis. A fixed nonzero subspace $E\subseteq\R^d$ specifies a direction family through the orthogonal projector $P_E$ onto $E$: \begin{equation} u=\frac{P_E\varepsilon}{\norm{P_E\varepsilon}_2},\qquad \varepsilon\sim\mathcal N(0,I_d),\qquad \sigma_E\coloneqq\mathrm{Unif}(E\cap S^{d-1}). \label{eq:subspace-directions} \end{equation} Rotational invariance within $E$ gives $u\sim\sigma_E$. An image Fourier band defines such a subspace: The subspace coordinates are the retained modes in a real orthonormal Fourier basis. High-frequency (HF) and low-frequency probes measure variation along distinct groups of input coordinates, with the band boundaries specifying $P_E$ (App.~\ref{app:measurement}). The probes measure the smoothness of the response as these image coefficients vary. This coordinate view connects to anisotropic Besov models, which assign smoothness to directions in a chosen coordinate system, including intermediate representations in compositional models \citep{suzuki2021deep}. The isotropic--HF mixture is \begin{equation} \nu_\alpha=\alpha\sigma_{\R^d}+(1-\alpha)\sigma_{E_{\mathrm{HF}}}, \qquad 0<\alpha\le1, \label{eq:direction-mixture} \end{equation} with $\alpha=1/2$ for RSC.

\begin{corollary}[Besov equivalence with mixed direction families] \label{cor:mixed-besov} Retain (A1), (A3), and the radius jitter in Assumption~\ref{asm:ambient}. At each scale let the base-point-independent direction law be $\nu_j=\alpha_j\sigma_{\R^d}+(1-\alpha_j)\eta_j$, where $\eta_j$ is any probability law on $S^{d-1}$ and $1\ge\alpha_j\ge\alpha_*>0$. Then \eqref{eq:ambient} holds for the aggregate using $\nu_j$, with constants also depending on $\alpha_*$. In particular, the equivalence holds for \eqref{eq:direction-mixture}. \end{corollary}

The isotropic component controls all directions, while the subspace component increases the weight on a specified nuisance family. For $p=q=2$ and $0<s<2$, Proposition~\ref{prop:angular-besov} identifies the directional energy with the Fourier weight $\E_\nu|\omega\cdot u|^{2s}$, without radius jitter. A subspace law measures Besov regularity along the subspace coordinates; a direction law with positive definite second moment gives a full $B^s_{2,2}$ characterization. At $s=1$, Corollary~\ref{cor:gaussian-directions} expresses this population energy as a weighted sum of squared derivatives along covariance eigenvectors, with weights determined by the normalized Gaussian directions. The remainder of this subsection develops this connection and extends the directional characterization to direction-specific smoothness exponents. For data on a manifold, Proposition~\ref{prop:chart-besov} gives a localized Besov characterization in intrinsic coordinates with a bounded positive density.

Corollary~\ref{cor:mixed-besov} allows a subspace component in the jittered construction. Gaussian covariance operators provide a coordinate description of these directions. In the quadratic case, an angular Fourier weight characterizes the resulting Besov energy throughout $0<s<2$; the slope normalization used by RSC has an exact derivative representation at $s=1$.

\paragraph{Gaussian covariance coordinates.} For a centered Gaussian field $\eta(t)$ on an image domain $T$ with $\E\norm{\eta}_{L^2(T)}^2<\infty$, write $K(t,t')=\E[\eta(t)\eta(t')]$ for the field's covariance kernel. The covariance operator $(Cf)(t)=\int_T K(t,t')f(t') dt'$ has orthonormal eigenfunctions $e_k$ with nonnegative eigenvalues $\lambda_k$. The Gaussian field's Karhunen--Lo\`{e}ve representation is \begin{equation} \eta(t)=\sum_k\sqrt{\lambda_k}\zeta_k e_k(t), \qquad \zeta_k\overset{\mathrm{iid}}{\sim}\mathcal N(0,1), \label{eq:gp-kl} \end{equation} with convergence in mean square in $L^2(T)$ \citep[Sec.~4.1.3]{kanagawa2018gaussian}. The index $t$ specifies image coordinates, $e_k$ specifies a perturbation mode, and $\lambda_k$ gives the variance in that mode. These modes define coordinates for perturbing the input; the Besov energy measures the response $g$ as the mode coefficients vary. On a finite image grid, the construction becomes the covariance eigendecomposition of a Gaussian vector in $\R^d$. For a nonzero covariance matrix $C=\sum_{k=1}^d\lambda_k e_ke_k^\top\succeq0$ with an orthonormal eigenbasis, normalization gives the direction law \begin{equation} u_C=\frac{\sum_{k=1}^d\sqrt{\lambda_k}\zeta_k e_k} {(\sum_{k=1}^d\lambda_k\zeta_k^2)^{1/2}}, \qquad \nu_C\coloneqq\mathcal L(u_C). \label{eq:gaussian-covariance-directions} \end{equation} The unit direction retains the relative covariance structure, while the probe radius sets the perturbation magnitude.

For the population identities below, $L^2$ norms and integrals use Lebesgue measure on $D=\R^d$ or $\mathbb T^d$, normalized in the periodic case.

\begin{proposition}[Angular characterization of quadratic Besov energies] \label{prop:angular-besov} Let $\nu$ be a fixed probability law on $S^{d-1}$, independent of the base point, let $r_j=r_02^{-j}$ with $r_0>0$, and let $0<s<2$. Define \begin{equation} \begin{split} \mathcal E_{\nu,s}^{J}(g) &\coloneqq\sum_{j\in J}r_j^{-2s} \E_{u\sim\nu}\norm{\Delta^2_{r_ju}g}_{L^2(D)}^2,\\ \Psi_{\nu,s}(\omega)&\coloneqq\E_{u\sim\nu}|\omega\cdot u|^{2s}, \qquad M_\nu\coloneqq\E_{u\sim\nu}uu^\top. \end{split} \label{eq:angular-energy-def} \end{equation} For $g\in L^2(\R^d)$, \begin{equation} \mathcal E_{\nu,s}^{\mathbb Z}(g) \asymp\int_{\R^d}|\widehat g(\omega)|^2\Psi_{\nu,s}(\omega) d\omega. \label{eq:angular-fourier-energy} \end{equation} On $\mathbb T^d$, the corresponding expression is a sum over Fourier frequencies $\omega\in2\pi\mathbb Z^d$. If $M_\nu$ is positive definite, then $\Psi_{\nu,s}(\omega)\asymp\norm{\omega}_2^{2s}$ and \begin{equation} \norm{g}_{L^2(D)}^2+\mathcal E_{\nu,s}^{\mathbb N}(g) \asymp\norm{g}_{B^s_{2,2}(D)}^2. \label{eq:angular-besov-norm} \end{equation} For a nonzero subspace $E$, the law $\sigma_E$ instead gives $\Psi_{\sigma_E,s}(\omega)=\kappa_{\dim E,s}\norm{P_E\omega}_2^{2s}$, where $\kappa_{k,s}>0$ is the $2s$-moment of one coordinate of a uniform unit vector in $\R^k$. On $\R^d$, adding the $L^2$ term therefore gives the squared norm of $L^2(E^\perp;B^s_{2,2}(E))$. \end{proposition}

Thus an HF spectrum measures the response's regularity as the HF coordinates vary, integrated over the remaining coordinates. The mixture in \eqref{eq:direction-mixture} has a positive definite $M_\nu$ and gives \eqref{eq:angular-besov-norm} for every $0<s<2$, including both the training and evaluation exponents. Independent radius jitter in $[2^{-1/2},2^{1/2}]$ preserves these equivalences, and (A1) transfers these equivalences to density-weighted moments on the torus. The proof is in App.~\ref{app:proof-angular-besov}.

\begin{proposition}[Directional slope energy on an unjittered dyadic grid] \label{prop:directional-energy} With the notation of \eqref{eq:angular-energy-def}, set \begin{equation} G_\nu(r)\coloneqq r^{-2}\E_{u\sim\nu}\norm{\Delta_{ru}g}_{L^2(D)}^2. \label{eq:directional-energy-def} \end{equation} For $g\in L^2(D)$ and integers $a\le b$, the finite-band identity is \begin{equation} \mathcal E_{\nu,1}^{\{a,\ldots,b\}}(g) =4[G_\nu(r_b)-G_\nu(2r_a)]. \label{eq:directional-telescope} \end{equation} For $g\in H^1(D)$, the full-scale limit of the dyadic energy is \begin{equation} \mathcal E_{\nu,1}^{\mathbb Z}(g) =4\int_D\nabla g(x)^\top M_\nu\nabla g(x) dx. \label{eq:directional-dirichlet} \end{equation} If $M_\nu$ is positive definite, then \begin{equation} \norm{g}_{L^2(D)}^2+\mathcal E_{\nu,1}^{\mathbb N}(g) \asymp\norm{g}_{B^1_{2,2}(D)}^2 \label{eq:directional-besov-norm} \end{equation} for all $g\in L^2(D)$, with the value $+\infty$ allowed. The comparison constants depend on $r_0$, $d$, and the extreme eigenvalues of $M_\nu$. On the torus, \eqref{eq:directional-besov-norm} also holds when the base-point integral in $\mathcal E_{\nu,1}^{\mathbb N}$ uses any density satisfying (A1). \end{proposition}

\begin{corollary}[Gaussian covariance directions] \label{cor:gaussian-directions} For $\nu_C$ in \eqref{eq:gaussian-covariance-directions}, define \begin{equation} w_k(C)\coloneqq \E\left[\frac{\lambda_k\zeta_k^2} {\sum_{\ell=1}^d\lambda_\ell\zeta_\ell^2}\right]. \label{eq:gaussian-angular-weights} \end{equation} Then $M_{\nu_C}=\sum_{k=1}^d w_k(C)e_ke_k^\top$, and for $g\in H^1(D)$, \begin{equation} \mathcal E_{\nu_C,1}^{\mathbb Z}(g) =4\sum_{k=1}^d w_k(C)\norm{\partial_{e_k}g}_{L^2(D)}^2, \qquad \partial_{e_k}g\coloneqq e_k^\top\nabla g. \label{eq:gaussian-directional-besov} \end{equation} If $C$ is positive definite, \eqref{eq:angular-besov-norm} holds with $\nu=\nu_C$ for every $0<s<2$. At general $s$, the Gaussian directional energy's Fourier weight is $\Psi_{\nu_C,s}$; at $s=1$, \eqref{eq:gaussian-directional-besov} gives the exact covariance-coordinate decomposition. \end{corollary}

\begin{proof} In the covariance eigenbasis, sign symmetry of each Gaussian coefficient makes the off-diagonal entries of $\E[u_Cu_C^\top]$ zero. The diagonal entries $w_k(C)$ sum to one and are positive exactly when $\lambda_k>0$. Substituting this matrix into Proposition~\ref{prop:directional-energy} gives \eqref{eq:gaussian-directional-besov}; Proposition~\ref{prop:angular-besov} gives the norm equivalences. \end{proof}

For $C=P_E$, rotational invariance gives $\nu_C=\sigma_E$ and $M_{\nu_C}=P_E/\dim E$. The isotropic and HF Gaussian components have covariances $I_d$ and $P_{E_{\mathrm{HF}}}$, respectively. Mixing the normalized isotropic and HF direction laws as in \eqref{eq:direction-mixture} gives \begin{equation} \begin{split} M_{\nu_\alpha}&=\frac{\alpha}{d}I_d +\frac{1-\alpha}{d_{\mathrm{HF}}}P_{E_{\mathrm{HF}}},\\ \mathcal E_{\nu_\alpha,1}^{\mathbb Z}(g) &=\frac{4\alpha}{d}\norm{\nabla g}_2^2 +\frac{4(1-\alpha)}{d_{\mathrm{HF}}}\norm{P_{E_{\mathrm{HF}}}\nabla g}_2^2. \end{split} \label{eq:mixture-gradient-weight} \end{equation} Together with the $L^2$ term, this mixture energy is an elliptic, directionally weighted realization of $B^1_{2,2}=H^1$ for $0<\alpha\le1$: The HF component assigns additional weight to variation in the HF subspace while the isotropic component supplies a positive weight in every direction. The finite band retains the secant-slope interpretation of RSC. Under Lebesgue integration, \eqref{eq:directional-telescope} expresses RSC's second-difference energy as the increase in first-difference energy between the band endpoints. Under the training distribution, \eqref{eq:rsc-energy} samples the same finite-band differences as data-weighted moments; the corresponding translation correction is given in \eqref{eq:weighted-telescope}.

For a common base law, response, and radius convention, the quadratic mixture also decomposes at each finite band: \begin{equation} (R_{\nu_\alpha,s,2,2}^{(2),J})^2 =\alpha(R_{\mathrm{iso},s,2,2}^{(2),J})^2 +(1-\alpha)(R_{\mathrm{HF},s,2,2}^{(2),J})^2. \label{eq:finite-band-mixture} \end{equation} This finite-band decomposition follows by linearity of expectation at each scale and holds for data-weighted moments as well as Lebesgue integrals.

\paragraph{RSC population interpretation.} The direction law is $\nu_{1/2}$ in \eqref{eq:direction-mixture}. For each class margin, the full-scale Lebesgue counterpart is a directionally weighted $B^1_{2,2}$ energy: For $d_{\mathrm{HF}}=\dim E_{\mathrm{HF}}$, the associated gradient weight is $M_{\nu_{1/2}}=I_d/(2d)+P_{E_{\mathrm{HF}}}/(2d_{\mathrm{HF}})$ (Proposition~\ref{prop:directional-energy}). These gradient weights are the second-moment weights of the normalized isotropic and HF Gaussian components (Corollary~\ref{cor:gaussian-directions}). RSC samples a finite band of this directional energy under the training distribution. The data-weighted finite-band identity in Proposition~\ref{prop:weighted-telescope} separates the change in secant-slope energy from the change in the base measure under translation. Uniform $B^1_{\infty,\infty}$ control of the class margins on a box containing the probes also bounds this data-weighted objective (\eqref{eq:rsc-uniform-besov}). The second moment retains both the size and frequency of slope changes along the sampled nuisance directions.

\paragraph{Fixed coordinates and distinct smoothness exponents.} Let $Q$ be an orthogonal change of coordinates on $\R^d$ and write $\widetilde g(z)=g(Q^\top z)$. Then \begin{equation} \Delta_h^2g(x)=\Delta_{Qh}^2\widetilde g(Qx). \label{eq:linear-coordinate-difference} \end{equation} For a real orthonormal Fourier basis, $E_{\mathrm{HF}}$ and $E_{\mathrm{LF}}$ are coordinate subspaces in $z$. The subspaces' frequency labels refer to the spatial modes of an input image; the spectrum measures the smoothness of $\widetilde g$ as these image coefficients vary.

The anisotropic Besov construction assigns an exponent $\beta_i$ to each coordinate \citep{suzuki2021deep}. For $0<\beta_i<2$, one equivalent inhomogeneous $p=q=2$ anisotropic Besov norm on $\R^d$ is \begin{equation} \begin{split} \norm{g}_{B^{\boldsymbol\beta}_{2,2}(Q)}^2 &\asymp\norm{g}_2^2+ \sum_{k\ge0}2^{2k}\omega_2(\widetilde g, (2^{-k/\beta_i})_{i=1}^d)_2^2,\\ \omega_2(f,\mathbf t)_2 &\coloneqq\sup_{|v_i|\le t_i}\norm{\Delta_v^2f}_2. \end{split} \label{eq:anisotropic-modulus} \end{equation} The notation $B^{\boldsymbol\beta}_{2,2}(Q)$ records the chosen coordinate system. Grouping coordinates into orthogonal blocks gives a directional characterization of this norm.

\begin{proposition}[Block anisotropic Besov energy] \label{prop:block-anisotropic} Let $\R^d=E_1\oplus\cdots\oplus E_B$ be an orthogonal decomposition into nonzero subspaces, and choose $Q$ whose axes are orthonormal bases of these subspaces. Assign $s_b\in(0,2)$ and $\lambda_b>0$ to block $E_b$, and set $\beta_i=s_b$ on the coordinates of $E_b$. For $g\in L^2(\R^d)$, \begin{equation} \norm{g}_2^2+ \sum_{b=1}^B\lambda_b\sum_{j\in\mathbb Z}r_j^{-2s_b} \E_{u\sim\sigma_{E_b}}\norm{\Delta^2_{r_ju}g}_2^2 \asymp\norm{g}_{B^{\boldsymbol\beta}_{2,2}(Q)}^2. \label{eq:block-anisotropic-besov} \end{equation} The constants may depend on the block dimensions, exponents, and weights, but not on $g$. \end{proposition}

This formulation allows smoothness to differ between frequency bands. The RSC population energy in \eqref{eq:mixture-gradient-weight} has the common exponent $s_b=1$, with unequal weights on the HF block and the orthogonal complement of the HF block. Distinct $s_b$ in \eqref{eq:block-anisotropic-besov} encode different regularity orders in those coordinates. Proofs of the directional results are in Apps.~\ref{app:proof-mixed-besov} through~\ref{app:proof-block-anisotropic}.

\paragraph{Compositional representations and curved coordinates.} The coordinate map can also be part of a composition $g=F\circ\Phi$, as in the deep anisotropic models of \citet{suzuki2021deep}. In Figure~1 of that study, coordinates follow a data manifold and separate directions with different regularity. An intermediate representation can supply the corresponding coordinate variables; a multiresolution representation can further group these coordinates by spatial scale. Probing $F$ in those variables measures the directional regularity of $F$ directly. For a local chart $\Psi:U\subset\R^k\to\mathcal M$, the coordinate response $f=g\circ\Psi$ has differences \begin{equation} \begin{gathered} \Delta_v^2f(z) =g(\Psi(z+2v))-2g(\Psi(z+v))+g(\Psi(z)), \\ z,z+v,z+2v\in U. \end{gathered} \label{eq:chart-difference} \end{equation} The coordinate map and a direction law in the coordinate domain together specify the probes. A law supported on a curved $k$-dimensional manifold can have a bounded positive density in the manifold's $k$ intrinsic variables. Localizing within a chart yields a Besov characterization under that coordinate law.

\begin{proposition}[Besov energy in intrinsic coordinates] \label{prop:chart-besov} Let $\Psi:U\to\mathcal M$ be a $C^2$ chart, where $U\subset\R^k$ is a bounded open box. Choose $\chi\in C_c^\infty(U)$ and extend $f=\chi(g\circ\Psi)$ by zero to $\R^k$, with $f\in L^2(\R^k)$. Let $Z$ be supported on $U$ with density $0<c\le\rho_Z\le C<\infty$ there. Choose a fixed unit-direction law $\nu$ on $\R^k$ with $M_\nu\succ0$, independent of $Z$, and the independent radius jitter of Definition~\ref{def:spectrum}. If $2\sqrt2r_0<\operatorname{dist}(\operatorname{supp}\chi,\partial U)$, then for $0<s<2$, \begin{equation} \norm{f}_{L^2(\rho_Z)}^2+ \sum_{j\ge0}r_j^{-2s}\E\abs{\Delta^2_{r_j\xi u}f(Z)}^2 \asymp \norm{f}_{B^s_{2,2}(\R^k)}^2. \label{eq:chart-besov} \end{equation} Together with the same $L^2$ term, the analogous sum of coordinate-block energies in Proposition~\ref{prop:block-anisotropic}, with $j\ge0$, characterizes $B^{\boldsymbol\beta}_{2,2}(Q)$ of $f$ for an orthogonal decomposition of $\R^k$ and the exponents and weights specified there. The comparison constants are independent of $g$. \end{proposition}

\begin{proof} For every probe displacement $h$, the radius condition places the support of $\Delta_h^2 f$ inside $U$. Density comparison therefore gives $\E_Z|\Delta_h^2f(Z)|^2\asymp\|\Delta_h^2f\|_{L^2(\R^k)}^2$, and likewise for the $L^2$ term. Proposition~\ref{prop:angular-besov}, including the bounded-jitter version of the proposition, gives \eqref{eq:chart-besov}. For the block energies, apply Proposition~\ref{prop:block-anisotropic}; the omitted $j<0$ part is bounded by a constant times $\|f\|_2^2$ since $\|\Delta_h^2f\|_2\le4\|f\|_2$ and each $s_b>0$. \end{proof}

If $g=F\circ\Phi$ and $\Phi\circ\Psi=\operatorname{id}_U$, the localized response in this proposition is $\chi F$. A representation supplying these intrinsic coordinates thus exposes the response's Besov regularity under a density condition in dimension $k$. A uniform coordinate modulus gives the corresponding bound for arbitrary coordinate laws. In tubular coordinates $\Psi(z,n)$ with $\Psi(z,0)\in\mathcal M$, the response $g(\Psi(z,n))$ additionally records variation transverse to the manifold. Tangential and transverse coordinates can carry distinct exponents in the same block construction. The density and direction conditions apply in the coordinate variables being probed.

For input-space probes of a composition, write $\delta=\Phi(x+h)-\Phi(x)$. If $F$ is Lipschitz on a feature domain containing the following stencil points, direct subtraction gives \begin{equation} \abs{\Delta_h^2g(x)-\Delta_\delta^2F(\Phi(x))} \le\Lip(F)\norm{\Delta_h^2\Phi(x)}. \label{eq:composition-direction-remainder} \end{equation} The remainder records second differences of the coordinate map. To compare entire quadratic spectra, let $h_j=r_j\xi u$, put $\delta_j=\Phi(x+h_j)-\Phi(x)$, and define \begin{align*} A_j^\Phi(F)&\coloneqq (\E|\Delta_{\delta_j}^2F(\Phi(x))|^2)^{1/2},\\ R_{s,2,2}^{\Phi,J}(F)&\coloneqq \left(\sum_{j\in J}r_j^{-2s}A_j^\Phi(F)^2\right)^{1/2}. \end{align*} The expectation uses the same base points and probes as the input spectrum. For finite moments and finite $J$, the reverse triangle inequalities in $L^2$ and $\ell^2$ give \begin{equation} \abs{R^{(2),J}_{s,2,2}(g)-R_{s,2,2}^{\Phi,J}(F)} \le\Lip(F)R^{(2),J}_{s,2,2}(\Phi). \label{eq:composition-spectrum-control} \end{equation} Here the spectrum of the vector-valued map $\Phi$ uses the Euclidean norm in \eqref{eq:spectrum}. For a piecewise-affine representation, the term $R^{(2),J}_{s,2,2}(\Phi)$ records slope changes at activation boundaries. If $\norm{D^2\Phi}_{\mathrm{op}}\le K$ on all probe segments, the bound becomes \begin{equation} \abs{R^{(2),J}_{s,2,2}(g)-R_{s,2,2}^{\Phi,J}(F)} \le\Lip(F)K(\E\xi^4)^{1/2} \left(\sum_{j\in J}r_j^{4-2s}\right)^{1/2}. \label{eq:composition-spectrum-remainder} \end{equation} Indeed, $\norm{\Delta_{h_j}^2\Phi(x)}\le K\norm{h_j}^2$ in \eqref{eq:composition-direction-remainder}. For $0<s<2$, the right-hand side over $j\ge a$ is $O(r_a^{2-s})$, uniformly in the number of retained fine scales. Smooth coordinate curvature therefore contributes a summable remainder, while the induced increments $\delta_j$ describe how the probe law acts in feature coordinates. For linear Fourier coordinates the remainder vanishes exactly.

\subsection{Data-weighted differences and boundary regularity} \label{app:data-weighted-besov}

Replacing Lebesgue integration by $L^2(\mu)$ retains the difference operator and scale weights. The effect on the slope-energy identity can be expressed through translations of the base measure, even when $\mu$ is supported on a lower-dimensional set.

\begin{proposition}[Translation correction for data-weighted slope energy] \label{prop:weighted-telescope} Let $\mu$ be a probability measure on $D=\R^d$ or $\mathbb T^d$, and let $\nu$ be a fixed direction law independent of the base point. Write $\mu_h=(x\mapsto x+h)_\#\mu$ and, for scalar $g$ with finite displayed difference moments, define \begin{equation} \begin{split} G_{\mu,\nu}(r)&\coloneqq r^{-2}\E_\nu \norm{\Delta_{ru}g}_{L^2(\mu)}^2,\\ \mathcal D_{\mu,\nu}(r)&\coloneqq 2r^{-2}\E_\nu \int_D|\Delta_{ru}g(x)|^2 d(\mu_{ru}-\mu)(x). \end{split} \label{eq:measure-translation-correction} \end{equation} For $r_j=r_02^{-j}$ and integers $a\le b$, \begin{equation} \begin{split} \sum_{j=a}^b r_j^{-2}\E_\nu\norm{\Delta_{r_ju}^2g}_{L^2(\mu)}^2 &=4[G_{\mu,\nu}(r_b)-G_{\mu,\nu}(2r_a)]\\ &\quad+\sum_{j=a}^b\mathcal D_{\mu,\nu}(r_j). \end{split} \label{eq:weighted-telescope} \end{equation} If $g$ is globally $L$-Lipschitz and $d\mu=\rho dx$ with $\rho\in W^{1,1}(D)$, then \begin{equation} \begin{gathered} |\mathcal D_{\mu,\nu}(r)|\le2L^2r\norm{\nabla\rho}_{L^1},\\ \sum_{j=a}^b|\mathcal D_{\mu,\nu}(r_j)| \le4L^2r_a\norm{\nabla\rho}_{L^1}. \end{gathered} \label{eq:measure-correction-bound} \end{equation} \end{proposition}

The correction measures how much mass translation places on locations with large first differences. The correction vanishes for the uniform torus measure, recovering \eqref{eq:directional-telescope}. For a differentiable density, \eqref{eq:measure-correction-bound} makes the correction's fine-scale contribution summable. Applied conditionally on the class label and then averaged, \eqref{eq:weighted-telescope} gives the finite-band decomposition of RSC under the training distribution. The proof is in App.~\ref{app:proof-weighted-telescope}.

\paragraph{The critical kink scaling under a data measure.} For piecewise-affine responses, the leading moment depends on data mass near the interfaces and on the directions that cross those interfaces. Write $\mathcal H^{d-1}$ for surface measure and $t_+=\max\{t,0\}$.

\begin{proposition}[Data- and direction-weighted kink intensity] \label{prop:weighted-kinks} Let $g:\R^d\to\R$ be continuous and affine on each cell of a finite polyhedral partition. Let $X$ have a continuous, compactly supported density $\rho$, let $u\sim\nu$ be independent of $X$, and let $0<c\le\xi\le C<\infty$ be independent of $(X,u)$. For each distinct codimension-one interface $F$, choose a unit normal $n_F$ and write the gradient jump as $a_F n_F$. Set \begin{equation} \begin{split} \mathcal I_{\rho,\nu}(g) &\coloneqq\sum_F a_F^2\E_\nu|u\cdot n_F|^3 \int_F\rho(x) d\mathcal H^{d-1}(x),\\ A(r)^2&\coloneqq\E|\Delta_{r\xi u}^2g(X)|^2. \end{split} \label{eq:weighted-interface-intensity} \end{equation} Then \begin{equation} A(r)^2=\frac23\E[\xi^3]\mathcal I_{\rho,\nu}(g)r^3+o(r^3) \qquad(r\to0). \label{eq:weighted-kink-asymptotic} \end{equation} For a single planar hinge $g(x)=\ell(x)+a(n\cdot x-b)_+$ with affine $\ell$ and $\norm n_2=1$, the same conclusion holds for any base law whose signed-distance variable $T=n\cdot X-b$ has a density $f_T$ continuous near zero, with $\mathcal I_{\rho,\nu}(g)$ replaced by $a^2 f_T(0)\E_\nu|u\cdot n|^3$. \end{proposition}

Continuity forces the gradient jump to be normal to each interface. The factor $|u\cdot n_F|^3$ combines the squared directional slope jump with the thickness of the crossing region. The single-hinge statement only requires a density for signed distance to the interface: A smooth density with compact support in a manifold chart transverse to that interface supplies this local condition when the tangential gradient of signed distance is bounded away from zero on the relevant support. Thus the $r^{3/2}$ amplitude law also applies to such concentrated data distributions. The proof is in App.~\ref{app:proof-weighted-kinks}.

If the coefficient in \eqref{eq:weighted-kink-asymptotic} is positive, $r^{-3/2}A(r)$ approaches the square root of that coefficient. For a fixed number $L$ of successive fine scales, \begin{equation} \lim_{a\to\infty}\frac1L\sum_{j=a}^{a+L-1} (r_j^{-3/2}A(r_j))^2 =\frac23\E[\xi^3]\mathcal I_{\rho,\nu}(g). \label{eq:critical-band-intensity} \end{equation} This limit identifies the $s=3/2$ readout as an interface-intensity profile. For the same probe law, the slope-normalized contribution is $r^{-2}A(r)^2=(2/3)\E[\xi^3]\mathcal I_{\rho,\nu}(g)r+o(r)$: The $s=1$ and $s=3/2$ normalizations weight the same boundary contribution differently across scales. At this critical exponent the fine-scale sequence is bounded, as in the $q=\infty$ Besov aggregation, while summing the sequence's nonzero limiting value over infinitely many scales diverges for finite $q$. The measured $q=2$ aggregate retains a specified finite band of this critical profile.

\paragraph{Boundary mass and the moment order.} The signed-distance formulation also identifies how the critical exponent depends on the local data law. The signed-distance formulation applies to ambient probes from a concentrated distribution through a one-dimensional marginal.

\begin{proposition}[Local mass and the critical exponent] \label{prop:boundary-mass-exponent} Let $g(x)=\ell(x)+a(n\cdot x-b)_+$, where $\ell$ is affine and $\norm n_2=1$. Suppose $T=n\cdot X-b$ has a density in a neighborhood of zero satisfying, for some $\kappa>0$ and $c_+,c_-\ge0$, \begin{equation} f_T(t)=c_+t^{\kappa-1}+o(t^{\kappa-1}),\qquad f_T(-t)=c_-t^{\kappa-1}+o(t^{\kappa-1}) \quad(t\to0^+). \label{eq:boundary-density-power} \end{equation} Let $u\sim\nu$ be a unit direction independent of $X$, and let $0<\xi_-\le\xi\le\xi_+<\infty$ be independent of $(X,u)$. For $1\le p<\infty$, define \begin{equation} \begin{split} C_{p,\kappa}&=\int_0^2\min\{t,2-t\}^{p}t^{\kappa-1} dt,\\ I_{p,\kappa} &=|a|^p C_{p,\kappa}\E[\xi^{p+\kappa}]\\ &\quad\times\E_\nu[ c_-(u\cdot n)_+^{p+\kappa} +c_+(-u\cdot n)_+^{p+\kappa}]. \end{split} \label{eq:boundary-mass-intensity} \end{equation} Then \begin{equation} A_p(r)^p\coloneqq\E|\Delta_{r\xi u}^2g(X)|^p =I_{p,\kappa}r^{p+\kappa}+o(r^{p+\kappa}). \label{eq:boundary-mass-asymptotic} \end{equation} If $I_{p,\kappa}>0$, the critical scale normalization is $s_\kappa=1+\kappa/p$. For $0<\kappa<p$, the critical exponent lies in $0<s_\kappa<2$: The fine-scale sequence $\{r_j^{-s}A_p(r_j)\}$ belongs to every $\ell^q$, $0<q\le\infty$, when $s<s_\kappa$; at $s=s_\kappa$ the sequence is bounded with positive limit and belongs to $\ell^\infty$ alone. \end{proposition}

\begin{proof} The affine term cancels. Put $v=u\cdot n$ and $H=r\xi|v|$. For $v>0$, the second difference is supported on $-2H<T<0$, where the absolute second difference is $|a|H\min\{-T/H,2+T/H\}$. Consequently the conditional $p$-th moment of the second difference is \begin{align*} |a|^pH^{p+1}\int_0^2\min\{t,2-t\}^p f_T(-Ht) dt. \end{align*} Dividing by $H^{p+\kappa}$ and using \eqref{eq:boundary-density-power} gives $|a|^pc_-C_{p,\kappa}$. For $v<0$, the crossing interval is $0<T<2H$ and the same calculation gives $c_+$; for $v=0$ the difference vanishes. The density asymptotics bound the integrands by a constant times $\min\{t,2-t\}^pt^{\kappa-1}$. Since $|v|\le1$ and $\xi$ is bounded, dominated convergence also applies to the remaining expectation over $(u,\xi)$, proving \eqref{eq:boundary-mass-asymptotic}. Finally, $r_j^{-s}A_p(r_j)\sim I_{p,\kappa}^{1/p}r_j^{s_\kappa-s}$, which gives the stated sequence properties. \end{proof}

For a positive continuous signed-distance density, $\kappa=1$ and $c_+=c_-=f_T(0)$. Since $C_{p,1}=2/(p+1)$, the proposition recovers the $s=3/2$ quadratic kink law, including the asymptotic constant, at $p=2$. The factors in \eqref{eq:boundary-mass-intensity} separate the slope jump, the local data mass, and the direction's normal component. The parameter $p$ determines the critical exponent $1+\kappa/p$, while $q$ determines aggregation at that exponent. Below the critical exponent, the hinge's omitted dyadic tail decays geometrically. At criticality, the mean of $(r_j^{-s_\kappa}A_p(r_j))^p$ over a fixed fine band converges to $I_{p,\kappa}$; \eqref{eq:critical-band-intensity} gives the quadratic $\kappa=1$ case.

The boundary-mass law also determines the number of independent base points needed to resolve the spectrum at a fine scale. The following consequence of Propositions~\ref{prop:boundary-mass-exponent} and~\ref{prop:finite-band-estimation} quantifies the relative error of the empirical squared amplitude, uniformly in the number of probes per base point.

\begin{corollary}[Statistical resolution of the roughness spectrum] \label{cor:statistical-resolution} Retain the hypotheses and notation of Proposition~\ref{prop:boundary-mass-exponent}, and suppose $I_{2,\kappa}>0$. Fix the laws of $X$, $u$, and $\xi$ independently of the probe radius. For integers $N,K\ge1$, let $X_1,\ldots,X_N$ be independent copies of $X$, and let $(\xi_{ik},u_{ik})_{i\le N,k\le K}$ be independent copies of $(\xi,u)$, independent of the base points. Define \begin{equation} \begin{aligned} E(r)&\coloneqq A_2(r)^2,\\ \widehat E_{N,K}(r)&\coloneqq \frac1{NK}\sum_{i=1}^N\sum_{k=1}^K \abs{\Delta^2_{r\xi_{ik}u_{ik}}g(X_i)}^2. \end{aligned} \label{eq:resolution-estimator} \end{equation} There exist $r_*>0$ and $0<c\le C<\infty$, depending only on the fixed hinge and sampling laws, such that $E(r)>0$ and \begin{equation} \frac{c}{Nr^\kappa} \le \E\left[\left(\frac{\widehat E_{N,K}(r)}{E(r)}-1\right)^2\right] \le \frac{C}{Nr^\kappa} \label{eq:resolution-relative-error} \end{equation} for every $0<r<r_*$ and every $N,K\ge1$. In particular, for any deterministic sequences $r_N>0$ and integers $K_N\ge1$ with $\lim_{N\to\infty}r_N=0$, \begin{align*} \lim_{N\to\infty} \E\left[\left(\frac{\widehat E_{N,K_N}(r_N)}{E(r_N)}-1\right)^2\right]=0 \quad\Longleftrightarrow\quad \lim_{N\to\infty}Nr_N^\kappa=\infty. \end{align*} \end{corollary}

\begin{proof} Put $W_r=\abs{\Delta^2_{r\xi u}g(X)}^2$ and $w_r(X)=\E[W_r\mid X]$. The affine part of $g$ cancels, and the hinge second difference satisfies $W_r\le |a|^2\xi_+^2r^2$, so all the following second moments are finite. Applying Proposition~\ref{prop:boundary-mass-exponent} at moment orders $2$ and $4$ gives \begin{align*} \lim_{r\to0^+}\frac{\E W_r}{r^{2+\kappa}}&=I_{2,\kappa}>0,\\ \lim_{r\to0^+}\frac{\E W_r^2}{r^{4+\kappa}}&=I_{4,\kappa}<\infty. \end{align*} Consequently, for all sufficiently small $r$, $E(r)=\E W_r>0$ and $\E W_r^2/E(r)^2\le C_0r^{-\kappa}$ for some $C_0<\infty$.

Conditional independence of the $K$ probes at each base point, followed by the law of total variance, gives the single-scale variance identity used in Proposition~\ref{prop:finite-band-estimation}: \begin{align*} \E\widehat E_{N,K}(r)&=E(r),\\ \operatorname{Var}(\widehat E_{N,K}(r)) &=\frac{\operatorname{Var}(w_r(X))}{N}+ \frac{\E[\operatorname{Var}(W_r\mid X)]}{NK}. \end{align*} Since $K\ge1$, another application of the law of total variance bounds the right-hand side by $\operatorname{Var}(W_r)/N\le\E W_r^2/N$. Unbiasedness and the preceding moment bound prove the upper bound in \eqref{eq:resolution-relative-error}.

For the lower bound, retain $T=n\cdot X-b$ from Proposition~\ref{prop:boundary-mass-exponent}. If $|T|>2\xi_+r$, then all three stencil points lie on the same side of the hinge for every admissible probe, because $|r\xi u\cdot n|\le\xi_+r$. Thus $w_r(X)=0$ outside $B_r\coloneqq\{|T|\le2\xi_+r\}$. By \eqref{eq:boundary-density-power}, there exist $M,t_0>0$ such that $f_T(t)+f_T(-t)\le Mt^{\kappa-1}$ for almost every $0<t<t_0$. Hence, whenever $2\xi_+r<t_0$, \begin{align*} p_r\coloneqq\Prob(B_r) &=\int_0^{2\xi_+r}(f_T(t)+f_T(-t))dt\\ &\le \frac{M(2\xi_+)^\kappa}{\kappa}r^\kappa \coloneqq C_1r^\kappa. \end{align*} Since $E(r)>0$, necessarily $p_r>0$. Cauchy--Schwarz and the support property of $w_r(X)$ imply \begin{align*} E(r)^2 &=(\E[w_r(X)\mathbf{1}_{B_r}])^2 \le p_r\E[w_r(X)^2]. \end{align*} Choose $r_*>0$ small enough that all preceding bounds hold and $C_1r_*^\kappa\le1/2$. For $0<r<r_*$, these bounds give \begin{align*} \frac{\operatorname{Var}(w_r(X))}{E(r)^2} &\ge \frac1{p_r}-1 \ge \frac1{C_1r^\kappa}-1 \ge \frac1{2C_1r^\kappa}. \end{align*} The second term in the clustered variance identity is nonnegative, so unbiasedness gives the lower bound in \eqref{eq:resolution-relative-error} with $c=(2C_1)^{-1}$. The equivalence for shrinking radii follows from the two-sided bound, whose constants are independent of $N$ and $K$. \end{proof}

Multiplying both $E(r)$ and $\widehat E_{N,K}(r)$ by $r^{-2s}$ preserves the estimator's relative error. At $s_\kappa=1+\kappa/2$, the squared spectrum $r^{-2s_\kappa}E(r)$ converges to $I_{2,\kappa}>0$, while the amplitude $r^{-s_\kappa}A_2(r)$ converges to $\sqrt{I_{2,\kappa}}$. Corollary~\ref{cor:statistical-resolution} gives relative MSE $\Theta(N^{-1}r^{-\kappa})$ for the empirical squared spectrum. Relative root-mean-square accuracy $\eta$ for the empirical squared spectrum therefore requires $N$ of order $\eta^{-2}r^{-\kappa}$, uniformly in $K$. For positive continuous signed-distance densities, $\kappa=1$, so halving the radius doubles the required number of base points up to constants. Additional directions cannot remove base-point variance. These conclusions concern the specified hinge-model estimator, not arbitrary estimators or network responses.

\subsection{Uniform Besov control under data-dependent sampling} \label{app:uniform-besov}

\begin{remark}[Data-dependent sampling and uniform control] \label{rem:notnorm} For finite $p$, the moment in \eqref{eq:spectrum} weights changes by the frequency of those changes under $\mu$, which may concentrate on natural images. For the quadratic slope energy, Proposition~\ref{prop:weighted-telescope} expresses this measure dependence through an exact translation correction to the Lebesgue identity; Proposition~\ref{prop:weighted-kinks} relates the fine-scale amplitudes to data mass near slope-change boundaries. More generally, a signed-distance density of order $|t|^{\kappa-1}$ gives the hinge amplitude law $A(r)\asymp r^{1+\kappa/p}$ under nondegenerate crossing directions (Proposition~\ref{prop:boundary-mass-exponent}). A uniform comparison takes a supremum over a fixed domain $\Omega$ containing the probe stencils: \begin{equation} A_j(g)\le U_j^{(m)}(g;\Omega)\coloneqq \sup_{\substack{\norm{h}_2\le\sqrt2r_j\\ z,z+h,\ldots,z+mh\in\Omega}} \norm{\Delta_h^m g(z)}. \label{eq:uniform-amplitude} \end{equation} For scalar continuous responses on a box or the torus, the full $s$-weighted $\ell^q$ aggregate of $U_j^{(m)}$, together with $\norm{g}_\infty$, is equivalent to the classical $B^s_{\infty,q}$ (quasi-)norm for $0<s<m$ (Proposition~\ref{prop:uniform-besov}). This comparison holds for every base-point and direction law, including empirical laws. The deployed moments measure the same finite-difference structure under a task-relevant distribution, while the uniform counterpart controls the finite-difference structure throughout $\Omega$. \end{remark}

The uniform amplitudes in \eqref{eq:uniform-amplitude} give a Besov comparison that does not require a density for the data distribution. This comparison uses the classical modulus of smoothness, taking the supremum over base points and displacements in a fixed domain.

\begin{proposition}[Uniform Besov characterization and moment control] \label{prop:uniform-besov} Let $\Omega$ be a compact box with nonempty interior in $\R^d$, or the torus $\mathbb T^d$, and let $g\in C(\Omega)$ be scalar-valued. Fix an integer $m\ge1$, $0<s<m$, $0<q\le\infty$, and $r_j=r_02^{-j}$ with $r_0>0$. Then \begin{equation} \norm{g}_{B^s_{\infty,q}(\Omega)} \asymp \norm{g}_{L^\infty(\Omega)}+ \norm{\{r_j^{-s}U_j^{(m)}(g;\Omega)\}_{j\ge0}}_{\ell^q}. \label{eq:uniform-besov} \end{equation} For each $j$, let $(z,h_j)$ have any joint law with $\norm{h_j}_2\le\sqrt2r_j$ and $z,z+h_j,\ldots,z+mh_j\in\Omega$ almost surely. For $1\le p<\infty$, form the moments $A_j(g)$ and the corresponding aggregate $R^{(m)}_{s,p,q}(g)$ over $J\subseteq\{0,1,\ldots\}$ as in \eqref{eq:spectrum}. Then \begin{equation} R^{(m)}_{s,p,q}(g) \le \norm{\{r_j^{-s}U_j^{(m)}(g;\Omega)\}_{j\in J}}_{\ell^q} \le C\norm{g}_{B^s_{\infty,q}(\Omega)}. \label{eq:uniform-moment-control} \end{equation} The constants depend only on $(\Omega,m,s,q,r_0)$, independently of $g$, $p$, and the probe laws. On a box, the Besov space is the usual restriction space; on the torus, the differences are periodic. \end{proposition}

The proof is in App.~\ref{app:proof-uniform-besov}. For a continuous response, $U_j^{(m)}$ also equals the essential supremum under any joint probe law with full support on the admissible stencil set in \eqref{eq:uniform-amplitude}: Every neighborhood of a maximizer has positive probability. This equality depends on support, not on the probability weights within that support. A law on a smaller support gives the corresponding restricted supremum, while \eqref{eq:uniform-moment-control} applies to all such laws.

\paragraph{Application to the RSC energy.} For each class $y$, define the continuous margin function $g_y(x)\coloneqq f_\theta(x)_y-\max_{k\ne y}f_\theta(x)_k$ on a box $\Omega$ containing all training probe stencils. Let $J_{\mathrm{train}}$ index the radii in $\mathcal R_{\mathrm{train}}=\{4,2,1,0.5\}$ and put $V_j\coloneqq\max_y U_j^{(2)}(g_y;\Omega)$. Applying \eqref{eq:uniform-amplitude} to each labeled example in \eqref{eq:rsc} gives \begin{equation} \mathcal L_{\mathrm{RSC}} \le \frac{1}{|J_{\mathrm{train}}|} \sum_{j\in J_{\mathrm{train}}}(r_j^{-1}V_j)^2 \le C\max_y\norm{g_y}_{B^1_{\infty,\infty}(\Omega)}^2. \label{eq:rsc-uniform-besov} \end{equation} The last inequality uses that an average of squares is bounded by the largest square and applies \eqref{eq:uniform-besov} at $m=2$, $s=1$, $q=\infty$. The bound holds for any training-data distribution and the sampled nuisance directions. This bound relates the finite-band RSC energy to uniform Besov regularity while the penalty's $p=2$ moment weights the slope changes encountered on the data. The uniform $B^1_{\infty,\infty}$ control is finite for Lipschitz margins, including piecewise-affine ones.

\subsection{Dyadic reconstruction and prediction-churn control} \label{app:theory-dyadic}

The vision diagnostic is deliberately second-order, and second-order probing still controls a first-order failure. Proposition~\ref{prop:margin} bounds scale-$r$ prediction churn by the clean margin distribution plus the scale-$r$ first-order margin roughness, and the exact dyadic identity of Lemma~\ref{lem:dyadic-reconstruct} converts the measured second-order grid, together with one coarse first-order anchor that is necessary rather than proof slack, into population control of that first-order quantity and hence into a prediction-churn bound (Theorem~\ref{thm:dyadic-churn} and Corollary~\ref{cor:dyadic-churn}), neither step a smoothness approximation.

\paragraph{Margin intuition.} For a cat image with logits $8$, $5$, and $1$ for the classes cat, dog, and car, the margin is $8-5=3$. A positive margin means the true class wins; a negative margin means the true class loses. For a correctly classified input, a label-preserving perturbation retains the correct prediction whenever $\abs{m_\theta(x+h)-m_\theta(x)}<m_\theta(x)$: The change cannot close the initial gap. A larger positive margin tolerates a larger decrease before a rival can win. Robustness to an input perturbation depends on both this initial margin and the perturbation-induced change in the margin (Proposition~\ref{prop:margin}).

\begin{proposition}[Margin-roughness robustness bound] \label{prop:margin} For any $\gamma>0$, any $p>0$, and any label-preserving perturbation distribution $h$, \begin{equation} \Prob_{x,h}\{\hat y(x+h) \ne \hat y(x)\} \le \Prob_x\{m_\theta(x)<\gamma\} + \frac{\E_{x,h}\abs{\Delta_h m_\theta(x)}^p}{\gamma^p}, \end{equation} \textit{i.e.}, scale-$r$ prediction churn is controlled by the clean margin distribution plus the scale-$r$ first-order margin roughness. \end{proposition}

\begin{lemma}[Dyadic finite-difference reconstruction] \label{lem:dyadic-reconstruct} Let $J\ge1$ and let $g:\mathcal Z\to\mathcal Y$ take values in a vector space. For any $z,h$ such that the segment from $z$ to $z+2^Jh$ lies in $\mathcal Z$, \begin{equation} \Delta_h^1 g(z) =2^{-J}\Delta_{2^Jh}^1 g(z) -\sum_{a=0}^{J-1}2^{-(a+1)}\Delta_{2^ah}^2 g(z). \label{eq:dyadic-reconstruct} \end{equation} \end{lemma}

For the amplitude statement, fix $1\le p<\infty$, $J\ge1$, and a map $g:\mathcal Z\to\mathcal Y$ into a normed vector space. Consider a scale-coherent probe law: Draw $(z,\xi,u)$ once, put $h\coloneqq r\xi u$, and obtain the nominal scale $2^a r$ by replacing $h$ with $2^a h$, for $a=0,\dots,J$. Assume the segment from $z$ to $z+2^Jh$ lies in $\mathcal Z$ almost surely. Define the coupled population amplitudes by \begin{equation} A^{(m)}(2^a r) \coloneqq(\E\norm{\Delta^m_{2^a h}g(z)}^p)^{1/p}. \end{equation}

\begin{theorem}[Second-to-first-order dyadic amplitude control] \label{thm:dyadic-churn} Under this scale-coherent setup, Lemma~\ref{lem:dyadic-reconstruct} implies \begin{equation} A^{(1)}(r) \le 2^{-J}A^{(1)}(2^Jr) +\sum_{a=0}^{J-1}2^{-(a+1)}A^{(2)}(2^ar) \eqqcolon U_J(r). \label{eq:dyadic-a1} \end{equation} \end{theorem}

\begin{corollary}[Second-order prediction-churn control] \label{cor:dyadic-churn} Under the hypotheses of Theorem~\ref{thm:dyadic-churn}, if $g=m_\theta$ is the classification margin and the scale-$r$ perturbation $h$ is label-preserving, then for every $\gamma>0$, \begin{equation} \Prob\{\hat y(x+h)\ne\hat y(x)\} \le \Prob\{m_\theta(x)<\gamma\} +\frac{U_J(r)^p}{\gamma^p}. \label{eq:dyadic-churn} \end{equation} \end{corollary}

On the paper's dyadic grid, taking $r=r_j$ and $J\le j$ uses exactly the measured scales $2^ar_j=r_{j-a}$; independent Monte Carlo draws at different scales are allowed because only the population marginals at each scale enter~\eqref{eq:dyadic-a1}.

The coarse first-order anchor in Theorem~\ref{thm:dyadic-churn} measures the overall margin change at a larger radius. To see why RSC also needs the coarse first-order anchor for a robustness bound, consider a linearly decreasing margin $m_\theta(x+tu)=a-bt$ with $a,b>0$. This linear margin has zero second differences, so the RSC penalty does not detect this decline, yet the margin becomes negative when $t>a/b$. The anchor must therefore be measured or controlled separately from the second-order spectrum. Together with the second differences, the coarse first-order anchor bounds the margin change at finer scales; the clean-margin distribution then enters the prediction-stability bound in Corollary~\ref{cor:dyadic-churn}.

\subsection{Finite-scale damage, task alignment, and target-mismatch control} \label{sec:damage-theory} \label{app:theory-alignment}

This subsection develops the theory behind the allocation rules of Sec.~\ref{sec:control}. The quantization score of \eqref{eq:dquant} is the finite-scale roughness of the task-loss functional, and this identification makes that section's allocation rule an object we can reason about exactly, the one place where the comparison between the finite-scale reading and the scalar rests on a provable identity rather than only a measured correlation. This subsection sets up the configuration-level damage $\mathcal{D}(S)$ of a mixed-precision assignment $S$, the additive damage surrogate $\widehat{\mathcal{D}}(S)\coloneqq\sum_{\ell\in S}D_\ell$, and the configuration criterion $\mathcal C_{\mathrm{SKL}}$ that the drift-greedy rule re-measures, and then proves the alignment results; App.~\ref{app:theory-alloc} takes up additivity and allocation. The main proofs are collected in App.~\ref{app:proofs}.

\paragraph{Drift is not damage.} Write $D_\ell$ for the task damage of quantizing block $\ell$ and $G_\ell$ for the label-free forward-KL self-damage. Theorem~\ref{thm:drift-damage} makes the missing orientation term exact: The gap $D_\ell-G_\ell$ is the target mismatch paired with the perturbation's log-ratio direction. Corollary~\ref{cor:drift-damage-bound} turns the identity into the dimension-free Cauchy--Schwarz control $\abs{D_\ell-G_\ell}\le C_{\mathrm{tgt}}V_\ell$, a target-mismatch factor times a log-ratio-variance factor, and Proposition~\ref{prop:alignment-inversion} exhibits a three-outcome instance on which forward KL and the symmetric drift of \eqref{eq:bquant} alike order two quantized blocks incorrectly, so symmetrization does not remove the obstruction. Thus a large, precisely estimated output movement can be less damaging than a smaller movement because the larger movement points away from the task direction, and only the target-labeled finite difference measures task damage $D_\ell$, the signed first finite difference of $\E_x\mathrm{NLL}_\theta(x)$ along the quantization-noise direction, a task-aligned companion to Definition~\ref{def:spectrum} rather than a Besov (quasi-)seminorm.

\paragraph{The setup and the alignment results.} Fix a calibration distribution over inputs $x$ and a target low precision with quantizer $Q$. A mixed-precision configuration is a set $S\subseteq \{1,\dots,L\}$ of blocks placed at the low precision; block $\ell\in S$ contributes the weight residual $\delta_\ell \coloneqq Q(\theta_\ell)-\theta_\ell$, supported on block $\ell$'s coordinates, and the configuration realizes the stacked perturbation $\delta_S \coloneqq \sum_{\ell\in S}\delta_\ell$. With \begin{equation} \Delta(\delta) \coloneqq \E_x[\mathrm{NLL}_{\theta+\delta}(x) - \mathrm{NLL}_\theta(x)], \quad \mathcal{D}(S) \coloneqq \Delta(\delta_S), \quad D_\ell \coloneqq \Delta(\delta_\ell), \label{eq:damage-def} \end{equation} the damage of a perturbation, of a configuration, and of block $\ell$ alone (\eqref{eq:dquant}), the additive damage rule is $\widehat{\mathcal{D}}(S) \coloneqq \sum_{\ell\in S}D_\ell$, and damage-greedy re-evaluates $\mathcal{D}$ at each realized $S$. For the implemented drift-greedy rule, if $P_x\coloneqq p_\theta(\cdot\mid x)$ and $Q_{S,x}\coloneqq p_{\theta+\delta_S}(\cdot\mid x)$, the configuration criterion is $\mathcal C_{\mathrm{SKL}}(S)\coloneqq\E_x\dskl(P_x,Q_{S,x})^2$. Thus $B^{\mathrm{quant}}_{\ell,b} =\mathcal C_{\mathrm{SKL}}(\{\ell\})^{1/2}$ at the corresponding bit width, but the drift-greedy rule re-measures $\mathcal C_{\mathrm{SKL}}$ directly. Taking a square root preserves the ranking of individual blocks, but changes the numerical size of score reductions. For the fixed-$K$ greedy procedure, taking a square root also preserves which candidate upgrade is chosen: All candidates share the same current configuration, so minimizing either $\mathcal C_{\mathrm{SKL}}$ or $\sqrt{\mathcal C_{\mathrm{SKL}}}$ selects the same candidate. We state the objective explicitly to identify the quantity that the implementation re-measures. The results below, proved in App.~\ref{app:proofs}, identify the exact gap between the label-free score and the task damage; App.~\ref{app:theory-alloc} then gives a finite-path near-oracle bound for the additive rule and a worst-case impossibility result for scores restricted to infinitesimal curvature.

The drift--damage distinction is not merely terminological.  Let $q_x$ denote the target distribution at context $x$, a point mass at the observed token in the corpus experiment, let $P_x\coloneqq p_\theta(\cdot\mid x)$ be the full-precision output, and $Q_{\ell,x}\coloneqq p_{\theta+\delta_\ell}(\cdot\mid x)$, the output after quantizing block $\ell$. Define the task damage and the label-free forward-KL self-damage by \begin{equation} D_\ell\coloneqq\E_x\sum_y q_x(y)\log\frac{P_x(y)}{Q_{\ell,x}(y)}, \qquad G_\ell\coloneqq\E_x D_{\mathrm{KL}}(P_x\Vert Q_{\ell,x}). \label{eq:alignment-def} \end{equation} The following result makes the missing orientation term exact.

\begin{theorem}[Task-alignment identity for finite-scale drift and damage] \label{thm:drift-damage} Suppose the output alphabet is finite and $P_x(y),Q_{\ell,x}(y)>0$ for every output $y$, almost surely in $x$; $q_x$ may have smaller support, and assume the expectations in~\eqref{eq:alignment-def} and below are finite. Then \begin{equation} D_\ell-G_\ell =\E_x\sum_y(q_x(y)-P_x(y)) \log\frac{P_x(y)}{Q_{\ell,x}(y)}. \label{eq:alignment-id} \end{equation} \end{theorem}

$D_\ell$ is the signed first finite difference of $g(\theta)\coloneqq\E_x\mathrm{NLL}_\theta(x)$ along the quantization-noise direction $\delta_\ell$. The damage score is a task-aligned companion to Definition~\ref{def:spectrum}, not a Besov (quasi-)seminorm: The score averages a signed loss change at one realized scale rather than an absolute $p$-moment over a scale grid.

Two consequences of Theorem~\ref{thm:drift-damage} fix the operating range of the label-free score. To express the mismatch control, define \begin{equation} C_{\mathrm{tgt}}^2\coloneqq\E_x\chi^2(q_x\Vert P_x), \qquad V_\ell^2\coloneqq\E_x\operatorname{Var}_{Y\sim P_x} \left[\log\frac{P_x(Y)}{Q_{\ell,x}(Y)}\right], \label{eq:alignment-cv} \end{equation} where $\chi^2(q\Vert P)\coloneqq\sum_y(q(y)-P(y))^2/P(y)$.

\begin{corollary}[Target-mismatch control] \label{cor:drift-damage-bound} Under the conditions and notation of Theorem~\ref{thm:drift-damage}, the dimension-free Cauchy--Schwarz control is \begin{equation} \abs{D_\ell-G_\ell}\le C_{\mathrm{tgt}}V_\ell. \label{eq:alignment-bound} \end{equation} \end{corollary}

For the corpus point mass $q_x=\delta_{y_x}$, $\chi^2(q_x\Vert P_x)=P_x(y_x)^{-1}-1$: Low confidence in the target token amplifies exactly the gap in~\eqref{eq:alignment-bound}. This amplification also names the limit of label-free self-damage: Only the target-labeled finite difference measures task damage $D_\ell$ directly. Theorem~\ref{thm:drift-damage} and Corollary~\ref{cor:drift-damage-bound} are about the forward-KL member of the drift family; the proposition below shows that symmetrization does not remove the obstruction.

\begin{proposition}[Target mismatch can reverse drift and damage rankings] \label{prop:alignment-inversion} Without control of the target mismatch, there is no monotone comparison, even when both task damages are positive.  For one context, \begin{equation} q\coloneqq(1,0,0),\quad P\coloneqq(0.6,0.2,0.2),\quad Q_A\coloneqq(0.59,0.405,0.005),\quad Q_B\coloneqq(0.5,0.25,0.25) \label{eq:alignment-inversion} \end{equation} give $G_A\approx0.607>0.020\approx G_B$ but $D_A=\log(60/59)\approx0.0168<0.1823\approx\log(6/5)=D_B$. The symmetric drift of \eqref{eq:bquant} orders the same pair incorrectly: $\dskl(P,Q_A)^2\approx0.432>0.020\approx\dskl(P,Q_B)^2$. \end{proposition}

This instance realizes the claim stated at the head of this subsection, and localizes that claim: The obstruction is orientation, not estimation, so symmetrizing the divergence leaves the ordering just as wrong.

\subsection{Additivity, interaction, and near-oracle allocation} \label{app:theory-alloc}

This subsection gives an upper bound for the additive precision rule and a lower bound for every rule that reads a block only through the block's infinitesimal jet, using the notation of App.~\ref{sec:damage-theory}.

\paragraph{The additive rule from above.} The results below show that the leading defect of the additive surrogate is cross-block curvature, and give the near-oracle bound $\mathcal D(\widehat S)-\mathcal D(S^\star)\le 2\Lambda_k+2k\varepsilon$ under a finite-path mixed-curvature envelope $\Lambda_k$ and calibration error $\varepsilon$, together with the transfer of the near-oracle bound to label-free selection (Prop.~\ref{prop:damage}, Lem.~\ref{lem:alloc-modularity}, Thm.~\ref{thm:alloc-stability}); greedy re-measurement invokes none of that surrogate, and we claim no submodular guarantee for the greedy rule. Against that positive result, Theorem~\ref{thm:alloc-regret} establishes a limitation of deterministic rules restricted to infinitesimal block information even in an exactly additive regime: On an explicit class of exactly additive instances whose blocks share the same second-order jet, every deterministic rule reading an instance only through those jets, including HAWQ and the earlier Optimal Brain Damage approach, incurs minimax regret exactly $\min(K,L-K)M$, unbounded as $M\to\infty$. The conclusion is a task-level separation achieved by measuring the realized finite-scale damage. Where cross-block curvature is non-negligible, the interaction-aware greedy of Sec.~\ref{sec:rsc} removes the additive modeling step by re-measuring $\mathcal{D}$ along the greedy selection path, and we report that greedy rule alongside the additive rule for that reason.

\begin{proposition}[The leading additive defect is cross-block curvature] \label{prop:damage} Let $\Delta$ be $C^3$ on a neighborhood of the box $\mathcal{B}_S \coloneqq \{\sum_{\ell\in S}t_\ell \delta_\ell : t\in[0,1]^{|S|}\}$, let $H \coloneqq \nabla^2\Delta(0)$, and put $M_3\coloneqq\sup_{z\in\mathcal B_S}\norm{\nabla^3\Delta(z)}_{\mathrm{op}}$. The additive surrogate has the expansion \begin{equation} \mathcal{D}(S) - \sum_{\ell\in S}D_\ell = \underbrace{\sum_{\{\ell,\ell'\}\subseteq S}\delta_\ell^\top H\delta_{\ell'}} _{\text{cross-block curvature}} +\mathcal{R}, \qquad \abs{\mathcal{R}} \le \tfrac{M_3}{6}(\norm{\delta_S}^3 + \textstyle\sum_{\ell\in S}\norm{\delta_\ell}^3). \label{eq:damage-defect} \end{equation} \end{proposition}

The linear terms cancel identically, so the leading defect is the off-support, cross-block Hessian coupling.

Greedy re-measurement does not invoke this additive surrogate. Nor does greedy re-measurement invoke a submodular one. Selecting $k$ blocks to minimize $\mathcal D$ is a cardinality-constrained set-function problem, but the classical greedy guarantee for such problems \citep{nemhauser1978analysis} requires monotone submodularity, and \eqref{eq:damage-defect} shows the defect of the modular model to be a cross-block curvature form of either sign, and the interaction audits at the end of this subsection find that defect super-additive in the curvature-benign regime, so diminishing returns cannot be assumed. We therefore make no $1-1/e$ claim for the greedy rule and report that rule purely empirically; the guarantee we do prove (Theorem~\ref{thm:alloc-stability}) is for the additive rule, and that guarantee replaces submodularity by the uniform approximate-modularity envelope $\Lambda_k$ below.

Proposition~\ref{prop:damage} identifies the leading interaction term locally. The next lemma and theorem turn the same mechanism into an end-to-end allocation bound without a third-derivative expansion and with the calibration error included. That bound is the positive counterpart to the curvature-only impossibility result that follows.

Fix $L$ blocks and a high-precision budget $K\in\{1,\dots,L-1\}$, and put $k\coloneqq L-K$. Let $\mathcal B\coloneqq\{\sum_{\ell=1}^L t_\ell\delta_\ell:t\in[0,1]^L\}$ and suppose $\Delta$ is $C^2$ on a neighborhood of $\mathcal B$, with $\Delta(0)=0$. For $\ell\ne\ell'$, define the finite-path mixed-curvature envelope \begin{equation} I_{\ell\ell'}\coloneqq \sup_{t\in[0,1]^L} \abs{\delta_\ell^\top \nabla^2\Delta(t_1\delta_1+\dots+t_L\delta_L) \delta_{\ell'}}, \qquad \Lambda_k\coloneqq\max_{|S|=k}\sum_{\{\ell,\ell'\}\subseteq S}I_{\ell\ell'}. \label{eq:interaction-envelope} \end{equation}

\begin{lemma}[Finite-path approximate modularity] \label{lem:alloc-modularity} Under the preceding conditions and notation, every $k$-block low-precision configuration satisfies the uniform approximate-modularity bound \begin{equation} \abs*{\mathcal D(S)-\sum_{\ell\in S}D_\ell} \le\sum_{\{\ell,\ell'\}\subseteq S}I_{\ell\ell'} \le\Lambda_k. \label{eq:uniform-modularity} \end{equation} \end{lemma}

\begin{theorem}[Near-oracle finite-scale allocation under interaction and calibration error] \label{thm:alloc-stability} In the setting of Lemma~\ref{lem:alloc-modularity}, let $\widehat D_\ell$ be calibration estimates satisfying $\max_\ell\abs{\widehat D_\ell-D_\ell}\le\varepsilon$, let $\widehat S\in\arg\min_{|S|=k}\sum_{\ell\in S}\widehat D_\ell$, equivalently keeping the $K$ largest estimated finite-scale damages at high precision, and let $S^\star\in\arg\min_{|S|=k}\mathcal D(S)$. Then \begin{equation} 0\le\mathcal D(\widehat S)-\mathcal D(S^\star) \le 2\Lambda_k+2k\varepsilon. \label{eq:alloc-stability} \end{equation} \end{theorem}

Thus the realized finite-scale selector is exactly optimal when interactions and estimation error vanish, and degrades continuously, rather than catastrophically, when both are small.

Lemma~\ref{lem:alloc-modularity} and Theorem~\ref{thm:alloc-stability} together name the boundary of the additive rule. The relevant quantity is not a Hessian trace within one block but the sum of mixed directional curvatures along the entire quantization box. When that envelope is large, the guarantee correctly becomes loose. Theorem~\ref{thm:alloc-stability} also transfers Corollary~\ref{cor:drift-damage-bound} from per-block alignment to allocation: If $\widehat G_\ell$ estimates $G_\ell$ with $\max_\ell\abs{\widehat G_\ell-G_\ell}\le\eta$ and $\kappa\coloneqq\max_\ell C_{\mathrm{tgt}}V_\ell$, then selecting by the label-free self-damage gives, by~\eqref{eq:alignment-bound} and Theorem~\ref{thm:alloc-stability}, \begin{equation} \mathcal D(\widehat S_G)-\mathcal D(S^\star) \le 2\Lambda_k+2k(\kappa+\eta). \label{eq:alignment-allocation} \end{equation} The bound says precisely when drift allocation is licensed; the inversion in \eqref{eq:alignment-inversion} shows why no such license is distribution-free.

\paragraph{Infinitesimal curvature is minimax-insufficient.} Fix $L$ blocks, a high-precision budget $K\in\{1,\dots,L-1\}$, and constants $c>0$, $M>0$. Let $\mathcal{P}_{L,c,M}$ be the class of allocation instances in which the block damages are additive, $\mathcal{D}(S)=\sum_{\ell\in S}D_\ell$, an exactly separable, no-interaction subclass of the regime analyzed by Proposition~\ref{prop:damage}, and each block's low-precision loss profile $\phi_\ell(t)\coloneqq\Delta(t\delta_\ell)\in C^3([0,1])$ has second-order jet $(\phi_\ell(0),\phi_\ell'(0),\phi_\ell''(0))=(0,0,c)$ and realized damage $D_\ell=\phi_\ell(1)\in[c/2,c/2+M]$. For a low-precision set $S$ with $|S|=L-K$, write the regret $\mathrm{Reg}(S)\coloneqq\mathcal{D}(S)-\min_{|S'|=L-K}\mathcal{D}(S')$.

\begin{theorem}[Infinitesimal curvature is minimax-insufficient for allocation] \label{thm:alloc-regret} On the class $\mathcal{P}_{L,c,M}$ defined above, let $\mathcal A$ be any deterministic allocation rule that outputs a low-precision set of cardinality $L-K$ and reads an instance only through the block jets $\{(\phi_\ell(0),\phi_\ell'(0),\phi_\ell''(0))\}_{\ell=1}^{L}$. Then \begin{equation} \sup_{\mathcal{I}\in\mathcal{P}_{L,c,M}} \mathrm{Reg}_{\mathcal{I}}(\mathcal A(\mathcal{I})) =\min(K,L-K)M, \label{eq:alloc-regret} \end{equation} which is unbounded as $M\to\infty$. \end{theorem}

\noindent The proof is a one-line packing argument (App.~\ref{app:proofs}), and the conclusion is a task-level separation achieved by measuring the realized finite-scale damage: On this class, no deterministic function of the $r\to0$ curvature can allocate precision within $\min(K,L-K)M$ of optimal. The class is deliberately minimal, establishing the existence of instances on which infinitesimal scoring provably fails without asserting that measured models satisfy the identical-jet hypothesis. The class fixes an exactly separable regime where the additive rule is exact; when cross-block curvature is non-negligible, the interaction-aware greedy of Sec.~\ref{sec:rsc} removes the additive modeling step by re-measuring $\mathcal{D}$ along the greedy selection path, and the empirical comparisons establish the greedy rule's performance.

\paragraph{The evaluated configurations expose cross-block interaction.} If every block outside $S$ remains at full precision, the defect in \eqref{eq:damage-defect} is exactly the gap between joint damage and the sum of the single-block damages. The evaluated mixed configurations instead use an int8 complement, so the observed configuration gap also contains the complement's own damage and the complement's interactions with the int4 blocks. The int8-only damage is near-lossless, at most $0.003$ nats across the pool and an order of magnitude below the gap, which makes the gap a useful empirical interaction diagnostic, but the small int8-only damage alone does not bound those mixed interactions. This proxy is small and super-additive in the curvature-benign regime, at $+0.024$ nats for Pythia-410M $K=L/2$, and the additive ranking of configurations is preserved there at Spearman $0.88$. The one place the additive ranking is not preserved is where the interaction proxy is largest, at Pythia-2.8B $K=L/4$, where Spearman falls to $0.47$, an additivity break that greedy is designed to absorb, documented empirically in Sec.~\ref{sec:exp-quant}.

\section{Proofs} \label{app:proofs}

\subsection{Theorem~\ref{thm:besov-equiv} on ambient equivalence} \label{app:proof-besov} Throughout, $\lesssim$ hides constants depending only on $(d,m,s,p,q,c_\mu,C_\mu,r_0)$. Let $\sigma$ be the surface measure on $S^{d-1}$, normalized to unit mass, and for $t,r>0$ write the spherical mean and the normalized ball mean \begin{align*} M(t)^p &\coloneqq \int_{\mathbb{T}^d}\int_{S^{d-1}} \abs{\Delta^m_{tu}g(z)}^pd\sigma(u)dz,\\ V(r)^p &\coloneqq \abs{B_r}^{-1}\int_{\abs{h}\le r} \norm{\Delta^m_h g}^p_{L^p(\mathbb{T}^d)}dh = d r^{-d}\int_0^r t^{d-1}M(t)^pdt. \end{align*} For $r_0\le 1/(4m)$, every probed segment has length $\le m\sqrt2r_0<1/2$, so all differences are the standard periodic ones.

Exchanging the base measure comes first. By (A1), for every $j$, $c_\mu\E_\xi M(r_j\xi)^p \le A_j^p \le C_\mu\E_\xi M(r_j\xi)^p$.

The jitter law then becomes an annulus mean. The jitter $\xi$ has density $1/(\xi\ln 2)$ on $[2^{-1/2},2^{1/2}]$, so with $t=r_j\xi$, $\E_\xi M(r_j\xi)^p = \int_{\mathrm{ann}_j} M(t)^pdt/(t\ln 2)$, $\mathrm{ann}_j \coloneqq [r_j/\sqrt2,r_j\sqrt2]$. Since $t\asymp r_j$ on $\mathrm{ann}_j$ with absolute constants, $\E_\xi M(r_j\xi)^p \asymp N_j^p \coloneqq r_j^{-1}\int_{\mathrm{ann}_j}M(t)^pdt$.

Next we pass between annulus and ball means at the level of the aggregate, in both directions. To bound the annulus aggregate by the ball aggregate, note that $t^{d-1}\ge (r_j/\sqrt2)^{d-1}$ on $\mathrm{ann}_j$, so $N_j^p \lesssim r_j^{-d}\int_{\mathrm{ann}_j}t^{d-1}M(t)^pdt \lesssim V(\sqrt2r_j)^p \lesssim V(2r_j)^p$, using that $r\mapsto r^dV(r)^p$ is nondecreasing. Summing and bounding the one stray scale by $V(2r_0)\le 2^m\norm{g}_{L^p}$, \begin{align*} \left(\sum_{j\ge0}(r_j^{-s}N_j)^q\right)^{1/q} &\lesssim \left(\sum_{j\ge0}(r_j^{-s}V(r_j))^q\right)^{1/q} + \norm{g}_{L^p} \qquad\text{with suprema when }q=\infty. \end{align*} For the reverse comparison, decompose $[0,r_j]$ into the dyadic intervals $[r_{k+1},r_k]$, $k=j+i$. Each $[r_k/2,r_k]$ is covered by $\mathrm{ann}_{k+1}\cup\mathrm{ann}_k$ and carries $t^{d-1}\le r_k^{d-1}$, so $\int_{r_{k+1}}^{r_k}t^{d-1}M(t)^pdt \lesssim r_k^{d} (N_k^p+N_{k+1}^p)$ and hence $V(r_j)^p \lesssim \sum_{i\ge0}2^{-id}N_{j+i}^p$. With $b_k \coloneqq r_k^{-s}N_k$ and the summable kernel $\kappa_i \coloneqq 2^{-i(d+sp)}$, the bound reads $(r_j^{-s}V(r_j))^p \lesssim \sum_i \kappa_i b_{j+i}^p$; Minkowski's inequality in $\ell^{q/p}$ when $q\ge p$, the subadditivity of $x\mapsto x^{q/p}$ when $q<p$, or taking suprema when $q=\infty$ then gives $(\sum_j (r_j^{-s}V(r_j))^q)^{1/q} \lesssim (\sum_k b_k^q)^{1/q}$.

The classical characterization closes the chain. For $0<s<m$, $1\le p<\infty$, and $0<q\le\infty$, \begin{align*} \norm{g}_{B^s_{p,q}(\mathbb{T}^d)} &\asymp \norm{g}_{L^p} + \left(\sum_{j\ge0}(r_j^{-s}V(r_j))^q\right)^{1/q}. \end{align*} This equivalence is the means-of-differences characterization \citep[\S2.5.12]{triebel1983theory}, in the periodic setting \citep[Ch.~3]{schmeisser1987topics}, discretized to the dyadic grid $r_j\coloneqq r_02^{-j}$ by the standard dyadic discretization for averaged moduli of smoothness. The finitely many coarser scales in $[r_0,1]$ are absorbed into $\norm{g}_{L^p}$. The discretization uses the finite-difference dilation inequality for averaged moduli; mere monotonicity of the unnormalized ball integral gives only one direction and is not being used as a two-sided doubling estimate.

Chaining these four stages yields \eqref{eq:ambient}. \hfill$\square$

\subsection{Corollary~\ref{cor:mixed-besov} on mixed direction families} \label{app:proof-mixed-besov} Write $A_{j,\nu}$ and $A_{j,\mathrm{iso}}$ for the mixed and isotropic amplitudes under the same base measure and jitter. The mixture gives $A_{j,\nu}^p\ge\alpha_*A_{j,\mathrm{iso}}^p$. Hence $R_\nu\ge\alpha_*^{1/p}R_{\mathrm{iso}}$, and Theorem~\ref{thm:besov-equiv} supplies the Besov bound by $\norm{g}_p+R_\nu$. For the other direction, independence of the base point and direction, the density upper bound, and $\norm{r_j\xi u}\le\sqrt2r_j$ give \begin{align*} A_{j,\nu}\le C_\mu^{1/p}\omega_m(g,\sqrt2r_j)_p,\qquad \omega_m(g,t)_p\coloneqq\sup_{\norm{h}\le t}\norm{\Delta_h^m g}_{L^p}. \end{align*} The classical dyadic modulus characterization of $B^s_{p,q}$ \citep{triebel1983theory,devore1993constructive,schmeisser1987topics} bounds the $s$-weighted aggregate of the right-hand side by $C\norm{g}_{B^s_{p,q}}$. Both comparisons hold for $0<q<1$ as well as $q\ge1$, using monotonicity of the $\ell^q$ quasi-norm. \hfill$\square$

\subsection{Proposition~\ref{prop:angular-besov} on angular Besov energy} \label{app:proof-angular-besov} For $0<s<2$, $r_j=r_02^{-j}$, and $t\in\R$, the scalar dyadic multiplier satisfies \begin{align*} \sum_{j\in\mathbb Z}r_j^{-2s}|e^{ir_jt}-1|^4\asymp|t|^{2s}. \end{align*} For the upper bound use $|e^{iv}-1|\le\min\{2,|v|\}$ and split the sum at $r_j|t|=1$; the two geometric series converge because $0<s<2$. For the lower bound, when $t\ne0$ choose a dyadic radius with $r_j|t|\in[1/2,1]$ and retain that term. Plancherel's identity and Tonelli's theorem now prove \eqref{eq:angular-fourier-energy}, including infinite energies. The same argument applies to Fourier series on the torus.

For a unit vector $v$, put $Z=|v\cdot u|^2\in[0,1]$ and $\lambda_* = \lambda_{\min}(M_\nu)>0$. If $0<s\le1$, then $\E Z^s\ge\E Z\ge\lambda_*$; if $1\le s<2$, Jensen's inequality gives $\E Z^s\ge(\E Z)^s\ge\lambda_*^s$. The upper bound is $\E Z^s\le1$. Homogeneity therefore gives $\Psi_{\nu,s}(\omega)\asymp|\omega|^{2s}$. The coarser terms $j<0$ obey \begin{align*} \sum_{j<0}r_j^{-2s}\E_\nu\norm{\Delta^2_{r_ju}g}_2^2 \le16\norm{g}_2^2\sum_{j<0}r_j^{-2s} \lesssim r_0^{-2s}\norm{g}_2^2. \end{align*} Adding the $L^2$ term and using $H^s=B^s_{2,2}$ gives \eqref{eq:angular-besov-norm}. Rotational invariance within $E$ gives the formula for $\Psi_{\sigma_E,s}$. Fubini's theorem, the $B^s_{2,2}(E)$ Fourier characterization, and Plancherel in $E^\perp$ identify the corresponding norm as $L^2(E^\perp;B^s_{2,2}(E))$. For a fixed jitter value $\xi$, applying the scalar estimate to the grid $\xi r_j$ multiplies the comparison by $\xi^{2s}$. Averaging bounded positive jitter preserves the equivalence; (A1) compares each squared moment to the corresponding Lebesgue moment. \hfill$\square$

\subsection{Proposition~\ref{prop:directional-energy} on unjittered slope energy} \label{app:proof-directional-energy} Translation invariance of Lebesgue measure gives \begin{align*} \norm{\Delta_h^2g}_2^2 =4\norm{\Delta_hg}_2^2-\norm{\Delta_{2h}g}_2^2. \end{align*} Indeed, apply the parallelogram identity to $a=g(\cdot+2h)-g(\cdot+h)$ and $b=g(\cdot+h)-g$; translation invariance gives $\norm{a}_2=\norm{b}_2$. After averaging over $u$ and dividing by $r_j^2$, the identity becomes \begin{align*} r_j^{-2}\E_\nu\norm{\Delta_{r_ju}^2g}_2^2 =4[G_\nu(r_j)-G_\nu(2r_j)]. \end{align*} Since $2r_j=r_{j-1}$, summing proves \eqref{eq:directional-telescope}. For $g\in H^1(D)$, convergence of first difference quotients in $L^2$ and the gradient-norm bound $\norm{\nabla g}_2$ give \begin{align*} \lim_{r\to0}G_\nu(r) =\E_\nu\norm{u\cdot\nabla g}_2^2 =\int_D\nabla g^\top M_\nu\nabla g dx. \end{align*} Also $G_\nu(r)\le4r^{-2}\norm{g}_2^2\to0$ as $r\to\infty$. Taking the two endpoint limits proves \eqref{eq:directional-dirichlet}. The one-sided sum satisfies \begin{align*} \mathcal E_{\nu,1}^{\mathbb N}(g) =4\int_D\nabla g^\top M_\nu\nabla g dx-4G_\nu(2r_0). \end{align*} The subtracted term is at most $4r_0^{-2}\norm{g}_2^2$. Positive definiteness of $M_\nu$ therefore makes $\norm{g}_2^2+\mathcal E_{\nu,1}^{\mathbb N}(g)$ equivalent to the squared $H^1=B^1_{2,2}$ norm.

To include all $g\in L^2(D)$, suppose the one-sided energy is finite. The finite-band identity bounds $G_\nu(r_j)$ uniformly as $j\to\infty$. Plancherel's identity and Fatou's lemma then give a finite Fourier integral of $\abs{\widehat g(\omega)}^2\omega^\top M_\nu\omega$, which becomes a Fourier sum on the torus. If $M_\nu$ is positive definite, finiteness of the Fourier integral implies $g\in H^1(D)$, so the preceding argument applies. Under (A1), each squared difference moment is bounded between $c_\mu$ and $C_\mu$ times the corresponding Lebesgue moment, which proves the density-weighted statement. Finally, rotational invariance on $E$ and $\operatorname{tr}M_{\sigma_E}=1$ give $M_{\sigma_E}=P_E/\dim E$; linearity in $\nu$ yields \eqref{eq:mixture-gradient-weight}. \hfill$\square$

\subsection{Proposition~\ref{prop:block-anisotropic} on block anisotropic energy} \label{app:proof-block-anisotropic} Apply Proposition~\ref{prop:angular-besov} to each block $E_b$ with exponent $s_b$. The left side of \eqref{eq:block-anisotropic-besov} is equivalent to \begin{align*} \int_{\R^d}|\widehat g(\omega)|^2 \left(1+\sum_b\norm{P_{E_b}\omega}_2^{2s_b}\right)d\omega. \end{align*} The positive weights $\lambda_b$ enter the comparison constants.

For completeness, this multiplier follows directly from the anisotropic modulus in \eqref{eq:anisotropic-modulus}. For a rectangle of increments $|v_i|\le t_i$, the Fourier bound for each difference gives \begin{align*} \omega_2(f,\mathbf t)_2^2 \lesssim\int|\widehat f(\zeta)|^2 \sum_i\min\{1,|t_i\zeta_i|^4\} d\zeta. \end{align*} For the reverse comparison, average differences along each coordinate interval $[-t_i,t_i]e_i$ and then average over coordinates. These increments lie in the rectangle, and $(1/2)\int_{-1}^1|e^{iv a}-1|^4dv\asymp\min\{1,|a|^4\}$. The resulting integral is bounded by $\omega_2(f,\mathbf t)_2^2$, up to a dimension-dependent factor. Taking $t_i=2^{-k/\beta_i}$ and summing with weight $2^{2k}$, the same geometric-series argument, now with $k\ge0$, shows that \eqref{eq:anisotropic-modulus} is equivalent to \begin{align*} \int|\widehat{\widetilde g}(\zeta)|^2 \left(1+\sum_i|\zeta_i|^{2\beta_i}\right)d\zeta. \end{align*} The $L^2$ term absorbs $|\zeta_i|\le1$. Within block $b$, $\sum_{i\in b}|\zeta_i|^{2s_b}\asymp\norm{\zeta_b}_2^{2s_b}$. The orthogonal change of variables $\zeta=Q\omega$ now identifies the two Fourier expressions and proves \eqref{eq:block-anisotropic-besov}. \hfill$\square$

\subsection{Proposition~\ref{prop:weighted-telescope} on measure translation} \label{app:proof-weighted-telescope} Set $q_h(x)=g(x+h)-g(x)$. The parallelogram identity for $q_h(x+h)$ and $q_h(x)$ gives the pointwise identity \begin{align*} |\Delta_h^2g(x)|^2 =4|\Delta_hg(x)|^2-|\Delta_{2h}g(x)|^2 +2(|q_h(x+h)|^2-|q_h(x)|^2). \end{align*} Since $\int|q_h(x+h)|^2d\mu(x)=\int|q_h(x)|^2d\mu_h(x)$, integration, averaging over $h=ru$, and division by $r^2$ yield \begin{align*} r^{-2}\E_\nu\norm{\Delta_{ru}^2g}_{L^2(\mu)}^2 =4[G_{\mu,\nu}(r)-G_{\mu,\nu}(2r)]+\mathcal D_{\mu,\nu}(r). \end{align*} Summing and using $2r_j=r_{j-1}$ proves \eqref{eq:weighted-telescope}. If $d\mu=\rho dx$, then $d\mu_h(x)=\rho(x-h) dx$. The Lipschitz bound $|q_{ru}(x)|\le Lr$ and the $W^{1,1}$ translation inequality give \begin{align*} |\mathcal D_{\mu,\nu}(r)| \le2L^2\E_\nu\norm{\rho(\cdot-ru)-\rho}_{L^1} \le2L^2r\norm{\nabla\rho}_{L^1}. \end{align*} Finally, $\sum_{j=a}^b r_j\le2r_a$ proves \eqref{eq:measure-correction-bound}. \hfill$\square$

\subsection{Proposition~\ref{prop:weighted-kinks} on the weighted kink asymptotic} \label{app:proof-weighted-kinks} For the scalar hinge $t_+$ and a signed increment $\delta$, the second difference $\Delta_\delta^2(t_+)$ is triangular, supported on an interval of length $2|\delta|$ adjacent to zero, and satisfies \begin{align*} \int_{\R}|\Delta_\delta^2(t_+)|^2 dt =2\int_0^{|\delta|}v^2 dv=\frac23|\delta|^3. \end{align*} Away from lower-dimensional intersections of interfaces, continuity makes $g$ locally an affine function plus a hinge $a_F t_+$ in normal coordinate $t$. For a step $r\xi u$, the normal increment is $\delta=r\xi(u\cdot n_F)$. Integrate first in $t$ and then over $F$. Uniform continuity of $\rho$ on a fixed compact neighborhood of $\operatorname{supp}\rho$ permits replacing $\rho(y+tn_F)$ by $\rho(y)$ with an $o(r^3)$ error, uniformly in the bounded probes. The face contribution is therefore \begin{align*} \frac23 a_F^2 r^3\xi^3|u\cdot n_F|^3 \int_F\rho(y) d\mathcal H^{d-1}(y)+o(r^3). \end{align*} Continuity and the finite partition make $g$ globally Lipschitz. Inside the relevant compact set, the $O(r)$ neighborhood of the partition's faces of codimension at least two has volume $O(r^2)$. Stencils that cross multiple interfaces are confined to such neighborhoods for sufficiently small $r$, up to constants depending on the partition. The squared differences on these stencils are $O(r^2)$, so the total contribution from multiple-interface crossings is $O(r^4)$. Summing the single-interface contributions and averaging over $(\xi,u)$ proves \eqref{eq:weighted-kink-asymptotic}.

For a single hinge, the affine part cancels and the difference depends on $X$ only through $T=n\cdot X-b$. The same triangular integral, now against $f_T(t) dt$, gives \begin{align*} \E|\Delta_{r\xi u}^2g(X)|^2 =\frac23 a^2 f_T(0)\E[\xi^3]\E_\nu|u\cdot n|^3r^3+o(r^3). \end{align*} Continuity near zero makes the remainder uniform over bounded increments. Under the stated manifold transversality condition, local coordinates can use $T$ as one variable; integration over the remaining variables gives the continuous density of $T$. Finally, \eqref{eq:critical-band-intensity} follows by taking the limit of each of the $L$ normalized terms. \hfill$\square$

\subsection{Proposition~\ref{prop:uniform-besov} on uniform Besov control} \label{app:proof-uniform-besov} For $t>0$, write the modulus of smoothness as \begin{align*} \omega_m(g,t)_\infty\coloneqq \sup_{\substack{\norm{h}_2\le t\\ z,z+h,\ldots,z+mh\in\Omega}}\abs{\Delta_h^m g(z)}. \end{align*} On a box, all stencil points are required to lie in $\Omega$; on the torus, translations are periodic. The classical difference characterization on boxes \citep{devore1993constructive} and on the torus \citep[Ch.~3]{schmeisser1987topics} gives, for $0<s<m$ and $0<q<\infty$, \begin{align*} \norm{g}_{B^s_{\infty,q}(\Omega)}\asymp\norm{g}_\infty+ \left(\int_0^{t_0}[t^{-s}\omega_m(g,t)_\infty]^q\frac{dt}{t}\right)^{1/q}, \end{align*} with the integral expression replaced by a supremum when $q=\infty$. Any fixed cutoff $t_0>0$ can be used, with constants depending on $t_0$; the bounded range of scales between two cutoffs is absorbed into $\norm{g}_\infty$ using $\omega_m(g,t)_\infty\le2^m\norm{g}_\infty$. Take $t_j=\sqrt2r_j$. Monotonicity of $\omega_m(g,t)_\infty$ bounds the weighted modulus integral on $[t_{j+1},t_j]$ by the adjacent dyadic values, with factors depending only on $s$ and $q$. The coarsest term is again bounded by $2^m t_0^{-s}\norm{g}_\infty$. Since $U_j^{(m)}(g;\Omega)=\omega_m(g,t_j)_\infty$, this dyadic comparison proves \eqref{eq:uniform-besov}, including $q=\infty$.

Every admissible probe satisfies $\abs{\Delta^m_{h_j}g(z)}\le U_j^{(m)}(g;\Omega)$ pointwise. Taking $L^p$ moments of this bound gives $A_j(g)\le U_j^{(m)}(g;\Omega)$ under any probability law, without a measure change. Monotonicity of the $\ell^q$ aggregate and \eqref{eq:uniform-besov} now give \eqref{eq:uniform-moment-control}. \hfill$\square$

\subsection{Proposition~\ref{prop:trunc} on finite-grid truncation} For $j>j_1$, the density exchange and the annulus-mean identity above, together with the trivial bound of the annulus mean by the supremum modulus, give $A_j \lesssim \omega_m(g,\sqrt2r_j)_p$, and $\omega_m(g,t)_p \lesssim t^{s'}\norm{g}_{B^{s'}_{p,\infty}}$ for $0<s'<m$ \citep[Ch.~2]{devore1993constructive}. Hence $r_j^{-s}A_j \lesssim \norm{g}_{B^{s'}_{p,\infty}} r_j^{s'-s}$.  If $q<\infty$, the reverse triangle inequality in $\ell^q$ gives \begin{align*} 0\le R^{(m),\mathbb N}_{s,p,q}(g)-R_J(g) &\le \left(\sum_{j>j_1}(r_j^{-s}A_j)^q\right)^{1/q}; \end{align*} for $q=\infty$, the same difference is at most $\sup_{j>j_1}r_j^{-s}A_j$.  The geometric tail is therefore bounded by $C\norm{g}_{B^{s'}_{p,\infty}}r_0^{s'-s} 2^{-(j_1+1)(s'-s)}$, carrying an additional factor $(1-2^{-(s'-s)q})^{-1/q}$ when $q<\infty$, which is \eqref{eq:trunc} after absorbing constants. \hfill$\square$

\subsection{Theorem~\ref{thm:kink} on kink density} Since $f_N' = \tau'(Nx) \in \{\pm 1\}$ almost everywhere, we have $\Lip(f_N)=1$; and $f_N$ is piecewise affine, hence $f_N''=0$ off the kink set, which is finite per period. For the seminorm growth, we use the modulus characterization $\norm{f}_{\dot B^s_{p,q}}^q \asymp \int_0^1 [t^{-s}\omega_2(f,t)_p]^qdt/t$ with $\omega_2(f,t)_p \coloneqq \sup_{|h|\le t}\norm{\Delta^2_h f}_{L_p}$ \citep{devore1993constructive}. The function $f_N$ has $2N$ kinks per unit period, with a slope jump of magnitude $2$ at each. For $|h| = t \le 1/(4N)$, $\Delta^2_h f_N(x) = 0$ unless the segment $[x, x+2t]$ contains a kink, an event of measure $\le 4Nt$ per period, and on that event $\abs{\Delta^2_h f_N} \le 4t$ with $\abs{\Delta^2_h f_N} \ge t$ on a proportional sub-event, the distance to the kink being uniform on the segment. Hence $\omega_2(f_N,t)_p \asymp (Nt)^{1/p} t$ for $t \lesssim 1/N$. For $t \gtrsim 1/N$, monotonicity and the bound $\omega_2 \le 4\norm{f_N}_\infty \lesssim 1/N$ give $\omega_2(f_N,t)_p \asymp 1/N$: The lower bound follows, for example, by taking a step comparable to $1/N$, since at $h=1/(2N)$ the $L^p$ norm of the second difference is a positive $p$-dependent constant times $1/N$, and then using monotonicity. Splitting the integral at $t^* \coloneqq 1/N$: \begin{align*} \int_0^{1/N} [t^{1+1/p-s} N^{1/p}]^q \frac{dt}{t} &\asymp N^{q/p} N^{-(1+1/p-s)q} = N^{(s-1)q}, \\ \int_{1/N}^{1} [t^{-s} N^{-1}]^q \frac{dt}{t} &\asymp N^{sq}N^{-q} = N^{(s-1)q}, \end{align*} where the first integral converges precisely when $s < 1+1/p$ and the second is dominated by $t\asymp 1/N$ because $s>1>0$. Taking $q$-th roots yields $\norm{f_N}_{\dot B^s_{p,q}} \asymp N^{s-1}$; the case $q=\infty$ follows by evaluating the supremum at $t \asymp 1/N$. At $p=2$, $s=3/2$, $t^{-3/2}\omega_2(f_N,t)_2\asymp N^{1/2}$ for $t\lesssim1/N$. The supremum of the normalized modulus has this order, while the $q$-th power of the normalized modulus has a divergent $dt/t$ integral for every finite $q$, proving the endpoint statement. \hfill$\square$

\subsection{Lemma~\ref{lem:boundary-local} on local boundary crossing} Within one affine region $g(z) = Az+b$, all second differences cancel exactly: $\Delta^2_h g = (A(z+2h)+b) - 2(A(z+h)+b) + (Az+b) = 0$. Suppose the segment $t\mapsto z+tu$, $t\in[0,2r]$, crosses one boundary at $t=c$. Let $J$ be the jump of the affine linear map and put $\Gamma\coloneqq\norm{Ju}$, the norm of the jump in the directional derivative along $u$. Writing $\phi(t) \coloneqq g(z+tu)$, $\phi$ is piecewise affine with slope jump $Ju$, so a direct computation gives $\Delta^2 = \phi(2r)-2\phi(r)+\phi(0)=(Ju)\kappa(c)$, up to sign, with $\kappa(c)\coloneqq\min(c,2r-c)\in[0,r]$: For $c<r$, the telescoping leaves $(Ju)c$, and for $c>r$, symmetrically, $(Ju)(2r-c)$. Hence $\norm{\Delta^2} \le \Gamma r$ as claimed. \hfill$\square$

\subsection{Theorem~\ref{thm:drift-damage} on task alignment} For fixed $\ell$ and $x$, put $L_{\ell,x}(y)\coloneqq\log(P_x(y)/Q_{\ell,x}(y))$.  The positivity assumption makes every term finite.  Subtracting the two definitions in \eqref{eq:alignment-def} and using $D_{\mathrm{KL}}(P_x\Vert Q_{\ell,x})= \sum_yP_x(y)L_{\ell,x}(y)$ gives, without approximation, \begin{align*} D_\ell-G_\ell &=\E_x\sum_y(q_x(y)-P_x(y))L_{\ell,x}(y), \end{align*} which is~\eqref{eq:alignment-id}. \hfill$\square$

\subsection{Corollary~\ref{cor:drift-damage-bound} on target-mismatch control} Let $R_x(y)\coloneqq q_x(y)/P_x(y)$. Since $\E_{Y\sim P_x}[R_x(Y)-1]=\sum_y(q_x(y)-P_x(y))=0$, centering the log ratio does not change the inner product: \begin{align*} \sum_y(q_x(y)-P_x(y))L_{\ell,x}(y) &=\E_{Y\sim P_x}[(R_x(Y)-1) (L_{\ell,x}(Y)-\E_{P_x}L_{\ell,x})]. \end{align*} Cauchy--Schwarz under $P_x$ bounds the absolute value of this centered inner product by \begin{align*} (\E_{P_x}(R_x-1)^2)^{1/2} (\operatorname{Var}_{P_x}L_{\ell,x})^{1/2} &=\chi^2(q_x\Vert P_x)^{1/2} \operatorname{Var}_{P_x}(L_{\ell,x})^{1/2}. \end{align*} Taking the expectation over $x$, using the triangle inequality, and applying Cauchy--Schwarz once more to the two nonnegative factors gives \begin{align*} \abs{D_\ell-G_\ell} &\le \E_x[\chi^2(q_x\Vert P_x)^{1/2} \operatorname{Var}_{P_x}(L_{\ell,x})^{1/2}]\\ &\le (\E_x\chi^2(q_x\Vert P_x))^{1/2} (\E_x\operatorname{Var}_{P_x}(L_{\ell,x}))^{1/2} =C_{\mathrm{tgt}}V_\ell, \end{align*} proving~\eqref{eq:alignment-bound}. \hfill$\square$

\subsection{Proposition~\ref{prop:margin} on the margin bound} If $m_\theta(x)\ge\gamma$ and $\abs{\Delta_h m_\theta(x)}<\gamma$, then $m_\theta(x+h) > 0$, so the prediction persists at the probed point. Thus $\{\hat y \text{ changes}\} \subseteq \{m_\theta(x) < \gamma\} \cup \{\abs{\Delta_h m_\theta(x)} \ge \gamma\}$; apply the union bound and Markov's inequality with the $p$-th moment. \hfill$\square$

\subsection{Lemma~\ref{lem:dyadic-reconstruct} on the dyadic identity} For $J=1$, direct expansion of the finite differences gives \begin{align*} \Delta_{2h}^1g(z)-\Delta_h^2g(z) &=(g(z+2h)-g(z)) -(g(z+2h)-2g(z+h)+g(z)) =2\Delta_h^1g(z), \end{align*} and hence $\Delta_h^1g(z)=(1/2)\Delta_{2h}^1g(z)- (1/2)\Delta_h^2g(z)$. Apply this one-step identity successively to the first-difference term at scales $h,2h,\dots,2^{J-1}h$. An induction on $J$ then yields \begin{align*} \Delta_h^1g(z)&=2^{-J}\Delta_{2^Jh}^1g(z) -\sum_{a=0}^{J-1}2^{-(a+1)}\Delta_{2^ah}^2g(z), \end{align*} which is~\eqref{eq:dyadic-reconstruct}. The stated segment condition ensures that every value in this algebraic identity is defined. \hfill$\square$

\subsection{Theorem~\ref{thm:dyadic-churn} on dyadic amplitude control} Take the norm in the target space, then the $L^p$ norm over the common probe law. Since $p\ge1$, Minkowski's inequality and positive homogeneity give \begin{align*} (\E\norm{\Delta_h^1g(z)}^p)^{1/p} &\le 2^{-J}(\E\norm{\Delta_{2^Jh}^1g(z)}^p)^{1/p} +\sum_{a=0}^{J-1}2^{-(a+1)} (\E\norm{\Delta_{2^ah}^2g(z)}^p)^{1/p}, \end{align*} which is~\eqref{eq:dyadic-a1}. Scale coherence is used only to identify each term on the right with the population amplitude at nominal scale $2^ar$; the population expectation is unaffected if an implementation estimates those marginals with independent draws. \hfill$\square$

\subsection{Corollary~\ref{cor:dyadic-churn} on second-order churn control} For $g=m_\theta$, Proposition~\ref{prop:margin} gives $\Prob\{\hat y(x+h)\ne\hat y(x)\}\le \Prob\{m_\theta(x)<\gamma\}+A^{(1)}(r)^p/\gamma^p$. Substituting~\eqref{eq:dyadic-a1} proves~\eqref{eq:dyadic-churn}. \hfill$\square$

\subsection{Proposition~\ref{prop:alignment-inversion} on the target-mismatch obstruction} For~\eqref{eq:alignment-inversion}, direct substitution yields \begin{align*} D_A&=\log(60/59), & G_A&=0.6\log(60/59)+0.2\log(40/81)+0.2\log40,\\ D_B&=\log(6/5), & G_B&=0.6\log(6/5)+0.4\log(4/5). \end{align*} These exact expressions evaluate respectively to $D_A\approx0.016807$, $G_A\approx0.606746$, $D_B\approx0.182322$, and $G_B\approx0.020136$, giving both strict reversed inequalities.  Adding the reverse-KL terms and dividing by two gives $\dskl(P,Q_A)^2\approx0.432071$ and $\dskl(P,Q_B)^2\approx0.020273$, so the symmetric member of the drift family reverses the task ordering as well. \hfill$\square$

\subsection{Proposition~\ref{prop:damage} on damage decomposition} Second-order Taylor with integral remainder at $0$, using $\Delta(0)=0$, gives $\Delta(\delta_S) = g_1^\top\delta_S + (1/2)\delta_S^\top H\delta_S + R_S$ and $D_\ell = g_1^\top\delta_\ell + (1/2)\delta_\ell^\top H\delta_\ell + R_\ell$ with $g_1\coloneqq\nabla\Delta(0)$ and $\abs{R_\bullet}\le(M_3/6) \norm{\bullet}^3$. Since $\delta_S = \sum_{\ell\in S}\delta_\ell$, the linear parts obey $g_1^\top\delta_S = \sum_\ell g_1^\top\delta_\ell$ and cancel; the quadratic parts give $(1/2)\delta_S^\top H\delta_S - (1/2)\sum_\ell\delta_\ell^\top H\delta_\ell = \sum_{\{\ell,\ell'\}\subseteq S} \delta_\ell^\top H\delta_{\ell'}$ by bilinearity, disjoint supports notwithstanding, as $H$ couples across blocks; and $\mathcal{R} = R_S - \sum_\ell R_\ell$ obeys the stated bound by the triangle inequality. \hfill$\square$

\subsection{Lemma~\ref{lem:alloc-modularity} on finite-path approximate modularity} Fix a set $S=\{i_1,\dots,i_k\}$ in an arbitrary order and write $x_r\coloneqq\sum_{a=1}^r\delta_{i_a}$, with $x_0=0$. Telescoping along this path through the block box gives \begin{align*} \mathcal D(S)-\sum_{r=1}^kD_{i_r} &=\sum_{r=1}^k (\Delta(x_{r-1}+\delta_{i_r})-\Delta(x_{r-1}) -\Delta(\delta_{i_r})+\Delta(0)). \end{align*} For the $r$-th summand, set $G_r(y)\coloneqq\Delta(y+\delta_{i_r})-\Delta(y)$. The fundamental theorem of calculus, first along $t\mapsto tx_{r-1}$ and then along the $\delta_{i_r}$ direction, gives the exact identity \begin{align*} G_r(x_{r-1})-G_r(0) &=\int_0^1 x_{r-1}^\top (\nabla\Delta(tx_{r-1}+\delta_{i_r})- \nabla\Delta(tx_{r-1}))dt\\ &=\int_0^1\int_0^1 x_{r-1}^\top\nabla^2\Delta(tx_{r-1}+s\delta_{i_r}) \delta_{i_r}dsdt. \end{align*} Every point $tx_{r-1}+s\delta_{i_r}$ lies in $\mathcal B$, and $x_{r-1}=\sum_{a<r}\delta_{i_a}$. Therefore the triangle inequality and the definition in~\eqref{eq:interaction-envelope} imply \begin{align*} \abs{G_r(x_{r-1})-G_r(0)} &\le\sum_{a<r}I_{i_ai_r}. \end{align*} Summing over $r$ counts every unordered pair in $S$ exactly once and proves both inequalities in~\eqref{eq:uniform-modularity}. Notice that the variation of the Hessian along the full finite path is already inside the supremum; no Taylor remainder or third derivative is omitted. \hfill$\square$

\subsection{Theorem~\ref{thm:alloc-stability} on interaction-stable allocation} Let $S^\star$ minimize the true configuration damage and let $\widehat S$ minimize the estimated additive score. On the event $\max_\ell\abs{\widehat D_\ell-D_\ell}\le\varepsilon$, optimality of $\widehat S$ for the estimated score gives \begin{align*} \sum_{\ell\in\widehat S}D_\ell &\le \sum_{\ell\in\widehat S}\widehat D_\ell+k\varepsilon \le \sum_{\ell\in S^\star}\widehat D_\ell+k\varepsilon \le \sum_{\ell\in S^\star}D_\ell+2k\varepsilon. \end{align*} Apply~\eqref{eq:uniform-modularity} once above for $\widehat S$ and once below for $S^\star$: \begin{align*} \mathcal D(\widehat S) &\le\sum_{\ell\in\widehat S}D_\ell+\Lambda_k \le\sum_{\ell\in S^\star}D_\ell+2k\varepsilon+\Lambda_k \le\mathcal D(S^\star)+2k\varepsilon+2\Lambda_k. \end{align*} The lower bound by zero follows from the definition of $S^\star$, proving \eqref{eq:alloc-stability}. \hfill$\square$

\subsection{\eqref{eq:alignment-allocation} on label-free allocation transfer} If $\kappa=\max_\ell C_{\mathrm{tgt}}V_\ell$ and $\max_\ell\abs{\widehat G_\ell-G_\ell}\le\eta$, the triangle inequality and \eqref{eq:alignment-bound} give $\max_\ell\abs{\widehat G_\ell-D_\ell}\le\kappa+\eta$.  Substitute this uniform proxy error for $\varepsilon$ in Theorem~\ref{thm:alloc-stability} to obtain \eqref{eq:alignment-allocation}. \hfill$\square$

\subsection{Theorem~\ref{thm:alloc-regret} on the allocation minimax gap} Every instance of $\mathcal{P}_{L,c,M}$ presents the identical block jet $(0,0,c)$ at all $L$ blocks, so a deterministic rule that reads the instance only through those jets returns a low-precision set $S_0$ with $|S_0|=L-K$ that is the same for every instance; fix that set, and write $P\coloneqq\{1,\dots,L\}\setminus S_0$ for the protected set, so that $|P|=K$. For the lower bound, take one-dimensional blocks $\delta_\ell\coloneqq e_\ell$ and the separable global loss increase $\Delta(v)\coloneqq\sum_{\ell=1}^L((c/2)v_\ell^2+\beta_\ell v_\ell^3)$ with $\beta_\ell\in\{0,M\}$. The block profiles of $\Delta$ $\phi_\ell(t)=(c/2)t^2+\beta_\ell t^3$ have jet $(0,0,c)$, lie in $C^3([0,1])$, and realize $D_\ell=c/2+\beta_\ell\in[c/2, c/2+M]$; joint damages are additive. Thus every such instance is in $\mathcal{P}_{L,c,M}$, and even the full block Hessians and displacement norms exposed to standard Hessian-based scores are identical. Put $\beta_\ell=M$ on a set $G\subseteq S_0$ with $|G|=\min(K,L-K)$, feasible as $|S_0|=L-K\ge\min(K,L-K)$, and $\beta_\ell=0$ elsewhere. Since $|G|\le K$, at least $L-K$ blocks carry damage $c/2$, so the oracle low-precision set achieves $\min_{|S'|=L-K}\mathcal{D}(S')=(L-K)c/2$, whereas the rule pays $\mathcal{D}(S_0)=(L-K)c/2+M|G|$; hence $\mathrm{Reg}(S_0)=M\min(K,L-K)$. For the matching upper bound, take any instance and let $S^\star$ be an oracle low-precision set. The two sets $S_0$ and $S^\star$ both have size $L-K$, so $r\coloneqq|S_0\setminus S^\star|=|S^\star\setminus S_0|\le L-K$; also $S^\star\setminus S_0\subseteq P$, hence $r\le K$. The common blocks cancel, and, writing $D_\ell=c/2+\beta_\ell$ with $\beta_\ell\in[0,M]$, \begin{align*} \mathrm{Reg}(S_0) &=\sum_{\ell\in S_0\setminus S^\star}\beta_\ell -\sum_{\ell\in S^\star\setminus S_0}\beta_\ell \le rM\le\min(K,L-K)M. \end{align*} The two matching bounds prove \eqref{eq:alloc-regret}. \hfill$\square$

\section{Experimental setup: pools, models, and protocol} \label{app:setup-details}

This section details the design principle, recipes, pool composition, and protocol layers behind Sec.~\ref{sec:results}. The vision suite contains 80 checkpoints trained from scratch under strong standard recipes, namely 30 runs of a CIFAR-stem ResNet-18 on CIFAR-10 and 21 on CIFAR-100 \citep{krizhevsky2009learning}, 12 runs of WRN-28-10 on CIFAR-10 \citep{zagoruyko2016wide}, and 12 runs of the attention-based vision transformer ViT-S/4 trained from scratch with a DeiT-style recipe \citep{dosovitskiy2021image,touvron2021training}, each pool spanning recipe knobs including SAM and RSC. The two ResNet pools additionally contain a Jacobian penalty and the auxiliary RSC $+$ crossing variant of App.~\ref{app:crossing-penalty}. We evaluate these checkpoints under CIFAR-10-C and CIFAR-100-C \citep{hendrycks2019benchmarking}, per-channel symmetric weight RTN PTQ at int8, int6, and int4, and additive Gaussian input noise. The language-model suite is Pythia-160M to 6.9B \citep{biderman2023pythia} with a preregistered held-out check at 12B, seven modern deployment LLMs from six vendors (Sec.~\ref{sec:exp-modllm}), Qwen2.5, and Mamba, on WikiText-103 perplexity and LAMBADA, with per-block scores from 64 calibration sequences, the AWQ-style activation salience $\times$ relative weight MSE and HAWQ-V2 \citep{dong2019hawq,dong2020hawqv2} as principal allocation baselines, and the per-block $\Delta$NLL from quantizing that block alone as ground truth. The paragraphs below give the design principle behind these pools, full recipes and pool composition, the per-configuration seed counts and replication conventions, the complete baseline set, calibration budgets, the HAWQ-V2 trace estimator, the post hoc drift--damage diagnostic pass, estimator reliability, intervention-fairness accounting, and the compute infrastructure; App.~\ref{app:measurement} gives the measurement details.

\paragraph{Design principle: shared perturbation protocols.} Each experiment builds a pool of checkpoints trained with strong, standard, fully disclosed recipes, differing in seeds and recipe knobs. We then measure spectra and scalar proxies on clean data only, evaluate every checkpoint under the same perturbation protocols, and compare the targeted interventions and baselines under the stated clean-performance, compute, and precision budgets. Diagnostics are computed strictly from clean data and model internals; performance under perturbation is evaluated using external corruptions or quantizers.

\paragraph{Vision pools.} A CIFAR-stem ResNet-18 on CIFAR-10 and CIFAR-100 \citep{krizhevsky2009learning}: stochastic gradient descent with momentum 0.9 (Nesterov), learning rate 0.1, batch 128, weight decay $5\times10^{-4}$, cosine annealing, 200 epochs, random crop and horizontal flip, a standard strong recipe. Each pool variant uses 3 seeds. The CIFAR-10 variants are baseline, weight decay $0$, label smoothing 0.1, mixup 0.2, SAM with $\rho=0.05$, Jacobian penalty, RSC, RSC $+$ crossing, GroupNorm, and learning rate 0.3, for 30 runs; CIFAR-100 uses the same set without the weight decay $0$, GroupNorm, and learning-rate-0.3 knobs, for 21 runs.

To test architecture generality, we train a second pool with WRN-28-10 \citep{zagoruyko2016wide} on CIFAR-10 under the identical recipe, running 2 seeds of each of baseline, weight decay $0$, label smoothing, mixup, SAM, and RSC for 12 runs, and a third pool with an attention-based architecture: ViT-S/4 \citep{dosovitskiy2021image} with 12 layers, $d=384$, 6 heads, patch 4, and $\approx$21M parameters, trained from scratch with a standard DeiT-style recipe \citep{touvron2021training}, namely AdamW at learning rate $10^{-3}$ with weight decay 0.05 on matrices only, batch 512, 300 epochs with 10-epoch warmup then cosine, RandAugment$(2,9)$ plus crop and flip on hard labels, stochastic depth 0.1, and grad clip 1.0, again with 2 seeds of each of baseline, label smoothing, mixup, SAM, RSC, and no stochastic depth for 12 runs.

To probe hyperparameter sensitivity of the principal smoothness baseline, the CIFAR-10 pool additionally contains SAM at $\rho\in\{0.1, 0.2\}$ and an RSC $+$ SAM combination at a single seed each, and the CIFAR-100 pool a wall-clock-matched SAM arm of 300 epochs at 2 seeds, matching RSC's measured wall-clock, giving 80 vision runs total. Evaluation conditions: CIFAR-10-C and CIFAR-100-C over 15 corruptions $\times$ 5 severities \citep{hendrycks2019benchmarking}, per-channel symmetric weight PTQ at int8, int6, and int4, and additive Gaussian input noise.

\paragraph{Seeds and replication conventions.} Training seeds enter the vision suite as follows: Every primary configuration is replicated, at 3 seeds in the two ResNet-18 pools and 2 seeds in the WRN-28-10 and ViT-S/4 pools, and every replicated entry in Table~\ref{tab:vision-c10} and Tables~\ref{tab:vision-c100} through~\ref{tab:vision-vit} reports the mean $\pm$ half the range over the available measurements for that configuration, where half the range is the maximum minus the minimum, divided by two, with the CIFAR-10 baseline's corruption columns using two available seeds. Figure~\ref{fig:vision-int} shows the mean over seeds and whiskers spanning those seeds. The three auxiliary hyperparameter-sensitivity arms are the exception: SAM at $\rho\in\{0.1,0.2\}$ and RSC $+$ SAM are run at a single seed only and print as bare numbers, so no seed spread attaches to those particular numbers and those arms read as single-run probes of sensitivity rather than as replicated comparisons; the wall-clock-matched SAM arm is replicated at 2 seeds like the other CIFAR-100 rows. Two or three runs per configuration fix the direction of the intervention effects but not fine-grained effect sizes. Seeds enter the measurement side separately and independently of training: Spectra are estimated at a fixed probe seed with reliability quantified across three independent probe seeds (App.~\ref{sec:exp-reliability}), and the per-block LLM scores of Sec.~\ref{sec:exp-quant} are reported at a single fixed calibration draw, with the diagnostic pass re-measuring those scores at four fresh draws whose seeds are disjoint from that reported draw (App.~\ref{app:measurement}).

\paragraph{Pretrained LLMs.} Pythia-160M, 410M, 1.4B, 2.8B, and 6.9B \citep{biderman2023pythia}, trained on the Pile \citep{gao2020pile}, evaluated on WikiText-103 at token-level perplexity over non-overlapping 1024-token windows and on LAMBADA at strict greedy last-word accuracy. Per-block $B^{\mathrm{quant}}_{\ell,4}$ from 64 calibration sequences; allocation baselines: relative weight-MSE, activation salience $\times$ MSE (AWQ-style), first-$K$ and last-$K$, random, and HAWQ-V2 \citep{dong2019hawq,dong2020hawqv2}, the positive part of the per-block mean Hessian trace times the int4 weight perturbation, $(\mathrm{Tr}(H_\ell)/n_\ell)_+ \norm{Q_4(W_\ell)-W_\ell}_F^2$ (App.~\ref{app:measurement}); ground truth: per-block $\Delta$NLL when quantizing that block alone. The per-block drift score's collapse at 2.8B and 6.9B (Sec.~\ref{sec:exp-quant}) motivates a post hoc drift--damage diagnostic pass: Four independent fresh 64-sequence calibration draws per model yield per-sequence drift (\eqref{eq:bquant}), damage (\eqref{eq:dquant}), and top-1 flip rates per block, with bootstrap CIs and damage-score allocations evaluated exactly like the other allocation rows (App.~\ref{app:measurement}); a final forward-only depth-propagation probe at 2.8B and 6.9B separates magnitude from direction accounts of the dissociation (App.~\ref{app:depth}).

\paragraph{Scope of the vision endpoints.} Proposition~\ref{prop:margin} bounds prediction churn within the probed perturbation family and nowhere else, and Theorem~\ref{thm:dyadic-churn} shows that the second-order grid actually measured in vision controls that first-order term, while Corollary~\ref{cor:dyadic-churn} turns the control into a churn bound once one coarse anchor is supplied. The primary vision endpoint is the accuracy drop under additive Gaussian input noise at $\sigma=0.05$, which tests robustness at the probed noise scale. The CIFAR-C corruptions test how far the diagnostic carries beyond the probed family, with the noise-type corruptions nearest and blur, weather, and digital farthest \citep{yin2019fourier}.

\paragraph{Protocol and fairness.} Three further protocol layers are specified in App.~\ref{app:measurement}. Estimator reliability covers three independent probe seeds and two disjoint base-point halves per checkpoint, reported in App.~\ref{sec:exp-reliability}. The intervention-fairness accounting covers the identical base recipe and budget for every training-time intervention, the untuned first-run hyperparameters, the auxiliary SAM $\rho$-grid and wall-clock-matched SAM arms, and the $0.4\%$ pool tolerance behind matched clean accuracy. The signed-Rayleigh convention $\lambda_{\mathrm{dom}}\coloneqq v^\top Hv$ used by every vision table is recorded there as well, together with the compute infrastructure.

\section{Measurement details} \label{app:measurement}

\paragraph{Radius convention.} All spectra use dyadic radii $r_j \coloneqq r_0 2^{-j}$, $j = 0,\dots,7$, with per-sample log-uniform jitter in $[r_j/\sqrt2, r_j\sqrt2]$. Raw amplitudes $A_j$ are stored; $s$-weighted summaries $R_{s,p,q}$ are formed at analysis time. Unless stated otherwise $p=q=2$ and $m=2$ for vector-valued targets, with the $m=1$ divergence analogue used for output distributions.

\paragraph{Vision margin spectra.} Base points: 1024 test images, identical across checkpoints; 4 directions per family per scale; $r_0 = 16$ in the channel-normalized input space of per-pixel unit variance, where the coarsest scale corresponds roughly to severity-1 and severity-2 common corruptions and the finest to $\norm{h}_2 = 0.125$. The direction families are the isotropic Gaussian, normalized per sample; the HF and low-frequency families, Gaussian noise filtered by radial FFT masks, high meaning radial frequency $\ge 0.35$ and low meaning $\le 0.15$ in normalized units; and augmentation directions, normalized displacement toward a random 1-pixel to 2-pixel translation with brightness jitter. Margins are evaluated in \texttt{float32} with BatchNorm in evaluation mode, since fine-scale second differences fall below \texttt{bfloat16} resolution.

\paragraph{Margin normalization of spectrum quantities.} Margin-based quantities are reported under the empirical margin convention $\tilde R \coloneqq R / \overline{\abs{m}}$ with $\overline{\abs{m}} \coloneqq \E_x\abs{m_\theta(x)}$ on the same base points. This margin normalization expresses roughness relative to the model's typical margin magnitude and is invariant to a common positive rescaling of all logits. We apply the identical normalization to the margin-based empirical Lipschitz baseline; raw and margin-normalized forms are stored for both, and the reported vision tables use the margin-normalized forms.

\paragraph{Source diagnostics (vision).} The empirical unit-wise sign-flip rate $\widehat c_j$ is the fraction of ReLU pre-activations, taken over all ReLU modules, 17 for ResNet-18 and 25 for WRN-28-10, whose sign flips between $x$ and $x+r_j u$, averaged over 512 test points and 2 isotropic directions. The unconditional gradient-change statistic $\widehat J_j\coloneqq\E\norm{\nabla_x m(x+r_j u)-\nabla_x m(x)}_2$ is measured at $j\in\{0,2,4,6\}$ over 256 points. Both quantities are descriptive diagnostics of activation changes along the probes.

\paragraph{Reliability protocol.} Because $A_j$ is a Monte Carlo estimate over 1024 base points and 4 directions per scale, every checkpoint's margin spectra are re-measured with two additional independent probe seeds, using fresh directions and radius jitter on the same 1024 base points, and once on each of two disjoint 512-point base-point halves at the same probe seed. Reported statistics: mean within-checkpoint standard deviation (SD) across the three probe seeds, read against the between-checkpoint spread; SD of the per-checkpoint means across the pool; intraclass correlation coefficient ICC(1); split-half Pearson and Spearman correlations with Spearman--Brown correction. The pool-level results are in App.~\ref{sec:exp-reliability}.

\paragraph{HAWQ-V2 baseline (Pythia).} We estimate the per-block Hessian trace with the Hutchinson stochastic trace estimator \citep{hutchinson1989stochastic}, using 6 Rademacher probes restricted to the block's weight matrices and double-backpropagation on an fp32 model copy with eager attention over 4 calibration sequences of 1024 tokens, or 2 for Pythia-2.8B. The Hessian-vector product is accumulated one sequence at a time, since the mean-loss Hessian is the mean of per-sequence Hessians; we verified the accumulated estimator is exactly self-consistent, whereas batched double-backprop through the eager attention path is not, so the per-sequence form is used throughout. The allocator uses the nonnegative sensitivity convention \begin{align*} \mathrm{HAWQ}_\ell&\coloneqq (\mathrm{Tr}(H_\ell)/n_\ell)_+ \norm{Q_4(W_\ell) - W_\ell}_F^2, \qquad a_+\coloneqq\max\{a,0\}, \end{align*} following HAWQ-V2's sensitivity-times-perturbation criterion \citep{dong2020hawqv2}. The raw signed trace is reported separately in every H trace diagnostic column; only the allocation score takes the positive part of the trace. Allocation and evaluation use the same pipeline as all other scoring rules.

\paragraph{Drift--damage diagnostic pass (Pythia).} This post hoc pass responds to the drift score's rank collapse at 2.8B and 6.9B. For each model we take four independent calibration draws of 64 fresh sequences, at seeds disjoint from the single reported draw. Each block's int4 probe forward pass yields three per-sequence functionals against the cached full-precision logits on the same windows: mean symmetric KL, whose square root gives the drift amplitude in \eqref{eq:bquant}; the damage of \eqref{eq:dquant}, which is the paired NLL difference on the window's own next tokens; and the top-1 next-token flip rate. Reported statistics: pooled 256-sequence scores; Spearman against the 131,072-token ground truth, per draw and pooled, with percentile-bootstrap 95\% CIs over sequences from 2000 resamples; inter-draw pairwise rank agreement; depth-split correlations; and the damage-per-unit-drift profile. The drift score is reproduced almost exactly across those draws: The inter-draw pairwise rank agreement is $0.98$ to $0.99$, and the rank agreement with the reported draw is $0.97$ at 2.8B and $0.99$ at 6.9B. Damage-ranked additive top-$K$ allocations are evaluated exactly like every other allocation row, on full validation perplexity and LAMBADA, so a damage top-$K$ set that coincides with another rule's set carries the identical number.

\paragraph{Scalar baselines (vision).} Signed dominant-magnitude Hessian eigenvalue $\lambda_{\mathrm{dom}}$: power iteration over 20 steps, with early stop after stabilization, on the full-parameter training-loss Hessian over a fixed batch of 2048 training images, with BatchNorm in eval mode and Hessian-vector products accumulated over 256-image chunks, which is exact for the mean-batch Hessian and bounds the double-backprop graph; Hessian trace: the Hutchinson estimator \citep{hutchinson1989stochastic} with 8 Rademacher probes; parameter-gradient norm on the same fixed batch; empirical Lipschitz: the mean, max, and q99 of $\norm{\nabla_x m_\theta(x)}_2$ over 2048 test points, a sampled lower bound on the true constant, which is NP-hard to compute exactly and is overestimated by the usual layerwise upper bounds \citep{virmaux2018lipschitz}, so we report the sampled quantity that the comparison against a measured spectrum actually calls for.

\paragraph{Weight quantization protocol.} We apply per-output-channel symmetric uniform quantization to weight matrices: all convolutional and linear layers in vision, and transformer-block linear layers with embeddings kept in floating point for the LLMs. We use bit widths in $\{8,6,4\}$ for vision and $\{8,4\}$ for the LLMs. We use neither activation quantization nor reconstruction fine-tuning, so this evaluation isolates sensitivity to quantization noise in the weights.

\paragraph{GPTQ variant for the allocation-transfer check.} One-shot, non-sequential GPTQ: For every transformer-block Linear, the layer Hessian $H \coloneqq \sum_x x x^\top$ is accumulated over the 64 calibration sequences in a single full-precision forward pass; columns are then quantized in blocks of 128 with Cholesky-based error compensation \citep{frantar2022gptq} onto the same per-output-channel symmetric grid as RTN, with damping $0.01\overline{\mathrm{diag}(H)}$ and no activation reordering. Mixed configurations assemble the precomputed int8 and int4 GPTQ weights per block. In the quantizer-transfer check of App.~\ref{app:alloc-extra}, the int8 sets are the RTN-greedy selections, unchanged. In the full GPTQ-grid comparisons of App.~\ref{sec:exp-gptq}, the same GPTQ construction is used, but all per-block scores and allocation sets are re-measured and re-derived on the GPTQ weights.

\paragraph{RSC objective and hyperparameters.} We use $\lambda_{\mathrm{RSC}}=0.05$ throughout. Each step independently samples one direction family uniformly from isotropic and HF and one radius $r_j$ uniformly from $\mathcal R_{\mathrm{train}}=\{4,2,1,0.5\}$. Directions have unit $L^2$ norm, and the training penalty uses these radii without jitter. Define $A_{j,\mathrm{train}}^{(2)}\coloneqq (\E_{x,u}|\Delta^2_{r_j u}m_\theta(x)|^2)^{1/2}$ under this training-data and direction-mixture law. Then \begin{equation} \mathcal L_{\mathrm{RSC}} =\frac{1}{|\mathcal R_{\mathrm{train}}|} \sum_{r_j\in\mathcal R_{\mathrm{train}}} (r_j^{-1}A_{j,\mathrm{train}}^{(2)})^2. \label{eq:rsc-energy} \end{equation} The penalty $\mathcal L_{\mathrm{RSC}}$ is the data-weighted finite-band realization of the directional $s=1$, $p=q=2$ energy in Proposition~\ref{prop:directional-energy}, using $\nu_{1/2}$ from \eqref{eq:direction-mixture}. The RSC penalty is also bounded by the uniform class-margin Besov control in \eqref{eq:rsc-uniform-besov}. The regularization weight was set once at a round auxiliary-loss magnitude and never revisited, with no per-pool, per-metric, or per-scale adjustment. The additional settings for the crossing ablation are in App.~\ref{app:crossing-penalty}.

\paragraph{RSC evaluation.} For evaluation, we use $s=1.5$: The $A_j\asymp r_j^{3/2}$ law makes $r_j^{-3/2}A_j$ proportional to the square root of weighted boundary intensity in the resolved kink regime (Proposition~\ref{prop:weighted-kinks}, App.~\ref{sec:exp-toy}). This amplitude law gives the critical $B^{3/2}_{2,\infty}$ scaling near piecewise-affine slope changes; our finite $q=2$ aggregate accumulates the boundary-intensity contributions over the measured band. The evaluation aggregate places greater relative weight on small radii than the slope-normalized $s=1$ training energy. The primary evaluation diagnostic uses eight jittered scales, HF directions, test base points, and division by the mean absolute margin. Both the training penalty and the evaluation aggregate are data-weighted finite-band summaries of second differences, with the margin normalization expressing roughness relative to typical margin magnitude.

\paragraph{Signed-Rayleigh convention (vision).} For the empirical Hessian baseline, write $\lambda_{\mathrm{dom}}\coloneqq v^\top H v$ for the signed Rayleigh value returned by power iteration after convergence to an eigenvector of largest absolute eigenvalue; thus $\abs{\lambda_{\mathrm{dom}}}$ estimates $\norm{H}_{\mathrm{op}}$.  This signed Rayleigh value is not the algebraically largest eigenvalue $\lambda_{\max}(H)$ when the dominant-magnitude mode is negative.  Every vision table reports the signed value of $\lambda_{\mathrm{dom}}$.

\paragraph{Intervention fairness.} All training-time interventions run under the identical base recipe and budget; SAM uses the standard $\rho=0.05$, the Jacobian penalty $\lambda=0.01$, and RSC $\lambda_{\mathrm{RSC}}=0.05$, each set once, without tuning, from first-run defaults. Because a fixed $\rho$ could understate SAM, the pool additionally contains SAM at $\rho\in\{0.1,0.2\}$, commonly tuned SAM radii, and an RSC $+$ SAM combination that applies the RSC penalty at the SAM ascent point, testing whether the two mechanisms are additive. Per-step cost is reported from measured wall-clock, at SAM $\approx 2.1\times$ and RSC $\approx 3.2\times$; the Jacobian penalty requires double-backpropagation. Within the main pools, we do not grant slower methods extra steps, so those comparisons are budget-matched per epoch rather than per wall-clock hour. To close the remaining asymmetry directly, the CIFAR-100 pool contains a wall-clock-matched SAM arm of $300$ epochs at 2 seeds, matching RSC's measured wall-clock, so SAM is also compared at equal wall-clock (App.~\ref{app:vision-pools}). Matched clean accuracy always means within the pool's matching tolerance of $0.4\%$.

\paragraph{Infrastructure.} All experiments run on 8 NVIDIA GH200 (96GB) nodes; the 80 vision runs are trained in parallel across nodes. Our finite-difference probes use forward passes, while the Hessian baselines use automatic differentiation. The one exception in scale is the largest held-out model, Qwen2.5-32B, which no single node fits: The model's base statistics and greedy-candidate evaluations are sharded across the eight nodes and merged bit-identically; the Qwen2.5-32B HAWQ-V2 Hessian traces, whose plain fp32 autograd graph exceeds a 96GB node, are computed by streaming each block's weights through the graph from the bf16 checkpoint upcast to fp32, a mode validated before use against the single-device fp32 traces on Qwen2.5-14B, which the streaming mode matches at Spearman $1.0$ with bit-identical values; and the Qwen2.5-32B GPTQ solves are disk-backed. These arrangements affect memory use only; all methodological choices remain fixed, including the 32-sequence calibration slice in App.~\ref{app:modllm-extra}.

\begin{samepage} \paragraph{Practitioner-facing procedure and measured cost.} The roughness spectrum requires only forward passes. Choose the perturbed quantity $z$, such as an input, a block's weights, or intermediate features, and the response $g(z)$, such as a margin, an output distribution, a feature vector, or a loss. Choose a probe set $\{z_i\}_{i\le n}$ and direction families $\{\mathcal H\}$ appropriate to that quantity and the expected perturbations, such as isotropic, frequency-band, or augmentation directions. Fix dyadic radii $r_j\coloneqq r_0 2^{-j}$ for $j=0,\dots,7$, finite-difference order $m$, and moment $p$. For each family $\mathcal H$ and scale $j$, take $K$ direction draws, sample $u_{ik}\sim\mathcal H(z_i)$ with independent radius jitters $\xi_{ik}\in[2^{-1/2},2^{1/2}]$, and compute \begin{align*} (\widehat A_j)^p \coloneqq \tfrac1{nK}\textstyle\sum_{i=1}^{n}\sum_{k=1}^{K} \norm*{\sum_{a=0}^{m} (-1)^{m-a}\tbinom{m}{a} g(z_i + a\xi_{ik} r_j u_{ik})}^p, \end{align*} with the $\dskl$ analogue at $m=1$ used for distributions. Report the raw amplitudes $\widehat A_j$, the spectrum $B_j\coloneqq r_j^{-s}\widehat A_j$, the summaries $R_{s,p,q}$, and the log--log slope. For a second-order margin grid, Theorem~\ref{thm:dyadic-churn} converts the grid plus a coarse first-order anchor into a bound on the first-order amplitude, and Corollary~\ref{cor:dyadic-churn} turns that bound into a churn bound.

On one GH200 using autocast forwards, the complete CIFAR-10 ResNet-18 margin suite, covering four direction families, eight scales, four directions, and 1024 probes, together with source statistics and robustness evaluation, takes $45$ seconds. Per-block drift and damage measurement over 32 blocks and 64 sequences, followed by allocation evaluation, takes approximately $14$ minutes on Pythia-2.8B, while the same full per-block suite takes less than one hour on Pythia-6.9B. \end{samepage}

\paragraph{Choosing a diagnostic to run.} The experiments support two practical choices. For vision checkpoints, use the margin-normalized HF spectrum on the nuisance direction family and scale band relevant to the expected perturbations, and use RSC when training-time control is available. For mixed-precision allocation, score blocks with calibration damage and switch to interaction-aware greedy selection when the bit budget is tight or interactions are large. Theorem~\ref{thm:alloc-stability} states the additive rule's conditional near-oracle regime, and Corollary~\ref{cor:drift-damage-bound} states when label-free forward-KL drift may substitute for task damage.

\paragraph{Estimator sampling conventions: clustered probes.} \label{app:estvar} Our implementation reuses each base point for $k$ directions, $k=4$ per scale over 1024 base points, so the $D_i$ within a base point are dependent. We therefore treat the 1024 base-point cluster means in vision as the sampling units and do not count the $4096$ direction draws as independent, which is conservative relative to treating all direction draws as independent.

\section{Reference implementation of the spectrum estimator} \label{app:code}

Definition~\ref{def:spectrum} is deployed through the estimator reproduced below: the same code path that App.~\ref{sec:exp-besovequiv} validates against closed-form norms, that App.~\ref{sec:exp-reliability} re-measures under independent probe seeds, and that produces every spectrum reported in this paper. The estimator is given as source because the conventions a prose description leaves open are exactly the details a reimplementation has to get right: where the radius jitter is applied, which axis carries the $p$-moment, at what point the $s$-weighting enters, and in what precision the probed function is evaluated. Listing~\ref{lst:estimator} is the model-agnostic core and Listing~\ref{lst:margin} the vision margin instantiation of Sec.~\ref{sec:method}; both are the deployed functions with the package qualifiers dropped, the docstrings shortened, the helpers named in the captions elided, and the vision entry point renamed, so that the two read as one file. The quantization-noise constructions of Eqs.~\ref{eq:bquant} and~\ref{eq:dquant} share the scale--direction--representation indexing but are not literal instances of Definition~\ref{def:spectrum}; the measurement protocol for these quantization scores is specified in App.~\ref{app:measurement} instead.

\begin{lstlisting}[caption={The estimator of Definition~\ref{def:spectrum}. \texttt{fd\_amplitude} realizes one scale of \eqref{eq:spectrum} from the finite-difference stencil of \eqref{eq:fd}; \texttt{spectrum} and \texttt{aggregate} form $B_j$ and $R^{(m)}_{s,p,q}$ from stored amplitudes.}, label={lst:estimator}]
import math
import numpy as np
import torch

# Stencil of Eq. 1:
#   Delta^m_h g(z) = sum_a (-1)^{m-a} binom(m,a) g(z + a h).
FD_COEF = {1: [-1.0, 1.0], 2: [1.0, -2.0, 1.0]}


def dyadic_radii(r0, n_scales):
    """The grid r_j = r0 * 2^{-j} for j from 0 through n_scales - 1."""
    return [r0 * 2.0 ** (-j) for j in range(n_scales)]


def log_jitter(n, device, generator=None, ndim=1):
    """Per-sample log-uniform xi in [2^{-1/2}, 2^{1/2}], multiplied
    into the direction, so the step obeys ||h|| = xi * r_j."""
    f = torch.exp2(
        torch.rand(n, device=device, generator=generator) - 0.5)
    return f.view(n, *([1] * (ndim - 1)))


@torch.no_grad()
def fd_amplitude(eval_g, z, u, r, order=2, p=2):
    """One scale of Eq. 2, from m + 1 forwards and no backward:
    A = (E_z ||Delta^m_{r u} g(z)||^p)^{1/p}.  eval_g maps a tensor
    shaped like the base points z to one row per base point; u
    carries the jitter; the norm is Euclidean within a row."""
    coefs = FD_COEF[order]
    acc = None
    for a, c in enumerate(coefs):
        out = eval_g(z + a * r * u)
        val = c * out.reshape(out.shape[0], -1).float()
        acc = val if acc is None else acc + val
    norms = acc.norm(dim=1)
    return norms.pow(p).mean().pow(1.0 / p).item()


def spectrum(amps, radii, s):
    """B_j = r_j^{-s} A_j.  The raw A_j are what gets stored; s
    enters here, at analysis time, so the smoothness index is
    swept without re-measuring."""
    return [a * (r ** (-s)) for a, r in zip(amps, radii)]


def aggregate(amps, radii, s, q=2):
    """R^{(m)}_{s,p,q} = (sum_j B_j^q)^{1/q} on the finite grid,
    read as a supremum at q = inf."""
    B = spectrum(amps, radii, s)
    if math.isinf(q):
        return max(B)
    return float(np.sum(np.power(B, q)) ** (1.0 / q))
\end{lstlisting}

Four conventions are fixed by this code rather than by \eqref{eq:spectrum} alone. The jitter multiplies the direction, not the radius argument, so the realized step is $\norm{h}=\xi r_j$. For each scale and direction draw, an independent jitter is sampled for every base point and shared across the stencil points associated with that base point. The $p$-moment is taken after the Euclidean norm is contracted over the output axis of a row, so vector-valued and scalar targets are handled by the same estimator and the moment never mixes base points with output coordinates. The $s$-weighting is applied by \texttt{spectrum} at analysis time, which is what makes the stored $A_j$ and not $B_j$ the primitive, as Sec.~\ref{sec:method} states; the stored record adds the aggregates at $s\in\{0.5,1,1.5\}$ and the log--log slope of $A_j$ against $r_j$, every one of those derived quantities formed from the stored amplitudes. Finite-difference accumulation uses \texttt{float32}, through the \texttt{.float()} cast applied to the outputs of \texttt{eval\_g}. In the vision implementation, \texttt{margin\_eval\_fn} also evaluates the margin in \texttt{float32}. A second difference at the finest radius is a near-canceling sum of three comparable numbers, and at $m=2$ the surviving digits fall below \texttt{bfloat16} resolution. App.~\ref{sec:exp-besovequiv} reports the residual float32-against-float64 agreement of this accumulation on probe functions whose amplitudes are known in closed form.

\Needspace{36\baselineskip}
\begin{lstlisting}[caption={The vision margin instantiation of Sec.~\ref{sec:method}: $g=m_\theta$ as a function of the input, at $m=2$, $p=2$, over eight dyadic radii and four direction families. \texttt{sample\_dirs} draws the unit-norm families of App.~\ref{app:measurement}, and \texttt{margin\_eval\_fn} evaluates $m_\theta$ in \texttt{float32}.}, label={lst:margin}]
RADII = dyadic_radii(16.0, 8)   # L2 radii, normalized inputs
FAMILIES = ["iso", "highfreq", "lowfreq", "aug"]


@torch.no_grad()
def margin_spectra(model, x, y, n=1024, k_dirs=4, seed=0):
    """Margin second-difference amplitudes per (family, scale).

    margin_eval_fn wraps the model as g = m_theta of the input and
    evaluates the margin in fp32.  The n base points are a fixed
    prefix of the test set, identical across checkpoints; only the
    directions and the radius jitter are redrawn per probe seed.
    """
    gen = torch.Generator(device=x.device).manual_seed(seed)
    xs, ys = x[:n], y[:n]
    f = margin_eval_fn(model, ys)
    out = {}
    for fam in FAMILIES:
        amps = []
        for r in RADII:
            vals = []
            for _ in range(k_dirs):
                # sample_dirs: unit L2 norm per sample
                u = sample_dirs(xs, fam, gen)
                u = u * log_jitter(
                    len(xs), x.device, gen, ndim=u.dim())
                vals.append(
                    fd_amplitude(f, xs, u, r, order=2, p=2))
            # The k_dirs draws are pooled in quadrature: the
            # p = 2 case of Eq. 2.
            amps.append(
                float(torch.tensor(vals).pow(2).mean().sqrt()))
        out[fam] = amps    # raw A_j, one curve per family
    return out
\end{lstlisting}

The direction draws are pooled across the $K=4$ draws in quadrature rather than arithmetically. At the deployed $p=2$, that pooling is an identity, not an approximation: Each \texttt{fd\_amplitude} call returns $(\sum_i\norm{\cdot}^2/n)^{1/2}$ over the shared base points, so the root-mean-square over draws is exactly the single-stage estimator $(\sum_{i,k}\norm{\cdot}^2/(nK))^{1/2}$ of App.~\ref{app:measurement}. The two stages would differ at $p\ne2$, and no result in this paper uses one. Because the $K$ draws share the $n$ base points, the base points and not the $nK$ draws are the sampling units, which is the clustering convention App.~\ref{app:estvar} adopts for every interval reported on these spectra.

\section{Validating the canonical estimator against closed-form norms} \label{sec:exp-besovequiv}

Theorem~\ref{thm:besov-equiv} says that under (A1) through (A3) the estimator's population target is an equivalent Besov (quasi-)seminorm. That claim is falsifiable with no reference to networks: We run probe functions whose predicted in-band scaling, sharp regularity, or chosen coefficient seminorm is known, through the very code path that every spectrum in this paper uses, with the same finite-difference amplitude estimator, log-uniform jitter, dyadic grid, and $m=2$, and compare. Table~\ref{tab:besovequiv} runs four families on the circle.  Finite truncations of the ideal lacunary series $g_\alpha\coloneqq\sum_k 2^{-k\alpha}\cos(2\pi 2^k x+\varphi_k)$ place one frequency per Littlewood--Paley block. The target in-band scaling exponent of these lacunary truncations is $\alpha$, and a fixed dyadic Littlewood--Paley coefficient seminorm is closed form, $\mathcal N_{\mathrm{LP}}(g_\alpha) = \norm{\cos}_{L^p} (\sum_k 2^{k(s-\alpha)q})^{1/q}$, read as a supremum for $q=\infty$; the ideal infinite random Fourier series of fractional Brownian motion (fBm) type has critical index $H$, while the finite fBm synthesis targets the corresponding in-band exponent; a square wave has the sharp index $1/p$, which depends on $p$, together with a fully predicted amplitude in the square wave's single-crossing fine-scale window, $A_j = J_h(2\rho\E[\xi]r_j)^{1/p}$ for jump density $\rho$, jump height $J_h$, and radius jitter $\xi$; a smooth control saturates at the stencil order.

Three observables validate the estimator--theory connection in this canonical regime, beginning with the two-sided equivalence interval. Across all $108$ $(\alpha,\mathrm{window},s,p,q)$ lacunary cells, the measured-to-exact ratio stays inside $[2.6,10.9]$, an interval of width $4.2\times$ and at most $3.3\times$ within any fixed $(p,q)$, while the exact seminorm sweeps a $482\times$ range: The estimator tracks the ambient norm two-sidedly with constants that move two orders of magnitude less than the norms being estimated, which is what norm equivalence means and all that equivalence promises. The second observable is the pair of in-band exponents and constants. Interior windows, defined as windows with spectral mass on both sides of every probed scale, recover the target exponent to $0.004$; grid-edge windows bias the fitted slope by up to $0.086$, which is the truncation effect of Proposition~\ref{prop:trunc} measured directly; slope recovery holds for the fBm-type series to $0.024$; the square wave matches the predicted exponent $1/p$ to $0.008$ and the predicted leading constant to $0.993$ to $0.998$ across $p\in\{1,2,4\}$, exponent and constant alike, neither quantity a fit; and the smooth control saturates at $1.987$ against the stencil order $m=2$. Numerical agreement is the third. The float32 path agrees with a float64 oracle to relative $6.3\times10^{-7}$ over every cell used above, while the finest smooth-control scale shows a $4\%$ float32 deviation: The cancellation floor is expected, detected, and excluded by the fitting window, applying the precision constraint identified in App.~\ref{sec:exp-spectrum}.

The exact periodic check above is one-dimensional. As a separate validation in two dimensions, we also sample $d=2$ with isotropic $u\sim\mathrm{Unif}(S^1)$ on randomly oriented planar lacunary waves over the unit square. The random wave vectors of these planar waves are generally nonlattice, so these rows are not periodic torus characters and the same coefficient aggregate $\mathcal N_{\mathrm{LP}}$ is a reference quantity rather than an exact periodic Besov (quasi-)seminorm.  We therefore use the slope as the mathematical check: That slope recovers the target in-band exponent to $0.006$ on interior windows.  The descriptive reference ratios keep all $36$ two-dimensional ratio cells inside $[2.4,6.9]$, comparable to but not contained in the one-dimensional interval.

\begin{table}[t!] \centering \caption{Validation of the estimator against known Besov structure. For each construction, the table compares the predicted exponent $s^\star$ with the fitted log--log slope $\hat s$ and, where available, a Littlewood--Paley coefficient reference $\mathcal N_{\mathrm{LP}}$. Interior windows test the in-band scaling; grid-edge windows expose finite-grid truncation. The 2-D coefficient aggregate is a reference rather than an exact periodic seminorm. NA indicates that the coefficient reference is not defined for that row.} \label{tab:besovequiv} \small \begin{tabular}{l c c c c} \toprule construction & target $s^\star$ & slope $\hat s$ & $|\hat s-s^\star|$ & $\widehat R/\mathcal N_{\mathrm{LP}}$ \\ \midrule lacunary (interior), $\alpha=0.3$ & $0.3$ & 0.304 & 0.004 & 5.57 \\ lacunary (interior), $\alpha=0.5$ & $0.5$ & 0.500 & 0.000 & 6.22 \\ lacunary (interior), $\alpha=0.7$ & $0.7$ & 0.699 & 0.001 & 7.46 \\ \addlinespace lacunary (grid-edge), $\alpha=0.3$ & $0.3$ & 0.276 & 0.024 & 6.38 \\ lacunary (grid-edge), $\alpha=0.5$ & $0.5$ & 0.442 & 0.058 & 7.14 \\ lacunary (grid-edge), $\alpha=0.7$ & $0.7$ & 0.614 & 0.086 & 8.51 \\ \midrule lacunary 2-D (interior), $\alpha=0.3$ & $0.3$ & 0.306 & 0.006 & 4.74 \\ lacunary 2-D (interior), $\alpha=0.5$ & $0.5$ & 0.501 & 0.001 & 4.95 \\ lacunary 2-D (interior), $\alpha=0.7$ & $0.7$ & 0.699 & 0.001 & 5.62 \\ \midrule fBm-type, $H=0.3$ & $0.3$ & 0.305 & 0.005 & NA \\ fBm-type, $H=0.5$ & $0.5$ & 0.488 & 0.012 & NA \\ fBm-type, $H=0.7$ & $0.7$ & 0.724 & 0.024 & NA \\ \midrule square wave, $p=1$ & $1/p=1.00$ & 1.008 & 0.008 & NA \\ square wave, $p=2$ & $1/p=0.50$ & 0.504 & 0.004 & NA \\ square wave, $p=4$ & $1/p=0.25$ & 0.252 & 0.002 & NA \\ \midrule smooth $\cos(2\pi x)$ & $m=2$ (sat.) & 1.987 & 0.013 & NA \\ \bottomrule \end{tabular} \end{table}

\section{Additional results} \label{app:extra}

\subsection{Mechanism: finite-scale distinctions missed by scalar proxies} \label{sec:exp-toy}

\begin{figure}[t!] \centering \includegraphics[width=\linewidth]{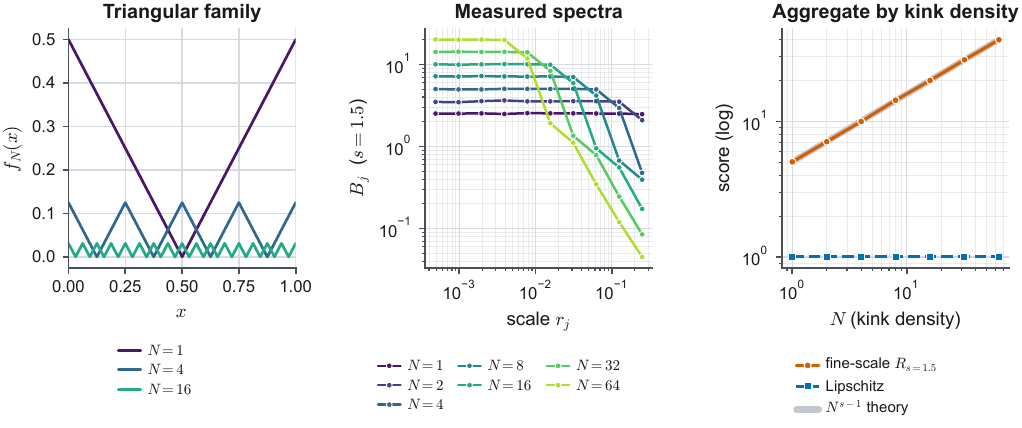} \caption{The blind spot shared by the Hessian and the Lipschitz constant. Left: $f_N(x)\coloneqq\tau(Nx)/N$ for $N\in\{1,4,16\}$; all members have Lipschitz constant 1 and zero curvature almost everywhere. Middle: Measured spectra $B_j$ at $s=1.5$ localize the kink scale. Right: The fixed-grid spectrum aggregate grows as $N^{0.500}$, matching the finite-grid kink law $N^{s-1}=N^{0.5}$, while the Lipschitz constant is flat.} \label{fig:toy} \end{figure}

We first verify the finite-grid kink law underlying Theorem~\ref{thm:kink} on the theorem's blind-spot construction. For $f_N(x) \coloneqq \tau(Nx)/N$, $N \in \{1,2,\dots,64\}$, the empirical Lipschitz constant, a maximum over $2\times10^{5}$ grid pairs, is $1.000$ for every $N$, and the second derivative is zero almost everywhere: Both scalar proxies are constant across the family. We use $s=1.5$, $p=2$, $m=2$, the critical exponent $s=1+1/p$. Theorem~\ref{thm:kink} gives $f_N\in B^{3/2}_{2,\infty}$ with seminorm of order $N^{1/2}$, and $f_N\notin B^{3/2}_{2,q}$ for finite $q$. The resolved fine-scale profile of $f_N$ has $B_j\asymp N^{1/2}$; the $q=2$ readout aggregates this critical profile over a fixed finite band, as in \eqref{eq:critical-band-intensity}. The measured ten-scale spectrum separates the family monotonically: The aggregate over the four finest scales grows from $5.03$ at $N=1$ to $40.1$ at $N=64$, and a log--log fit of this fine-scale aggregate against $N$ gives an empirical exponent of $0.500$, matching that finite-grid prediction $N^{s-1}=N^{0.5}$ (Figure~\ref{fig:toy}). The scale-resolved curves additionally localize each $f_N$'s kink spacing: $B_j$ transitions from the saturation regime to the kink-dominated regime at $r_j \approx 1/N$, information that no scalar summary retains. Per-sample radius jitter, $\norm{h}\asymp r_j$ as specified in App.~\ref{app:measurement}, is essential here: Exact dyadic radii resonate with the periodic kink grid of dyadic $N$ and would silently zero out entire scales, a useful reminder that the classical $\norm{h}\asymp 2^{-j}$ convention is not a technicality.

\FloatBarrier

\subsection{Auxiliary activation-crossing penalty} \label{app:crossing-penalty}

RSC is the default vision intervention. We also evaluated an auxiliary activation-crossing penalty that encourages agreement of ReLU pre-activation signs along the same nuisance probes. The combined variant, RSC $+$ crossing, uses the objective \begin{equation} \mathcal L_{\mathrm{task}}+\lambda_{\mathrm{RSC}}\mathcal L_{\mathrm{RSC}} +\lambda_{\mathrm{cross}}\mathcal L_{\mathrm{cross}}, \end{equation} where \begin{equation} \mathcal L_{\mathrm{cross}}\coloneqq \E\left[\mathrm{mean}_{\ell,k} \sigma\left(-\frac{a_{\ell,k}(x)a_{\ell,k}(x+r_j u)}{\tau_\ell^2}\right)\right]. \label{eq:lcross} \end{equation} Here $\sigma$ is the logistic sigmoid, $a_{\ell,k}$ are ReLU pre-activations, and $\tau_\ell$ is the detached mean absolute value of these pre-activations on the unperturbed batch, floored at $10^{-6}$, as in the implementation. The auxiliary term uses every other ReLU module and $\lambda_{\mathrm{cross}}=0.1$, with $\lambda_{\mathrm{RSC}}=0.05$ and the same radius and direction sampling as RSC. These weights were fixed without a tuning sweep. This variant was evaluated only in the two ResNet-18 pools, with three training seeds each; no cross-only training arm is reported.

The surrogate is motivated by the affine-region interpretation of second differences, but does not weight a unit's sign change by the effect of that sign change on the output margin. Lemma~\ref{lem:boundary-local} involves the directional derivative jump of the probed function, not only the occurrence of a crossing. Moreover, changes of the maximal rival class can contribute to margin second differences without a ReLU sign change. We therefore make no claim that minimizing this surrogate bounds margin roughness or is required for the RSC intervention.

Tables~\ref{tab:vision-c10} and~\ref{tab:vision-c100} retain the complete comparison. Adding the crossing term does not improve any of the five reported mean accuracy endpoints in either ResNet pool: Noise-corruption accuracy changes from $71.19\%$ to $70.69\%$ on CIFAR-10 and from $45.30\%$ to $44.98\%$ on CIFAR-100; Gaussian accuracy changes from $82.46\%$ to $82.25\%$ and from $51.86\%$ to $50.90\%$, respectively. The reported HF roughness rounds to the same values for the two methods, $0.06$ and $0.13$. The three-seed comparison provides no evidence of an added robustness benefit at the evaluated weight; the three-seed comparison does not resolve very small effects or the performance of other weights. RSC $+$ crossing is consequently reported as an auxiliary ablation; the WRN and ViT transfer experiments use RSC alone.

\subsection{Vision pools: full tables and intervention controls} \label{app:vision-pools}

\begin{figure}[t!] \centering \includegraphics[width=\linewidth]{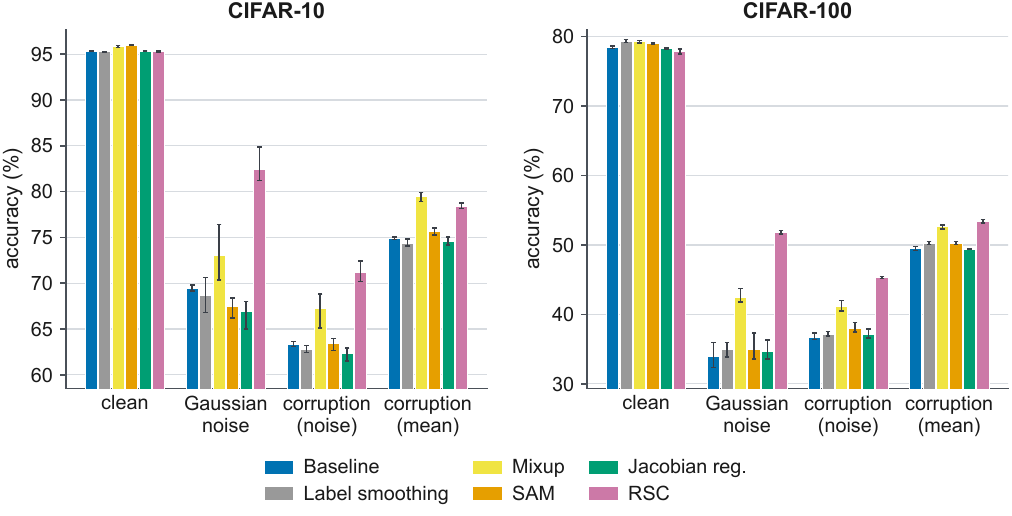} \caption{RSC and comparison recipes. Clean accuracy, Gaussian-noise accuracy at $\sigma=0.05$, and noise-family and mean corruption accuracy; bars show means and whiskers span the available seeds. Auxiliary crossing results and int4-PTQ outcomes are reported in App.~\ref{app:vision-pools}.} \label{fig:vision-int} \end{figure}

This subsection presents the four checkpoint pools of Sec.~\ref{sec:exp-vision}, namely CIFAR-10, CIFAR-100, WRN-28-10, and ViT-S/4, in a common format, together with the intervention results and controls. Figure~\ref{fig:vision-int} summarizes the two ResNet pools. Estimator-reliability figures follow in App.~\ref{sec:exp-reliability}. The ViT-S/4 pool spans two seeds of every knob, $n=12$ across baseline, label smoothing, mixup, SAM, no stochastic depth, and RSC, and RSC's clean gain there is $91.44\pm0.20$ against a baseline $90.31\pm0.22$.

\paragraph{Compute and BatchNorm controls.} At matched wall-clock on CIFAR-100, where RSC trains for $200$ epochs and SAM for $300$, RSC exceeds SAM by $6.5$ noise-corruption and $15.2$ Gaussian points, with clean accuracy $77.90\%$ against $79.26\%$ (Table~\ref{tab:vision-c100}). Recalibrating BatchNorm on clean training data leaves RSC's noise-family advantage essentially intact: The advantage moves from $+7.9$ to $+6.0$ points on CIFAR-100 and from $+7.3$ to $+6.4$ on CIFAR-10, disfavoring running-statistics changes alone as the explanation.

\paragraph{The intervention transfers to the attention-based architecture.} On the ViT-S/4 pool (Table~\ref{tab:vision-vit}), RSC transfers without retuning, at the same $\lambda_{\mathrm{RSC}}$ as in the ResNet pools: $+6.5$ noise-corruption, $+8.4$ Gaussian, and $+3.4$ mean-corruption points over the baseline. Unlike on CIFAR-10 ResNet and WRN, where clean accuracy is matched, RSC also adds $+1.1$ points clean and exceeds SAM on every column at identical epochs. The margin diagnostic does not move on this pool: The ViT-S/4 baseline already sits at the pool floor of $\tilde R^{\mathrm{hf}}_{1.5}=0.06$, so this pool tests whether the intervention transfers, not whether the diagnostic moves.

\paragraph{Three qualifications on the intervention comparison.} GroupNorm achieves the best raw mean-corruption accuracy on CIFAR-10, $82.5\%$, at a $1.0\%$ clean-accuracy cost (Table~\ref{tab:vision-c10}), consistent with BatchNorm's known corruption-statistics fragility, so recipe choices remain strong levers; combining RSC with SAM preserves SAM's clean accuracy at $96.0\%$ while improving every robustness column over SAM alone, $65.6$ against $63.4$ noise and $73.5$ against $67.5$ Gaussian, so the two mechanisms compose additively rather than competing; and the SAM $\rho$-grid and the RSC $+$ SAM composition are unreplicated auxiliary arms, so no run-to-run spread attaches to those particular numbers, unlike the replicated primary comparisons (App.~\ref{app:setup-details}).

\begin{table}[t!] \centering \caption{CIFAR-10 checkpoint pool. Here and throughout the replicated vision results, entries give mean $\pm$ half the range across training seeds, where half the range is the maximum minus the minimum, divided by two. Bare entries are unreplicated auxiliary arms. Baseline corruption columns use two available seeds. RSC $+$ crossing is the auxiliary crossing ablation of App.~\ref{app:crossing-penalty}. Accuracy columns show clean, CIFAR-10-C, noise-family corruption, int4 PTQ, and Gaussian-noise performance. Diagnostics are HF margin roughness, the signed dominant-magnitude Hessian eigenvalue, and relative empirical Lipschitz slope.} \label{tab:vision-c10} \small \begin{tabular}{l ccccc} \toprule & \multicolumn{5}{c}{Accuracy: clean and under perturbations} \\ \cmidrule(lr){2-6} Method & Clean & Corr. & Corr.(noise) & int4 & Gauss \\ \midrule Baseline & 95.31 $\pm$0.03 & 74.92 $\pm$0.12 & 63.36 $\pm$0.27 & 94.88 $\pm$0.15 & 69.39 $\pm$0.31 \\ No weight decay & 92.47 $\pm$0.37 & 72.27 $\pm$0.26 & 66.99 $\pm$1.89 & 91.49 $\pm$0.53 & 76.26 $\pm$2.47 \\ Label smoothing & 95.28 $\pm$0.02 & 74.34 $\pm$0.37 & 62.76 $\pm$0.38 & 94.98 $\pm$0.17 & 68.69 $\pm$1.92 \\ Mixup & 95.88 $\pm$0.12 & 79.46 $\pm$0.49 & 67.22 $\pm$1.85 & 95.39 $\pm$0.15 & 73.06 $\pm$3.03 \\ GroupNorm & 94.30 $\pm$0.17 & 82.48 $\pm$0.19 & 74.14 $\pm$0.25 & 94.09 $\pm$0.11 & 78.63 $\pm$0.27 \\ Learning rate 0.3 & 94.74 $\pm$0.09 & 73.21 $\pm$0.60 & 62.72 $\pm$1.11 & 94.11 $\pm$0.13 & 68.46 $\pm$2.49 \\ SAM & 95.99 $\pm$0.05 & 75.62 $\pm$0.38 & 63.41 $\pm$0.65 & 95.57 $\pm$0.01 & 67.47 $\pm$1.11 \\ SAM $\rho=0.1$ & 96.07 & 75.46 & 62.65 & 95.85 & 67.77 \\ SAM $\rho=0.2$ & 96.21 & 75.84 & 63.36 & 95.86 & 65.43 \\ Jacobian reg. & 95.36 $\pm$0.05 & 74.53 $\pm$0.42 & 62.34 $\pm$0.70 & 95.09 $\pm$0.15 & 66.89 $\pm$1.51 \\ \addlinespace RSC (default) & 95.30 $\pm$0.07 & 78.42 $\pm$0.29 & 71.19 $\pm$1.13 & 95.04 $\pm$0.07 & 82.46 $\pm$1.81 \\ RSC $+$ crossing (aux.) & 95.22 $\pm$0.18 & 77.85 $\pm$0.43 & 70.69 $\pm$0.52 & 94.93 $\pm$0.25 & 82.25 $\pm$1.19 \\ RSC $+$ SAM (ours) & 96.00 & 76.35 & 65.60 & 95.83 & 73.49 \\ \bottomrule \end{tabular}

\vspace{0.7em} \begin{tabular}{l ccc} \toprule & \multicolumn{3}{c}{Diagnostics on clean data} \\ \cmidrule(lr){2-4} Method & $\tilde R^{\mathrm{hf}}_{1.5}$ & $\lambda_{\mathrm{dom}}$ & $\widetilde\Lip$ \\ \midrule Baseline & 0.18 $\pm$0.02 & 14.30 $\pm$12.80 & 10.4 $\pm$0.9 \\ No weight decay & 0.10 $\pm$0.01 & 0.94 $\pm$0.48 & 4.7 $\pm$0.7 \\ Label smoothing & 0.22 $\pm$0.04 & 27.16 $\pm$2.96 & 16.9 $\pm$2.6 \\ Mixup & 0.37 $\pm$0.04 & 301.80 $\pm$6.69 & 16.6 $\pm$1.1 \\ GroupNorm & 0.23 $\pm$0.02 & 27.67 $\pm$12.28 & 20.7 $\pm$1.9 \\ Learning rate 0.3 & 0.21 $\pm$0.02 & 38.65 $\pm$24.85 & 11.0 $\pm$0.5 \\ SAM & 0.15 $\pm$0.01 & 3.03 $\pm$0.99 & 6.3 $\pm$0.3 \\ SAM $\rho=0.1$ & 0.17 & 4.21 & 6.5 \\ SAM $\rho=0.2$ & 0.17 & 13.07 & 6.9 \\ Jacobian reg. & 0.17 $\pm$0.00 & 6.81 $\pm$2.89 & 10.2 $\pm$0.5 \\ \addlinespace RSC (default) & 0.06 $\pm$0.00 & 41.38 $\pm$42.48 & 7.4 $\pm$1.2 \\ RSC $+$ crossing (aux.) & 0.06 $\pm$0.00 & 6.49 $\pm$2.70 & 5.9 $\pm$0.8 \\ RSC $+$ SAM (ours) & 0.09 & 2.56 & 5.4 \\ \bottomrule \end{tabular} \end{table}

\begin{table}[t!] \centering \caption{CIFAR-100 checkpoint pool, columns and reporting conventions as in Table~\ref{tab:vision-c10}.} \label{tab:vision-c100} \small \fitwidth{%
\begin{tabular}{l ccccc} \toprule & \multicolumn{5}{c}{Accuracy: clean and under perturbations} \\ \cmidrule(lr){2-6} Method & Clean & Corr. & Corr.(noise) & int4 & Gauss \\ \midrule Baseline & 78.46 $\pm$0.19 & 49.53 $\pm$0.28 & 36.72 $\pm$0.45 & 77.50 $\pm$0.15 & 34.00 $\pm$1.79 \\ Label smoothing & 79.35 $\pm$0.18 & 50.25 $\pm$0.22 & 37.13 $\pm$0.33 & 78.45 $\pm$0.63 & 34.91 $\pm$1.03 \\ Mixup & 79.24 $\pm$0.17 & 52.63 $\pm$0.29 & 41.22 $\pm$0.74 & 77.82 $\pm$0.20 & 42.50 $\pm$0.96 \\ SAM & 78.97 $\pm$0.11 & 50.28 $\pm$0.23 & 38.03 $\pm$0.69 & 77.99 $\pm$0.09 & 34.92 $\pm$1.86 \\ SAM (wall-clock-matched) & 79.26 $\pm$0.10 & 50.64 $\pm$0.18 & 38.77 $\pm$0.16 & 78.35 $\pm$0.09 & 36.70 $\pm$0.90 \\ Jacobian reg. & 78.30 $\pm$0.10 & 49.42 $\pm$0.05 & 37.17 $\pm$0.63 & 76.90 $\pm$0.41 & 34.61 $\pm$1.37 \\ \addlinespace RSC (default) & 77.90 $\pm$0.39 & 53.40 $\pm$0.25 & 45.30 $\pm$0.13 & 76.89 $\pm$0.63 & 51.86 $\pm$0.24 \\ RSC $+$ crossing (aux.) & 77.79 $\pm$0.13 & 52.99 $\pm$0.36 & 44.98 $\pm$0.53 & 76.55 $\pm$0.51 & 50.90 $\pm$1.66 \\ \bottomrule \end{tabular}}

\vspace{0.7em} \begin{tabular}{l ccc} \toprule & \multicolumn{3}{c}{Diagnostics on clean data} \\ \cmidrule(lr){2-4} Method & $\tilde R^{\mathrm{hf}}_{1.5}$ & $\lambda_{\mathrm{dom}}$ & $\widetilde\Lip$ \\ \midrule Baseline & 0.48 $\pm$0.05 & 12.39 $\pm$2.85 & 12.4 $\pm$1.2 \\ Label smoothing & 0.51 $\pm$0.05 & 29.28 $\pm$1.81 & 14.7 $\pm$2.2 \\ Mixup & 0.51 $\pm$0.01 & 46.70 $\pm$7.37 & 14.0 $\pm$0.9 \\ SAM & 0.47 $\pm$0.04 & 4.85 $\pm$0.52 & 10.5 $\pm$0.4 \\ SAM (wall-clock-matched) & 0.48 $\pm$0.01 & 4.28 $\pm$0.01 & 11.3 $\pm$0.0 \\ Jacobian reg. & 0.51 $\pm$0.02 & 11.15 $\pm$1.07 & 12.6 $\pm$1.1 \\ \addlinespace RSC (default) & 0.13 $\pm$0.00 & 23.61 $\pm$5.17 & 6.4 $\pm$0.4 \\ RSC $+$ crossing (aux.) & 0.13 $\pm$0.01 & 27.68 $\pm$6.18 & 6.1 $\pm$0.7 \\ \bottomrule \end{tabular} \end{table}

\begin{table}[t!] \centering \caption{WRN-28-10 pool (CIFAR-10), second architecture, identical recipe, columns, and reporting conventions as in Table~\ref{tab:vision-c10}.} \label{tab:vision-wrn} \small \begin{tabular}{l ccccc} \toprule & \multicolumn{5}{c}{Accuracy: clean and under perturbations} \\ \cmidrule(lr){2-6} Method & Clean & Corr. & Corr.(noise) & int4 & Gauss \\ \midrule Baseline & 96.25 $\pm$0.04 & 76.20 $\pm$0.25 & 63.42 $\pm$0.70 & 96.08 $\pm$0.11 & 66.68 $\pm$0.94 \\ No weight decay & 94.03 $\pm$0.05 & 72.72 $\pm$0.20 & 63.30 $\pm$0.91 & 93.79 $\pm$0.01 & 69.56 $\pm$1.30 \\ Label smoothing & 96.01 $\pm$0.03 & 75.58 $\pm$0.23 & 63.48 $\pm$0.94 & 95.68 $\pm$0.02 & 67.28 $\pm$2.33 \\ Mixup & 96.71 $\pm$0.07 & 80.98 $\pm$0.07 & 68.50 $\pm$0.88 & 96.45 $\pm$0.03 & 74.26 $\pm$1.45 \\ SAM & 96.86 $\pm$0.01 & 77.56 $\pm$0.30 & 65.30 $\pm$0.47 & 96.64 $\pm$0.02 & 69.25 $\pm$0.95 \\ \addlinespace RSC (default) & 96.12 $\pm$0.00 & 80.51 $\pm$0.37 & 73.66 $\pm$0.75 & 96.11 $\pm$0.02 & 84.17 $\pm$0.69 \\ \bottomrule \end{tabular}

\vspace{0.7em} \begin{tabular}{l ccc} \toprule & \multicolumn{3}{c}{Diagnostics on clean data} \\ \cmidrule(lr){2-4} Method & $\tilde R^{\mathrm{hf}}_{1.5}$ & $\lambda_{\mathrm{dom}}$ & $\widetilde\Lip$ \\ \midrule Baseline & 0.20 $\pm$0.01 & 9.71 $\pm$4.97 & 16.4 $\pm$0.6 \\ No weight decay & 0.16 $\pm$0.02 & 0.83 $\pm$0.63 & 7.5 $\pm$1.7 \\ Label smoothing & 0.24 $\pm$0.01 & 93.42 $\pm$2.43 & 24.1 $\pm$2.2 \\ Mixup & 0.40 $\pm$0.03 & 298.31 $\pm$71.37 & 25.9 $\pm$0.2 \\ SAM & 0.19 $\pm$0.00 & 1.76 $\pm$0.12 & 8.3 $\pm$0.9 \\ \addlinespace RSC (default) & 0.05 $\pm$0.00 & 8.35 $\pm$3.64 & 6.5 $\pm$0.3 \\ \bottomrule \end{tabular} \end{table}

\begin{table}[t!] \centering \caption{ViT-S/4 pool (CIFAR-10), third (attention-based) architecture, DeiT-style 300-epoch recipe; columns and reporting conventions as in Table~\ref{tab:vision-c10}.} \label{tab:vision-vit} \small \begin{tabular}{l ccccc} \toprule & \multicolumn{5}{c}{Accuracy: clean and under perturbations} \\ \cmidrule(lr){2-6} Method & Clean & Corr. & Corr.(noise) & int4 & Gauss \\ \midrule Baseline & 90.31 $\pm$0.22 & 79.00 $\pm$0.34 & 73.01 $\pm$0.20 & 89.85 $\pm$0.26 & 77.28 $\pm$0.33 \\ Label smoothing & 90.02 $\pm$0.04 & 78.36 $\pm$0.03 & 72.40 $\pm$0.06 & 89.74 $\pm$0.14 & 77.07 $\pm$0.40 \\ Mixup & 89.68 $\pm$1.00 & 77.75 $\pm$1.27 & 71.84 $\pm$1.26 & 89.21 $\pm$0.98 & 77.14 $\pm$1.00 \\ No stochastic depth & 89.21 $\pm$0.23 & 77.66 $\pm$0.23 & 71.83 $\pm$0.24 & 89.03 $\pm$0.02 & 75.86 $\pm$0.73 \\ SAM & 88.53 $\pm$0.82 & 77.08 $\pm$0.85 & 71.72 $\pm$0.42 & 88.35 $\pm$0.74 & 76.38 $\pm$0.41 \\ \addlinespace RSC (default) & 91.44 $\pm$0.20 & 82.35 $\pm$0.19 & 79.50 $\pm$0.54 & 91.12 $\pm$0.07 & 85.72 $\pm$0.13 \\ \bottomrule \end{tabular}

\vspace{0.7em} \begin{tabular}{l ccc} \toprule & \multicolumn{3}{c}{Diagnostics on clean data} \\ \cmidrule(lr){2-4} Method & $\tilde R^{\mathrm{hf}}_{1.5}$ & $\lambda_{\mathrm{dom}}$ & $\widetilde\Lip$ \\ \midrule Baseline & 0.06 $\pm$0.00 & 0.02 $\pm$0.01 & 4.4 $\pm$0.1 \\ Label smoothing & 0.11 $\pm$0.00 & 2.16 $\pm$0.38 & 18.8 $\pm$3.4 \\ Mixup & 0.10 $\pm$0.00 & 23.64 $\pm$2.53 & 8.2 $\pm$0.0 \\ No stochastic depth & 0.07 $\pm$0.00 & 0.25 $\pm$0.17 & 4.5 $\pm$0.4 \\ SAM & 0.06 $\pm$0.00 & 0.03 $\pm$0.01 & 3.9 $\pm$0.1 \\ \addlinespace RSC (default) & 0.06 $\pm$0.00 & 0.50 $\pm$0.03 & 10.9 $\pm$1.3 \\ \bottomrule \end{tabular} \end{table}

\subsection{Measurement reliability of the spectrum estimator} \label{sec:exp-reliability}

\begin{figure}[t!] \centering \includegraphics[width=\linewidth]{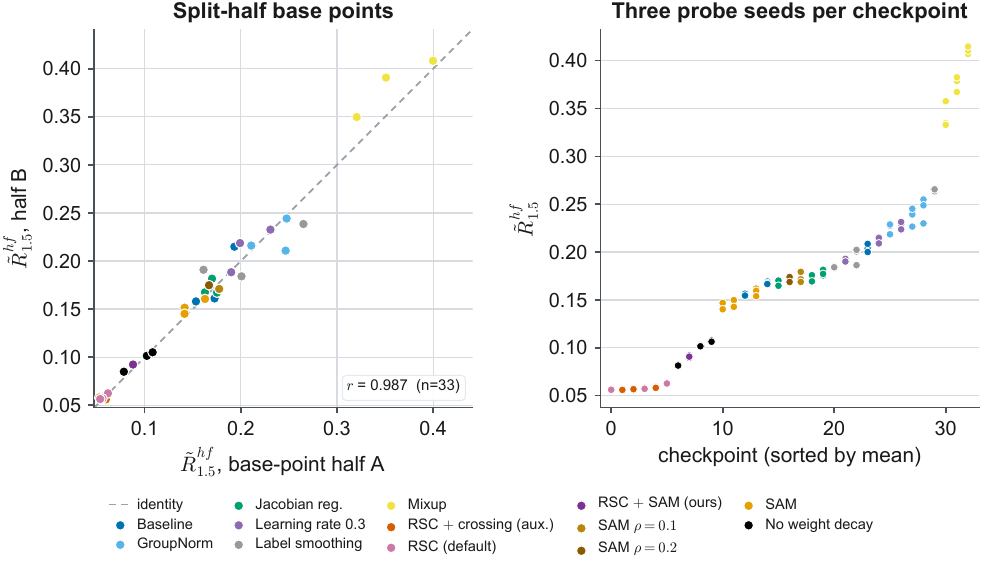} \caption{Measurement noise is far below model differences. Left: the CIFAR-10 primary predictor re-estimated on two disjoint base-point halves; each point is a checkpoint, colored by the checkpoint's training method. Right: the same predictor under three independent probe seeds per checkpoint, with checkpoints sorted by mean; the vertical scatter, which is probe noise, is invisible at the scale of the between-checkpoint spread.} \label{fig:reliability} \end{figure}

For the four trained vision pools, the probe-seed test-retest ICC(1) of the primary predictor is $0.997$ to $0.999$, and the within-checkpoint probe noise is $25$ to $48\times$ smaller than the between-checkpoint spread; split-half correlations over disjoint base-point sets are $0.962$ to $0.987$, rising to $0.980$ to $0.993$ after Spearman--Brown correction. Measurement noise is negligible relative to the observed between-checkpoint differences over the measured finite band (Figure~\ref{fig:reliability}). App.~\ref{app:measurement} specifies the protocol. Relative accuracy at shrinking radii requires separate sampling control (Corollary~\ref{cor:statistical-resolution}).

\FloatBarrier

\subsection{Allocation comparisons: uncertainty, additivity, and quantizer transfer} \label{app:alloc-extra}

\begin{figure}[t!] \centering \includegraphics[width=\linewidth]{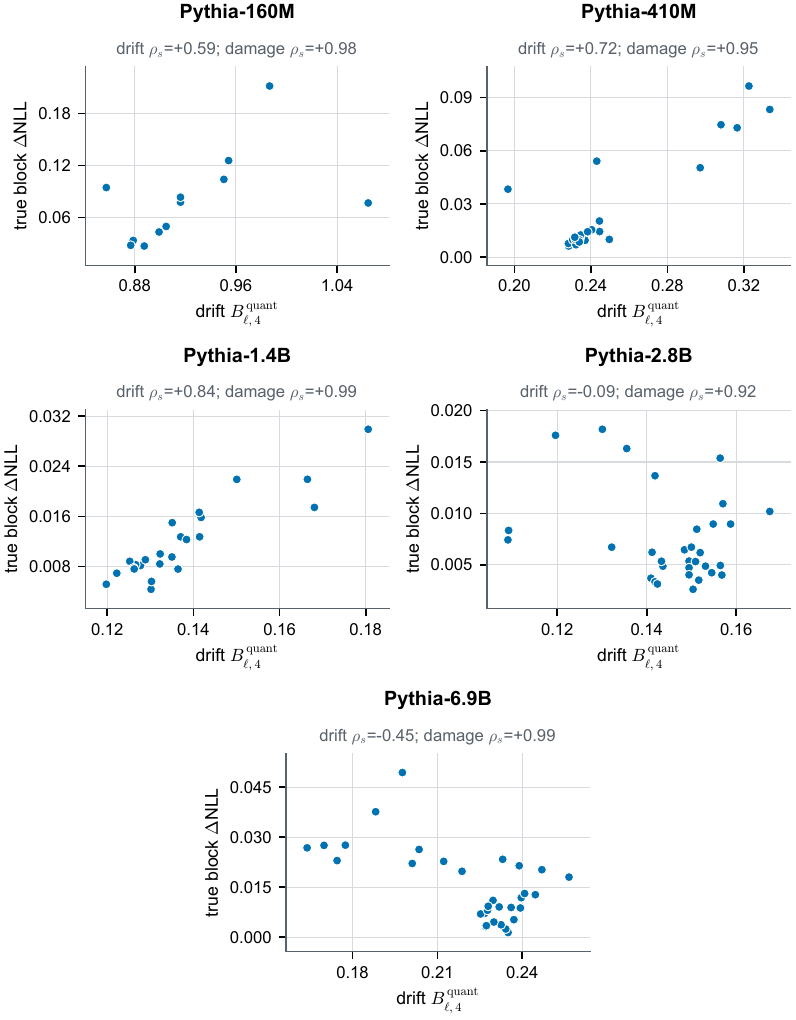} \caption{Pythia block scores. For each model, calibration-only drift is plotted against realized single-block int4 $\Delta$NLL, and the header gives the drift and damage rank correlations. Each point represents one block.} \label{fig:pythia} \end{figure}

\begin{figure}[t!] \centering \includegraphics[width=\linewidth]{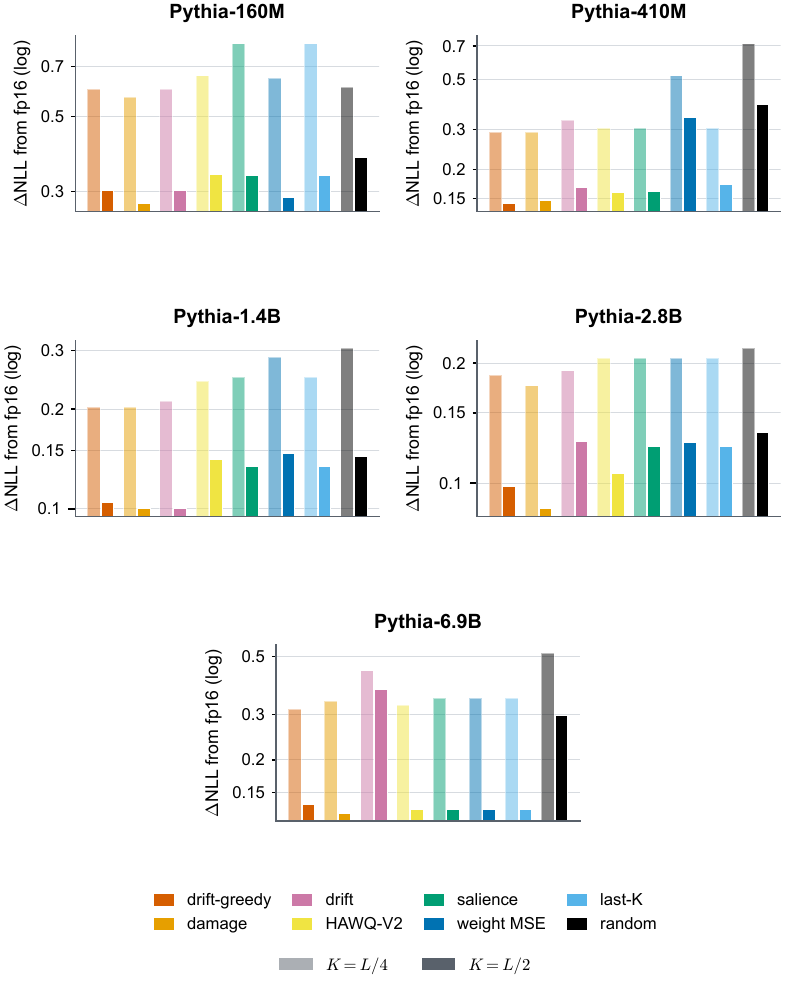} \caption{Pythia allocation quality under per-channel RTN. For each model, bars show configuration $\Delta$NLL relative to fp16 at $K=L/4$ (light) and $K=L/2$ (dark), with a logarithmic vertical axis. The methods and results correspond to Table~\ref{tab:pythia}.} \label{fig:pythia-allocation} \end{figure}

This section provides the detailed allocation analyses supporting Sec.~\ref{sec:exp-quant}, together with the full Pythia results and diagnostic figures. Figures~\ref{fig:pythia} and~\ref{fig:pythia-allocation} visualize the block scores and allocation results, respectively.

\begin{table}[t!] \centering \caption{Spearman correlation between calibration-only block scores and the realized $\Delta$NLL from quantizing one block to int4. HAWQ-V2 combines the positive-part mean Hessian trace with weight perturbation; drift and damage are defined in Eqs.~\ref{eq:bquant} and~\ref{eq:dquant}. The drift column uses the originally reported single 64-sequence calibration draw; damage is pooled over four fresh 64-sequence draws in the post hoc diagnostic pass.} \label{tab:pythia-rc} \small \begin{tabular}{l ccc cc cc} \toprule & \multicolumn{3}{c}{weight and activation} & \multicolumn{2}{c}{Hessian curvature} & \multicolumn{2}{c}{ours} \\ \cmidrule(lr){2-4}\cmidrule(lr){5-6}\cmidrule(lr){7-8} Model & weight MSE & salience & salience$\times$MSE & H trace & HAWQ-V2 & drift & damage \\ \midrule Pythia-160M & 0.71 & 0.68 & 0.68 & 0.43 & 0.41 & 0.59 & 0.98 \\ Pythia-410M & 0.15 & 0.83 & 0.83 & 0.67 & 0.69 & 0.72 & 0.95 \\ Pythia-1.4B & 0.44 & 0.63 & 0.63 & 0.60 & 0.76 & 0.84 & 0.99 \\ Pythia-2.8B & 0.50 & 0.61 & 0.61 & 0.37 & 0.64 & $-0.09$ & 0.92 \\ Pythia-6.9B & 0.96 & 0.96 & 0.96 & 0.46 & 0.83 & $-0.45$ & 0.99 \\ \bottomrule \end{tabular} \end{table}

\begin{table}[t!] \centering \caption{Pythia mixed-precision allocation under per-channel RTN. WikiText-103 perplexity is lower-is-better; $K$ of $L$ blocks use int8 and the rest int4. Baselines are positional, weight MSE, salience, and HAWQ-V2. Our calibration-only rules use additive drift, additive damage, or drift-greedy remeasurement of mean symmetric KL. The greedy column denotes drift-greedy throughout the allocation tables unless damage-greedy is explicitly named. LBD is LAMBADA accuracy for the drift-greedy configuration. Bold marks the lowest unrounded perplexity per model and budget; equal displayed values may differ before rounding. The upper part reports reference values and nonfunctional baselines; the lower part reports HAWQ-V2 and the functional rules for the same model--budget rows.} \label{tab:pythia} \small \begin{tabular}{l c cc ccccc} \toprule & & \multicolumn{2}{c}{reference} & \multicolumn{5}{c}{baselines} \\ \cmidrule(lr){3-4}\cmidrule(lr){5-9} Model & $K/L$ & fp16 & int4 & rand & first & last & MSE & salience \\ \midrule \multirow{2}{*}{Pythia-160M} & 3/12 & \multirow{2}{*}{50.57} & \multirow{2}{*}{156.81} & 93.11 & 138.30 & 114.83 & 96.56 & 114.83 \\ & 6/12 & & & 73.82 & 102.32 & 70.62 & 67.48 & 70.62 \\ \addlinespace \multirow{2}{*}{Pythia-410M} & 6/24 & \multirow{2}{*}{18.43} & \multirow{2}{*}{42.47} & 37.74 & 38.66 & 24.99 & 30.94 & 24.99 \\ & 12/24 & & & 27.15 & 34.16 & 21.92 & 25.93 & 21.63 \\ \addlinespace \multirow{2}{*}{Pythia-1.4B} & 6/24 & \multirow{2}{*}{13.11} & \multirow{2}{*}{19.69} & 17.79 & 17.91 & 16.84 & 17.46 & 16.84 \\ & 12/24 & & & 15.14 & 16.40 & 15.00 & 15.18 & 15.00 \\ \addlinespace \multirow{2}{*}{Pythia-2.8B} & 8/32 & \multirow{2}{*}{11.34} & \multirow{2}{*}{15.45} & 14.10 & 14.22 & 13.92 & 13.92 & 13.92 \\ & 16/32 & & & 12.96 & 13.46 & 12.83 & 12.87 & 12.83 \\ \addlinespace \multirow{2}{*}{Pythia-6.9B} & 8/32 & \multirow{2}{*}{10.56} & \multirow{2}{*}{19.40} & 17.73 & 18.50 & 14.92 & 14.92 & 14.92 \\ & 16/32 & & & 14.24 & 17.13 & 12.01 & 12.01 & 12.01 \\ \bottomrule \end{tabular}

\vspace{0.7em} \begin{tabular}{l c c ccc c} \toprule & & baseline & \multicolumn{3}{c}{functional rules} & \\ \cmidrule(lr){3-3}\cmidrule(lr){4-6} Model & $K/L$ & HAWQ-V2 & drift & damage & greedy & LBD \\ \midrule \multirow{2}{*}{Pythia-160M} & 3/12 & 97.70 & 92.24 & \textbf{89.43} & 92.24 & 5.3 \\ & 6/12 & 70.86 & 68.38 & \textbf{66.63} & 68.38 & 9.1 \\ \addlinespace \multirow{2}{*}{Pythia-410M} & 6/24 & 24.99 & 25.64 & \textbf{24.69} & \textbf{24.69} & 33.5 \\ & 12/24 & 21.59 & 21.78 & 21.33 & \textbf{21.24} & 41.7 \\ \addlinespace \multirow{2}{*}{Pythia-1.4B} & 6/24 & 16.71 & 16.19 & \textbf{16.06} & \textbf{16.06} & 58.5 \\ & 12/24 & 15.09 & \textbf{14.49} & 14.49 & 14.55 & 58.5 \\ \addlinespace \multirow{2}{*}{Pythia-2.8B} & 8/32 & 13.93 & 13.73 & \textbf{13.51} & 13.66 & 52.3 \\ & 16/32 & 12.60 & 12.87 & \textbf{12.36} & 12.51 & 60.1 \\ \addlinespace \multirow{2}{*}{Pythia-6.9B} & 8/32 & 14.62 & 16.43 & 14.80 & \textbf{14.47} & 42.3 \\ & 16/32 & 12.01 & 15.37 & \textbf{11.97} & 12.09 & 53.3 \\ \bottomrule \end{tabular} \end{table}

\paragraph{Functional allocation achieves the lowest perplexity at equal bit budgets across the tested model sizes.} With $K$ of $L$ blocks at int8 and the rest at int4, one of the functional rules, additive top-$K$ by drift or damage or greedy, gives the lowest perplexity in every one of the ten model--budget settings, marked in bold in Table~\ref{tab:pythia}. With each identical-set tie credited once, the damage rule accounts for six lowest-perplexity cells: both 160M budgets, both 2.8B budgets, 6.9B $K=16$, and 410M $K=6$, where the damage rule independently selects exactly greedy's block set; at 160M $K=6$ that rule also overtakes the weight-MSE rule that had edged the drift rule. Greedy accounts for three: 410M $K=12$, 1.4B $K=6$, again the same set as damage, and 6.9B $K=8$; at 1.4B $K=12$ the drift and damage rules are equal to within $0.003\%$, $14.493$ against $14.494$, both ahead of every baseline.

On Pythia-1.4B at $K=L/4$, greedy recovers $55\%$ of the int4-to-fp16 perplexity gap against $45\%$ for HAWQ-V2, $43\%$ for salience, and $34\%$ for weight-MSE; at $K=L/2$ the additive rules recover $79\%$ against $70\%$ for HAWQ-V2 and $71\%$ for salience. On Pythia-2.8B the damage rule is best at both budgets, $13.51$ against $13.92$ to $14.22$ at $K=8$ and $12.36$ against $12.60$ to $13.46$ at $K=16$, recovering $47\%$ and $75\%$ of the gap against $37\%$ and $69\%$ for HAWQ-V2. On Pythia-6.9B greedy is best at $K=8$, $14.47$ against $14.62$ for HAWQ-V2, the closest baseline, and the damage rule at $K=16$, $11.97$ against $12.01$ for weight-MSE, salience, HAWQ-V2, and last-$K$, which all converge to the same set here; the difference is $0.4\%$.

Composing the two functional fixes, so that greedy re-measurement is driven by the calibration damage score rather than by drift, refines the $K=L/2$ settings at the largest model sizes, reaching $12.31$ against $12.36$ for additive damage at 2.8B $K=16$ and selecting the lowest-perplexity sets at both 6.9B budgets, but not uniformly: At 2.8B $K=8$ the composition gives $13.54$ against $13.51$. We report that composition as a consistency probe rather than as a further rule.

\paragraph{Baseline rankings vary across model sizes, and salience-based selections largely follow depth.} The baseline rank correlations vary with model size: Weight-MSE reaches $0.71$ at 160M but falls to $0.15$ at 410M; activation salience climbs to $0.83$ at 410M but drops to $0.63$ at 1.4B; HAWQ-V2 reaches $0.41$ at 160M and $0.69$ and $0.76$ at 410M and 1.4B, respectively (Table~\ref{tab:pythia-rc}). Because pre-norm residual-stream magnitudes grow with depth, the salience$\times$MSE ranking largely follows a positional deepest-layers-first ordering: The salience$\times$MSE top-$K$ set coincides exactly with last-$K$ in nine of the ten model--budget settings. Compare the last and salience columns of Table~\ref{tab:pythia}. The predictive performance of salience$\times$MSE largely reflects depth. At 6.9B, where the true per-block damage is nearly depth-monotone, weight-MSE and salience score high, $0.96$ in Table~\ref{tab:pythia-rc}, while the functional allocations achieve lower perplexity.

\begin{table}[t!] \centering \caption{Paired evaluation-window bootstrap for the committed rule against HAWQ-V2. Each selected block set is re-evaluated on the same 128 WikiText-103 windows; $\Delta=\mathrm{ppl}(\text{HAWQ-V2})-\mathrm{ppl}(\text{ours})$ uses 10,000 paired resamples. Positive $\Delta$ favors our rule. The committed rule is additive damage at $K=L/4$ and $L/2$ and drift-greedy at the smaller budgets $K=\lfloor L/8\rfloor$ and $\lfloor L/16\rfloor$, without per-cell selection.} \label{tab:allocci} \small \begin{tabular}{l c cc c c c} \toprule & & \multicolumn{2}{c}{perplexity} & & & \\ \cmidrule(lr){3-4} Model & $K/L$ & HAWQ-V2 & ours & $\Delta$ & $95\%$ CI ($\Delta$) & result \\ \midrule \multicolumn{7}{l}{$K=L/4$ and $L/2$: committed damage rule, Table~\ref{tab:pythia}} \\ \multirow{2}{*}{160M} & 3/12 & 97.70 & 89.43 & $+8.27$ & $[+6.99, +9.52]$ & ours \\ & 6/12 & 70.86 & 66.63 & $+4.23$ & $[+3.52, +4.94]$ & ours \\ \addlinespace \multirow{2}{*}{410M} & 6/24 & 24.99 & 24.69 & $+0.30$ & $[+0.20, +0.40]$ & ours \\ & 12/24 & 21.59 & 21.33 & $+0.26$ & $[+0.19, +0.33]$ & ours \\ \addlinespace \multirow{2}{*}{1.4B} & 6/24 & 16.71 & 16.06 & $+0.64$ & $[+0.58, +0.72]$ & ours \\ & 12/24 & 15.09 & 14.49 & $+0.60$ & $[+0.54, +0.66]$ & ours \\ \addlinespace \multirow{2}{*}{2.8B} & 8/32 & 13.93 & 13.51 & $+0.41$ & $[+0.37, +0.46]$ & ours \\ & 16/32 & 12.60 & 12.36 & $+0.24$ & $[+0.21, +0.28]$ & ours \\ \addlinespace \multirow{2}{*}{6.9B} & 8/32 & 14.62 & 14.80 & $-0.18$ & $[-0.25, -0.11]$ & HAWQ \\ & 16/32 & 12.01 & 11.97 & $+0.05$ & $[+0.02, +0.08]$ & ours \\ \midrule \multicolumn{7}{l}{Smaller budgets: committed drift-greedy rule, Table~\ref{tab:tight}} \\ \multirow{2}{*}{1.4B} & 1/24 & 18.98 & 18.98 & $+0.00$ & $[+0.00, +0.00]$ & tie \\ & 3/24 & 18.51 & 17.45 & $+1.06$ & $[+0.97, +1.16]$ & ours \\ \addlinespace \multirow{2}{*}{2.8B} & 2/32 & 15.29 & 14.72 & $+0.56$ & $[+0.51, +0.62]$ & ours \\ & 4/32 & 14.97 & 14.28 & $+0.69$ & $[+0.63, +0.75]$ & ours \\ \addlinespace \multirow{2}{*}{6.9B} & 2/32 & 17.95 & 17.46 & $+0.49$ & $[+0.40, +0.58]$ & ours \\ & 4/32 & 16.40 & 16.29 & $+0.11$ & $[+0.04, +0.19]$ & ours \\ \bottomrule \end{tabular} \end{table}

\paragraph{Paired intervals resolve the equal-budget differences.} The positive gaps over HAWQ-V2 are $0.24$ to $0.65$ perplexity across $410$M to $2.8$B. Table~\ref{tab:allocci} supplements the point estimates in Table~\ref{tab:pythia} with paired bootstrap intervals. Each allocation's marginal $95\%$ perplexity interval is $\approx2$ perplexity wide and the two overlap almost completely: At $410$M and $K=L/4$, the damage rule gives $[23.65, 25.72]$ against $[23.93, 26.04]$ for HAWQ-V2. The two allocations are evaluated on the same $128$ windows, so the marginal intervals for both allocations share evaluation-window variation. Paired resampling accounts for that shared variation when estimating the difference.

The paired interval on the gap is an order of magnitude tighter and separates the rules in all ten cells: Damage is ahead in $9$ of the $10$ model--budget cells and behind in the remaining one. Even the sub-half-perplexity positive gaps are resolved: $410$M gives $+0.30$ with interval $[+0.20, +0.40]$ at $p<10^{-3}$, and $2.8$B at $K=L/2$ gives $+0.24$ with interval $[+0.21, +0.28]$, also at $p<10^{-3}$. Committing to one damage rule here, with no per-cell rule choice, makes the evaluation a comparison of one rule against one baseline rather than a best-of-rules comparison.

The two exceptions are both on Pythia-6.9B, where several allocation methods achieve similar perplexity: The committed damage rule falls significantly behind HAWQ-V2 at $K=L/4$, at $-0.18$ with interval $[-0.25, -0.11]$, though the interaction-aware greedy variant recovers that cell at $+0.14$ with interval $[+0.08, +0.21]$, and the committed damage rule's $K=L/2$ edge, while clearing zero, is practically negligible at $+0.05$ with interval $[+0.02, +0.08]$.

The smaller-budget part of Table~\ref{tab:allocci} makes the complementary point motivated by Theorem~\ref{thm:alloc-regret}: Once the int8 budget is small enough that a single mis-ranked block cannot be hidden, the committed greedy rule's margin over HAWQ-V2 is interval-separated in five of the six smaller-budget settings, from $+0.11$ to $+1.06$ perplexity; the sixth is the trivial $K=1$ tie. The allocation advantage is modest on the largest model at $K=L/2$, and becomes clearly separated by the interval when tight budgets make individual block mis-rankings consequential.

\paragraph{Greedy remeasurement accounts for block interactions.} Every top-$K$ rule above scores each block in isolation from the others and assumes joint damage is roughly the sum of per-block damages. We audit a close empirical proxy for this assumption: For each evaluated allocation with high-precision set $H$, so that blocks $\ell\notin H$ remain int4, compare the additive prediction $\sum_{\ell\notin H} \Delta\mathrm{NLL}_\ell$ with the measured joint $\Delta$NLL. Because the reported complement $H$ is int8 rather than fp16, this discrepancy also contains the int8 complement's damage and interactions with the int4 blocks; the int8-only damage is at most $0.003$ nats here, but the comparison is not a direct estimate of the mixed-curvature envelope; see the caveat at the end of App.~\ref{app:theory-alloc}.

The Spearman correlation between predicted and actual joint damage is $0.98$ to $1.00$ at 160M, $0.88$ to $1.00$ at 410M, and $0.88$ to $0.97$ at 1.4B, and drops to $0.47$ at 2.8B for $K=L/4$, against $0.85$ at $K=L/2$; at 6.9B the correlation is $0.74$ at $K=L/4$ and $1.00$ at $K=L/2$. Two regimes follow. Where additivity holds, an accurate per-block score is all one needs. Empirically, the additive damage rule has the lowest perplexity in all four settings at 160M and 2.8B and at 6.9B $K=16$; the additive damage rule's 2.8B $K=L/4$ result comes despite the weak $0.47$ additivity audit in that cell. This near-additive case is the regime quantified by Theorem~\ref{thm:alloc-stability}: If the audited near-additivity extends uniformly over candidate sets, the uniform mixed-curvature envelope and calibration error give an explicit near-oracle regret bound.

Where additivity degrades, at $K=L/4$ on 2.8B and 6.9B, which particular combination of blocks interacts begins to matter: The greedy variant re-measures the mean symmetric KL $\mathcal C_{\mathrm{SKL}}$ of whichever mixed configuration is current at each step, from calibration data only, never relies on additivity, and delivers the best allocation at 6.9B $K=8$ while staying within $1.1\%$ of the damage rule at 2.8B $K=8$. The lesson mirrors the paper's theme at the parameter-noise scale: Re-measuring a joint finite-scale functional, output drift here or task damage in the damage-greedy variant, removes the additive modeling step precisely where that step becomes unreliable.

\paragraph{Drift-greedy also improves allocation at smaller model sizes.} The drift-greedy rule improves over the additive drift rule in seven of the ten settings, reaching $24.69$ against $25.64$ perplexity at 410M and $K=L/4$ for instance, selects exactly the additive rule's sets in the two 160M settings, and loses only once, by $0.4\%$, at 1.4B, $K=L/2$, where the additive rules perform better. All selection uses training-domain calibration sequences, and all rules are evaluated on the same held-out data.

\paragraph{Allocation gains complement quantizer error compensation.} We use plain RTN so that the sensitivity signal is not confounded with reconstruction machinery, which leaves open whether a better quantizer would absorb the allocation's advantage. On Pythia-410M we rerun the pipeline under a one-shot GPTQ variant \citep{frantar2022gptq} with Cholesky-based error compensation on the same per-channel grid and calibration Hessians from the same 64 sequences, as specified in App.~\ref{app:measurement}: GPTQ improves uniform int4 from $42.5$ to $30.6$ perplexity, and the greedy int8 sets chosen under RTN roughness still recover $80\%$ and $90\%$ of the remaining int4-to-fp16 gap at $K=L/4$ and $K=L/2$, lowering perplexity from $30.6$ to $20.9$ and $19.7$ against fp16 $18.4$, compared with $74\%$ and $88\%$ recovery under RTN. The two mechanisms compose: Error compensation shrinks the noise per block, roughness-guided allocation decides where the remaining noise does the least functional damage, and the allocation transfers across quantizers without re-measurement.

The same evaluation on Pythia-1.4B and 2.8B shows that allocation gains persist at larger model sizes: GPTQ's error compensation alone already brings uniform int4 to within $13\%$ and $14\%$ of fp16, at $14.8$ and $12.9$ against $13.1$ and $11.3$ perplexity, and the RTN-chosen greedy sets, transferred with no re-measurement, still recover a further $34\%$ and $57\%$ on 1.4B and $31\%$ and $60\%$ on 2.8B of the residual int4-to-fp16 gap at $K=L/4$ and $K=L/2$. The allocation headroom at $K=L/4$ shrinks as the base quantizer approaches fp16, but the $K=L/2$ allocation recovers at least $57\%$ of the residual gap at every scale.

The composition extends to the outlier-aware AWQ format (App.~\ref{sec:exp-awq}): With the entire comparison rebuilt on AWQ's activation-aware scaling, the damage rule yields lower perplexity than HAWQ-V2 at $K=L/4$ on every model and is at parity at $K=L/2$. The same matched-budget comparison on a controlled SpQR-style dense-and-sparse variant (App.~\ref{sec:exp-spqr}) matches or improves over the HAWQ-V2 allocation in all twelve model--budget cells. Each of these comparisons isolates the allocation rule within one quantization format.

\begin{table}[t!] \centering \caption{RTN perplexity across bit budgets. $K$ of $L$ blocks use int8 and the rest int4; smaller $K$ is a tighter budget. The same drift-greedy rule is used in every cell. Bold marks improvement over HAWQ-V2. Each part includes seven held-out LLMs not used to design the rule, and repeats the fp16 and uniform-int4 references for readability.} \label{tab:tight} \small \begin{tabular}{l cc cc cc} \toprule & \multicolumn{2}{c}{reference} & \multicolumn{2}{c}{$K=L/2$} & \multicolumn{2}{c}{$K=L/4$} \\ \cmidrule(lr){2-3}\cmidrule(lr){4-5}\cmidrule(lr){6-7} Model & fp16 & int4 & HAWQ-V2 & \shortstack{drift-\\greedy} & HAWQ-V2 & \shortstack{drift-\\greedy} \\ \midrule Pythia-1.4B & 13.11 & 19.69 & 15.09 & \textbf{14.55} & 16.71 & \textbf{16.06} \\ Pythia-2.8B & 11.34 & 15.45 & 12.60 & \textbf{12.51} & 13.93 & \textbf{13.66} \\ Pythia-6.9B & 10.56 & 19.40 & 12.01 & 12.09 & 14.62 & \textbf{14.47} \\ \midrule \multicolumn{7}{l}{Held-out LLMs}\\ Phi-3.5-mini (3.8B) & 6.79 & 11.20 & 8.78 & \textbf{7.82} & 10.06 & \textbf{8.75} \\ OLMo-2-7B & 10.14 & 15.28 & 11.91 & \textbf{10.68} & 12.57 & \textbf{11.39} \\ Granite-3.1-8B & 5.74 & 8.03 & 6.45 & \textbf{6.10} & 6.96 & \textbf{6.49} \\ Hermes-3 & 8.50 & 13.33 & 10.07 & \textbf{9.53} & 10.82 & \textbf{10.00} \\ Mistral-NeMo (12B) & 6.48 & 17.00 & 7.38 & \textbf{7.07} & 7.81 & \textbf{7.53} \\ Qwen2.5-14B & 6.55 & 9.16 & 7.15 & \textbf{7.01} & 7.55 & \textbf{7.36} \\ Qwen2.5-32B & 6.16 & 8.48 & 6.80 & \textbf{6.39} & 7.25 & \textbf{6.73} \\ \bottomrule \end{tabular}

\vspace{0.7em} \begin{tabular}{l cc cc cc} \toprule & \multicolumn{2}{c}{reference} & \multicolumn{2}{c}{$K=L/8$} & \multicolumn{2}{c}{$K=L/16$} \\ \cmidrule(lr){2-3}\cmidrule(lr){4-5}\cmidrule(lr){6-7} Model & fp16 & int4 & HAWQ-V2 & \shortstack{drift-\\greedy} & HAWQ-V2 & \shortstack{drift-\\greedy} \\ \midrule Pythia-1.4B & 13.11 & 19.69 & 18.51 & \textbf{17.45} & 18.98 & 18.98 \\ Pythia-2.8B & 11.34 & 15.45 & 14.97 & \textbf{14.28} & 15.29 & \textbf{14.72} \\ Pythia-6.9B & 10.56 & 19.40 & 16.40 & \textbf{16.29} & 17.95 & \textbf{17.46} \\ \midrule \multicolumn{7}{l}{Held-out LLMs}\\ Phi-3.5-mini (3.8B) & 6.79 & 11.20 & 10.71 & \textbf{9.58} & 11.19 & \textbf{10.42} \\ OLMo-2-7B & 10.14 & 15.28 & 14.43 & \textbf{12.55} & 15.17 & \textbf{14.18} \\ Granite-3.1-8B & 5.74 & 8.03 & 7.77 & \textbf{6.82} & 8.33 & \textbf{7.18} \\ Hermes-3 & 8.50 & 13.33 & 11.20 & \textbf{10.36} & 11.46 & \textbf{10.68} \\ Mistral-NeMo (12B) & 6.48 & 17.00 & 8.07 & \textbf{7.88} & 16.64 & \textbf{8.28} \\ Qwen2.5-14B & 6.55 & 9.16 & 7.88 & \textbf{7.79} & 8.12 & 8.23 \\ Qwen2.5-32B & 6.16 & 8.48 & 7.40 & \textbf{7.04} & 7.48 & \textbf{7.27} \\ \bottomrule \end{tabular} \end{table}

\paragraph{The advantage concentrates where the budget is tight.} The allocations achieve similar perplexity at $K=L/4$ or $L/2$ in several settings, including Pythia-6.9B under GPTQ (App.~\ref{sec:exp-gptq}). Table~\ref{tab:tight} examines smaller budgets using the same drift-greedy rule in every cell, providing a comparison of one fixed rule with HAWQ-V2.

At $K=L/2$, each rule protects many high-sensitivity blocks, so HAWQ-V2 and the greedy rule nearly coincide: within $0.10$ perplexity on Pythia-2.8B, at $+0.09$, with HAWQ-V2 marginally ahead on 6.9B by $0.08$. As the int8 budget tightens to $L/8$ and $L/16$, where only two to four blocks can be kept at int8 and a single mis-ranked block can no longer be masked by a generous budget, the greedy rule's edge widens, to $+0.69$ perplexity for 2.8B at $L/8$ and $+0.49$ for 6.9B at $L/16$, and to $+1.06$ on the smaller Pythia-1.4B.

The same pattern extends beyond Pythia. Table~\ref{tab:tight} repeats the evaluation with the fixed rule on the seven held-out LLMs introduced in Sec.~\ref{sec:exp-modllm}, where the greedy rule yields lower perplexity than HAWQ-V2 in $27$ of the $28$ model--budget cells, and the edge again grows toward the tight end, most sharply where the Hessian proxy carries no rank information at all. The following cases describe the ranking differences and settings where drift-greedy ties with or falls behind HAWQ-V2.

\paragraph{Block-ranking errors at tight budgets.} Table~\ref{tab:tight} reflects concrete block-ranking mismatches. At $K=2$ on Pythia-6.9B, HAWQ-V2 spends the entire protection budget on blocks $\{18,20\}$ whose realized int4 damage is minor and misses the two most-damaging blocks $\{30,31\}$, while the task-aligned damage rule protects exactly $\{30,31\}$. On Pythia the trend has one exception at $K=1$ on Pythia-1.4B, where protecting a single block of $24$ produces a tie among the evaluated rules.

The same widening appears outside the Pythia family. The edge grows toward the tight end just as on Pythia, with Granite running from $+0.35$ at $L/2$ to $+1.15$ perplexity at $L/16$, and the blind spot is starkest on Mistral-NeMo-12B, whose HAWQ-V2 sensitivity score bears essentially no correlation with realized int4 damage, at rank correlation $0.04$: At $K=2$ HAWQ-V2 collapses to $16.6$ perplexity, barely improving on the uniform-int4 $17.0$, the damage that allocation is meant to relieve, while the greedy rule holds at $8.3$.

On Phi-3.5, drift-greedy yields $0.96$ lower perplexity at the $K=L/2$ RTN budget in Table~\ref{tab:tight}. The GPTQ diagnostic also shows a ranking mismatch: The HAWQ-V2 score has correlation $-0.20$ with true int4 damage (Table~\ref{tab:modllm-rc}). Qwen2.5-32B shows a similar pattern at the largest model size: With the HAWQ-V2 score on this model weakly correlated with realized damage at rank correlation $0.12$, the greedy rule has lower perplexity at all four budgets, by $0.41$ at $L/2$, peaking at $0.52$ at $L/4$, and $0.21$ at $L/16$; the edge is already present at $K=L/2$ rather than emerging only under tightening.

The complementary case occurs at $K=3$ of $48$ on Qwen2.5-14B, where HAWQ-V2's three-block set coincides with the per-block oracle exactly, at $\{4,46,47\}$ and $8.12$ perplexity, and the label-free drift rule also finds the oracle set, while the greedy rule misses oracle block~$4$ and lands at $8.23$, the one tight cell in the study where the infinitesimal proxy points at exactly the right blocks.

\FloatBarrier \subsection{HAWQ-V2 calibration-budget sweep} \label{app:hawqbudget}

This subsection tests a calibration-budget explanation for HAWQ-V2's deficit against the interaction-aware greedy; Table~\ref{tab:hawqbudget} reports the full sweep.

A calibration-budget sweep over an eight-fold range, from 4 to 32 sequences, shows that the HAWQ-V2 comparison is not explained by fewer Hessian-trace samples (Table~\ref{tab:hawqbudget}).

\begin{table}[t!] \centering \caption{HAWQ-V2 calibration-budget sweep on Pythia-2.8B under RTN. Across an $8\times$ sequence range, correlation with block damage plateaus near $0.66$; by 16 sequences, the ranking and top-8 set match the 32-sequence result. Perplexity changes by less than $0.02$. Reference rows do not use a HAWQ-V2 calibration budget; NA marks entries outside that sweep.} \label{tab:hawqbudget} \small \begin{tabular}{l ccc cc} \toprule & \multicolumn{3}{c}{allocation stability} & \multicolumn{2}{c}{WikiText-103 perplexity} \\ \cmidrule(lr){2-4}\cmidrule(lr){5-6} \shortstack{Rule and HAWQ\\calibration budget} & $\rho$(truth) & $\rho(n=32)$ & top-8 & $K=L/4$ & $K=L/2$ \\ \midrule HAWQ-V2, $n_{\mathrm{seq}}=4$ & 0.61 & 0.93 & 0.88 & 13.93 & 12.62 \\ HAWQ-V2, $n_{\mathrm{seq}}=16$ & 0.66 & 1.00 & 1.00 & 13.92 & 12.60 \\ HAWQ-V2, $n_{\mathrm{seq}}=32$ & 0.66 & 1.00 & 1.00 & 13.92 & 12.60 \\ \midrule damage-greedy, 96 seq. & \multicolumn{3}{c}{NA} & \textbf{13.54} & \textbf{12.31} \\ uniform int4 & \multicolumn{3}{c}{NA} & \multicolumn{2}{c}{15.45} \\ \bottomrule \end{tabular} \end{table}

\paragraph{Sensitivity to the HAWQ-V2 calibration budget.} The default Pythia-2.8B HAWQ-V2 trace uses two calibration sequences, while the damage-greedy reference in Table~\ref{tab:hawqbudget} uses 96. To assess the effect of the HAWQ-V2 calibration budget, we re-estimate the HAWQ-V2 trace on Pythia-2.8B while sweeping the number of calibration sequences over an $8\times$ range, $n_{\mathrm{seq}}\in\{4,16,32\}$, and re-solve the allocation at each (Table~\ref{tab:hawqbudget}, App.~\ref{app:hawqbudget}). The resulting allocations change little over this range: The rank correlation with the true per-block int4 damage is $0.61$ at the smallest budget and plateaus at $\approx 0.66$; the HAWQ-V2 block selection matches the most-calibrated HAWQ-V2 answer with top-8 overlap $1.00$ and rank correlation $1.00$ against the most-calibrated allocation at $n_{\mathrm{seq}}=32$ by $n_{\mathrm{seq}}=16$; and the resulting perplexity moves by less than $0.02$. For context, the reported 96-sequence damage-greedy result is $0.38$ and $0.29$ perplexity below HAWQ-V2's 32-sequence allocation at $L/4$ and $L/2$. Although the sweep does not equalize HAWQ-V2 to 96 sequences, exact allocation stability from 16 to 32 sequences and the sub-$0.02$ perplexity movement strongly disfavor under-resolved trace estimation over the tested range as the explanation for this gap.

\FloatBarrier \subsection{Depth propagation of block-quantization noise} \label{app:depth}

\begin{figure}[t!] \centering \includegraphics[width=\linewidth]{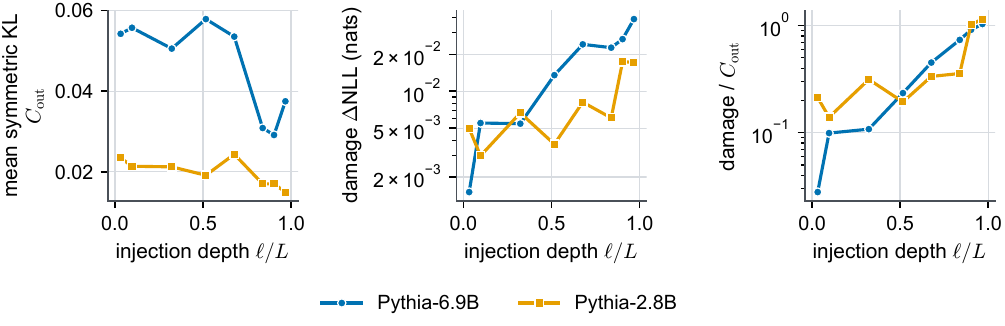} \caption{Depth propagation of block-int4 perturbations at Pythia-6.9B and 2.8B, by injection depth. Left: mean symmetric KL at the output $C_{\mathrm{out}}$. Middle: task damage, log scale. Right: damage divided by $C_{\mathrm{out}}$, log scale. Shallow-block noise produces more output drift and less damage; the ratio increases with depth, at Spearman $1.00$ on 6.9B and $0.88$ on 2.8B.} \label{fig:depth} \end{figure}

\paragraph{Drift and damage separate with depth.} At 6.9B, damage per unit of output movement increases with block depth, with $\rho_s=0.96$. Shallow-block noise reaches the output with more distributional movement than deep-block noise but causes $7\times$ less task damage (Table~\ref{tab:depth} and Figure~\ref{fig:depth}). The relevant distinction is the direction of movement relative to the next-token target, not movement magnitude alone. Damage restores the block ranking to $0.92$ at 2.8B and $0.99$ at 6.9B without additional forward passes. Figure~\ref{fig:qual-token-dissociation} in App.~\ref{sec:exp-spectrum} shows the same dissociation at token level.

The dissociation diagnosis of Sec.~\ref{sec:exp-quant} leaves two candidate mechanisms for large but benign shallow-block drift: Either later blocks and normalization layers progressively re-absorb the perturbation, a magnitude account, or the perturbation survives but points in task-irrelevant output directions, a direction account. A forward-only depth-propagation probe separates the two accounts. For each probe block $\ell$, we quantize that block alone to int4 and track the induced perturbation on 64 calibration sequences, a fresh draw disjoint from the scoring draw: the relative residual-stream drift $\E\norm{h'_d-h_d}/\E\norm{h_d}$ at every subsequent depth $d$, and the output-relevant divergence obtained by applying the model's own readout, consisting of the final LayerNorm and unembedding, to the perturbed stream. At the last layer that divergence is mean symmetric KL, $C_{\mathrm{out}}\coloneqq\E_x\dskl(P_x,Q_x)^2$, the square of the drift-amplitude functional in \eqref{eq:bquant}. We report $C_{\mathrm{out}}$ in nats.

Table~\ref{tab:depth} reports the endpoints at 6.9B. The magnitude data do not support the magnitude account: Shallow perturbations are not absorbed; relative drift instead grows roughly threefold on the way to the readout, and shallow injections arrive with more final output movement than deep ones, $0.050$ to $0.056$ against $0.029$ to $0.038$ nats of mean symmetric KL. Calibration damage per unit of arriving movement rises strictly monotonically with injection depth, at Spearman $1.00$ over the eight probe blocks and $0.88$ at 2.8B, where the drift--damage dissociation is weaker, consistent with the milder pooled four-draw diagnostic-pass inversion there, at $-0.06$ against the original single-draw $-0.09$ of Table~\ref{tab:pythia-rc}. The network does not shrink shallow noise but routes that noise into directions the task readout is insensitive to, which is why a divergence that weights all output directions equally, the drift, mis-ranks blocks precisely at the scale where this routing becomes strong, while the calibration damage score, which projects the same movement onto the task, does not.

Two further details characterize the dissociation in Sec.~\ref{sec:exp-quant}. First, the drift--damage association differs across depth: At 6.9B the drift score's rank correlation against the truth is $-0.78$ on the deep half of the network, blocks 16 to 31, but $+0.22$ on the shallow half, blocks 0 to 15. The negative association is already present within the deep half. Second, the reversal also appears in another drift-family functional: A top-1 flip-rate probe, computed from the same forward passes, also has a negative correlation, at $-0.42$ on 6.9B. Quantitatively, at 6.9B the shallow probe blocks $\{1,3,10\}$ arrive at the readout with mean symmetric KL $0.054$ against $0.033$ for the deep blocks $\{26,28,30\}$, so the shallow blocks move the output more, yet cause $7\times$ less damage, $0.004$ against $0.029$ nats.

\begin{table}[t!] \centering \caption{Depth propagation at Pythia-6.9B: relative residual drift at the injection block and at the last layer, mean symmetric KL at the output $C_{\mathrm{out}}$ (nats), per-block calibration damage $\Delta$NLL (nats), and the damage-to-KL ratio. Damage per unit of mean symmetric KL rises strictly with depth, at $\rho_s=1.00$ here and $0.88$ at 2.8B.} \label{tab:depth} \small \begin{tabular}{c cc c c c} \toprule Block & drift@inj & drift@final & $C_{\mathrm{out}}$ & damage & damage/$C_{\mathrm{out}}$ \\ \midrule 1 & 0.084 & 0.252 & 0.054 & 0.0015 & 0.03 \\ 3 & 0.087 & 0.255 & 0.056 & 0.0055 & 0.10 \\ 10 & 0.077 & 0.244 & 0.051 & 0.0055 & 0.11 \\ 16 & 0.077 & 0.257 & 0.058 & 0.0135 & 0.23 \\ 21 & 0.081 & 0.248 & 0.054 & 0.0241 & 0.45 \\ 26 & 0.084 & 0.192 & 0.031 & 0.0227 & 0.73 \\ 28 & 0.099 & 0.188 & 0.029 & 0.0266 & 0.91 \\ 30 & 0.123 & 0.192 & 0.037 & 0.0388 & 1.03 \\ \bottomrule \end{tabular} \end{table}

 \FloatBarrier \subsection{Preregistered held-out test on Pythia-12B} \label{app:prereg}

This subsection details the preregistered evaluation summarized in Sec.~\ref{sec:exp-quant}, where the 160M to 6.9B drift reversal that motivated the calibration damage score is reported. We fixed five falsifiable predictions in a written preregistration before producing the Pythia-12B quantization and diagnostic outputs. That checkpoint has 36 layers and hidden size 5,120, and is the largest Pythia checkpoint, absent from the 160M to 6.9B pool that motivated the calibration damage score. We then ran that same forward-only pipeline with no code change.

All five criteria were met (Table~\ref{tab:pythia-held}): The drift score does not recover, at pooled Spearman $-0.221$ with bootstrap 95\% CI $[-0.228,-0.211]$, continuing the reversal seen at 2.8B and 6.9B and staying below the pre-set $+0.10$ ceiling, while the damage score stays robust across model sizes at $+0.973$, with CI $[+0.958,+0.977]$ and above the $+0.85$ floor, a dissociation gap of $+1.194$ against a pre-set $\ge +0.75$; the flip-rate probe co-moves with drift at $-0.216$, not damage. The damage-score allocation is the best of all rules at both audited budgets: perplexity $14.47$ at $K=9$ and $11.55$ at $K=18$, against $15.45$ and $11.67$ for the best rival.

The same depth structure recurs: Drift against the truth stays positive within each depth half, $0.69$ deep and $0.53$ shallow, yet reverses once pooled across depth, a depth-confounded direction error, not a loss of signal. Because the thresholds preceded the measurements, this evaluation provides an a priori test of the drift--damage account at an unseen Pythia model size.

\begin{table}[t!] \centering \caption{Preregistered Pythia-12B evaluation. Predictions and thresholds were fixed before the quantization and diagnostic outputs existed. Spearman correlations pool four fresh 64-sequence calibration draws against 131,072-token per-block int4 $\Delta$NLL; brackets are percentile-bootstrap 95\% CIs from 2,000 resamples. The primary endpoint is P2 $\wedge$ P3.} \label{tab:pythia-held} \small \begin{tabular}{lll} \toprule Preregistered prediction & Observed & Threshold \\ \midrule P1 low drift--damage correlation & $-0.221$ $[-0.228,-0.211]$ & $\le +0.10$ \\ P2 high damage--damage correlation & $+0.973$ $[+0.958,+0.977]$ & $\ge +0.85$ \\ P3 dissociation persists & $+1.194$ & gap $\ge +0.75$ \\ P4 flip tracks drift & $-0.216$ & $\le +0.30$ \\ P5 damage allocation best & $14.47$, $11.55$ perplexity & $\le +1.0$ of the best rival \\ \bottomrule \end{tabular} \end{table}

\paragraph{Post hoc comparison with HAWQ-V2.} The preregistered rival set did not include a Hessian baseline at 12B. A subsequent seeded 8-probe HAWQ-V2 trace \citep{dong2020hawqv2} on Pythia-12B, label- and Hessian-based and the most competitive of our allocation baselines, reaches $14.76$ and $12.25$ perplexity at $K=9$ and $K=18$, and the a priori damage rule yields lower perplexity than that baseline at both budgets, at $14.47$ and $11.55$. Interaction-aware allocation, also analyzed post hoc, widens the margin further: The greedy rules, which re-measure the current mixed configuration's calibration criterion at each upgrade step and never rely on additivity, reach $14.16$ and $11.51$ perplexity for damage-greedy and $14.19$ and $11.57$ for drift-greedy at $K=9$ and $K=18$. Both greedy variants improve on HAWQ-V2 at both budgets. Damage-greedy also improves on additive damage at both budgets, whereas drift-greedy improves on additive damage only at $K=9$.

\section{Allocation comparisons: architectures, formats, and targets} \label{app:contests}

Each subsection below rebuilds the mixed-precision comparison of Sec.~\ref{sec:exp-quant} end to end, with the same score definitions, the same baselines, and the same evaluation, on a different architecture line or weight format, as summarized in Sec.~\ref{sec:exp-generality}. The last two subsections leave the language-model perplexity comparison behind: One re-scores the evaluated allocations on downstream capability benchmarks, the other rebuilds the comparison on an ImageNet-scale vision model.

\subsection{Held-out modern deployment LLMs: per-block rankings and pooled oracle-set agreement} \label{app:modllm-extra}

This subsection reports the per-block rankings and pooled evidence behind Sec.~\ref{sec:exp-modllm}. One memory constraint applies to the comparisons reported there: On Mistral-NeMo-12B, Qwen2.5-14B, and Qwen2.5-32B, whose 131,072- and $151$k-token vocabularies make the GPTQ comparison's cached base-model logits exceed host memory at the default $64$ calibration sequences, the GPTQ comparison uses $32$ sequences instead. Every rule and baseline in those comparisons consumes the identical slice, so the head-to-head comparison is unaffected; the RTN comparisons and every other model use the default.

A direct test of external validity is to freeze the method, every score definition, threshold, and hyperparameter alike, and run the identical pipeline on instruction-tuned decoder LLMs used in deployment, none of which was ever used to design the roughness or damage scores; the only per-model input is a WikiText-103 calibration slice tokenized with the model's own tokenizer. The allocation methods were developed on Pythia and evaluated on two further families of our selection, Qwen2.5 and Mamba (Apps.~\ref{sec:exp-qwen} and~\ref{sec:exp-mamba}). The sweep reported here is a held-out evaluation across the 3.8B to 32B size range where allocation differences are expected to narrow, as for Pythia-6.9B in App.~\ref{sec:exp-gptq}: The base-family sweep of App.~\ref{sec:exp-qwen} never touched the instruction-tuned 14B scale, and the checkpoints are off-the-shelf releases spanning six vendors and pretraining pipelines. The checkpoints' exact Hugging Face identifiers, followed in parentheses by the short labels used in tables, are \path{microsoft/Phi-3.5-mini-instruct} (Phi-3.5) \citep{abdin2024phi3}, \path{allenai/OLMo-2-1124-7B-Instruct} (OLMo-2) \citep{olmo2024olmo2}, \path{ibm-granite/granite-3.1-8b-instruct} (Granite-3.1) \citep{ibm2024granite31}, \path{NousResearch/Hermes-3-Llama-3.1-8B} (Hermes-3), a Llama 3.1 8B fine-tune \citep{teknium2024hermes3,grattafiori2024llama3}, \path{mistralai/Mistral-Nemo-Instruct-2407} (Mistral-NeMo) \citep{mistral2024nemo}, and \path{Qwen/Qwen2.5-14B-Instruct} (Qwen2.5-14B) and \path{Qwen/Qwen2.5-32B-Instruct} (Qwen2.5-32B) \citep{yang2024qwen25}. Every comparison in this paper, here, in Sec.~\ref{sec:exp-quant}, and in the remaining subsections of App.~\ref{app:contests}, fixes the quantizer, RTN or GPTQ, and varies only which blocks receive the int8 budget, so each comparison isolates the allocation rule against Hessian-proxy and parametric-proxy rivals: The AWQ-style salience baseline transplants AWQ's saliency heuristic into an allocation rule, while the AWQ scaling format is a separate axis. Apps.~\ref{sec:exp-awq} and~\ref{sec:exp-spqr} run the same comparison on that axis too, rebuilding the comparison on the AWQ scaling format and on a controlled SpQR-style dense-and-sparse variant.

\begin{table}[t!] \centering \caption{Held-out GPTQ mixed-precision allocation. At each budget, $K$ of $L$ blocks use GPTQ-int8 and the rest use GPTQ-int4. Lower perplexity is better. Greedy denotes drift-greedy. Bold marks the lowest unrounded perplexity per model and budget; equal displayed values may differ before rounding. Other conventions follow Table~\ref{tab:pythia}.} \label{tab:modllm-gptq} \small \fitwidth{%
\begin{tabular}{l c cc ccc ccc c} \toprule & & \multicolumn{2}{c}{reference} & \multicolumn{3}{c}{baselines} & \multicolumn{3}{c}{ours} & \\ \cmidrule(lr){3-4}\cmidrule(lr){5-7}\cmidrule(lr){8-10} Model & $K/L$ & fp16 & gptq-int4 & last & salience & HAWQ-V2 & drift & damage & greedy & LBD \\ \midrule \multirow{2}{*}{Phi-3.5-mini (3.8B)} & 8/32 & \multirow{2}{*}{6.79} & \multirow{2}{*}{8.05} & 7.53 & 7.53 & 7.70 & 7.56 & 7.53 & \textbf{7.50} & 60.0 \\ & 16/32 & & & 7.28 & 7.28 & 7.45 & 7.18 & \textbf{7.15} & 7.15 & 63.2 \\ \addlinespace \multirow{2}{*}{OLMo-2-7B} & 8/32 & \multirow{2}{*}{10.14} & \multirow{2}{*}{13.44} & 11.71 & 11.71 & 12.01 & 11.51 & 11.40 & \textbf{11.33} & 60.4 \\ & 16/32 & & & 10.63 & 10.63 & 11.02 & 10.68 & \textbf{10.49} & 10.58 & 66.4 \\ \addlinespace \multirow{2}{*}{Hermes-3} & 8/32 & \multirow{2}{*}{8.50} & \multirow{2}{*}{182.38} & 165.39 & 165.39 & 9.78 & 9.66 & \textbf{9.50} & 9.57 & 66.8 \\ & 16/32 & & & 156.20 & 156.20 & 9.26 & 9.28 & \textbf{8.99} & 9.18 & 70.0 \\ \addlinespace \multirow{2}{*}{Granite-3.1-8B} & 10/40 & \multirow{2}{*}{5.74} & \multirow{2}{*}{6.61} & 6.23 & 6.23 & 6.22 & 6.21 & \textbf{6.11} & 6.16 & 73.2 \\ & 20/40 & & & 6.10 & 6.10 & 6.04 & 6.04 & \textbf{5.95} & 5.99 & 75.6 \\ \addlinespace \multirow{2}{*}{Mistral-NeMo (12B)} & 10/40 & \multirow{2}{*}{6.48} & \multirow{2}{*}{7.81} & 7.20 & 7.20 & 7.31 & 7.16 & \textbf{7.14} & 7.15 & 70.8 \\ & 20/40 & & & 6.92 & 6.92 & 7.08 & 6.91 & \textbf{6.86} & 6.90 & 73.2 \\ \addlinespace \multirow{2}{*}{Qwen2.5-14B} & 12/48 & \multirow{2}{*}{6.55} & \multirow{2}{*}{7.64} & \textbf{7.10} & \textbf{7.10} & 7.13 & 7.12 & 7.10 & 7.14 & 70.4 \\ & 24/48 & & & 6.95 & 6.95 & 6.93 & 6.90 & \textbf{6.87} & 6.90 & 71.2 \\ \addlinespace \multirow{2}{*}{Qwen2.5-32B} & 16/64 & \multirow{2}{*}{6.16} & \multirow{2}{*}{7.12} & \textbf{6.56} & \textbf{6.56} & 6.78 & 6.66 & 6.59 & 6.59 & 73.2 \\ & 32/64 & & & \textbf{6.36} & 6.37 & 6.56 & \textbf{6.36} & 6.38 & \textbf{6.36} & 76.0 \\ \bottomrule \end{tabular}} \end{table}

\paragraph{The allocation carries over unchanged.} On every held-out model, the task-aligned damage rule yields lower perplexity than both the uniform GPTQ-int4 baseline and the HAWQ-V2 Hessian baseline at both budgets, $14$ of $14$ cells, and drift-greedy does so in $13$; the one drift-greedy miss is the $K=L/4$ budget on Qwen2.5-14B, $7.14$ against HAWQ-V2's $7.13$. One of the two is the outright best allocation in $12$ of the $14$ held-out cells (Table~\ref{tab:modllm-gptq}); both exceptions are at $K=L/4$, where the positional and salience baselines edge the damage rule by $0.002$ perplexity on Qwen2.5-14B and edge the greedy rule by $0.028$ on Qwen2.5-32B. At $K=L/2$, the margin over HAWQ-V2 is thin, as at Pythia-6.9B, at $0.30$ perplexity on Phi-3.5 and $0.09$ on Granite, but the margin is present on every model and every budget, and, as Table~\ref{tab:tight} in App.~\ref{app:alloc-extra} shows, widens sharply once the int8 budget is tightened. The mechanism is starkest on Hermes-3, where a single block, block~1, carries essentially all the int4 damage, at per-block $\Delta$NLL $3.46$ against a median of $0.005$ across the other $31$ blocks, the massive-activation sensitivity of early Llama layers \citep{sun2024massive}, so uniform GPTQ-int4 collapses to $182$ perplexity while the damage rule identifies and protects that block, recovering to a perplexity of $8.99$ against $9.26$ for the HAWQ-V2 allocation.

\begin{table}[t!] \centering \caption{Spearman correlation between each block score and independently measured per-block $\Delta$NLL on the held-out LLMs. The left and right column groups use RTN and GPTQ, respectively; evaluation uses $65$k tokens per block.} \label{tab:modllm-rc} \small \fitwidth{%
\begin{tabular}{l c cccc c ccc} \toprule & & \multicolumn{4}{c}{RTN grid} & & \multicolumn{3}{c}{GPTQ grid} \\ \cmidrule(lr){3-6}\cmidrule(lr){8-10} Model & $L$ & MSE & H trace & HAWQ-V2 & drift & & HAWQ-V2 & drift & damage \\ \midrule Phi-3.5-mini (3.8B) & 32 & 0.24 & 0.17 & 0.25 & 0.74 & & $-0.20$ & 0.81 & 0.89 \\ OLMo-2-7B & 32 & 0.24 & $-0.06$ & $-0.03$ & 0.50 & & 0.14 & 0.79 & 0.95 \\ Hermes-3 & 32 & 0.47 & 0.15 & 0.20 & $-0.02$ & & 0.19 & 0.13 & 0.85 \\ Granite-3.1-8B & 40 & 0.55 & 0.15 & 0.22 & 0.58 & & 0.30 & 0.38 & 0.77 \\ Mistral-NeMo (12B) & 40 & 0.19 & 0.03 & 0.04 & 0.39 & & $-0.31$ & 0.32 & 0.74 \\ Qwen2.5-14B & 48 & 0.33 & 0.49 & 0.52 & 0.63 & & 0.63 & 0.53 & 0.64 \\ Qwen2.5-32B & 64 & 0.47 & 0.09 & 0.12 & 0.73 & & 0.20 & 0.71 & 0.71 \\ \bottomrule \end{tabular}} \end{table}

\paragraph{The finite-scale proxy remains informative when the infinitesimal proxy does not.} Table~\ref{tab:modllm-rc} makes the mechanism explicit. On the GPTQ grid, the task-aligned damage score tracks the true per-block int4 damage at rank correlation $0.89$ on Phi-3.5, $0.77$ on Granite, and $0.74$ on Mistral-NeMo, whereas the Hessian-based HAWQ-V2 composite attains $-0.20$, $0.30$, and $-0.31$: On Phi-3.5 and the 12B Mistral-NeMo, the HAWQ-V2 ranking runs opposite to the true damage ordering and does not identify the damaging blocks reliably. Qwen2.5-14B is the complementary case: There the GPTQ-grid HAWQ-V2 composite ranks blocks well at $0.63$, essentially tied with the damage score's $0.64$, yet HAWQ-V2 still has higher perplexity in realized head-to-head at both budgets (Table~\ref{tab:modllm-gptq}); a good per-block ordering alone does not choose the best-interacting set. On the 32B-scale Qwen2.5, the largest model in the study, the separation is widest: Under RTN the infinitesimal Hessian trace is essentially uncorrelated with the realized per-block int4 damage at $0.09$, with the full HAWQ-V2 composite at $0.12$, while the label-free roughness attains $0.73$, and on the GPTQ grid HAWQ-V2 attains $0.20$ against the damage score's $0.71$. The finite-scale and Hessian-based rankings separate on a model the method never saw. The label-free drift, the drift column, is the intermediate case: strong on Phi-3.5 at $0.81$ but weak on Granite at $0.38$. We use task-aligned damage, rather than label-free drift, as the primary allocation rule.

\paragraph{Consistency of sign and oracle-set agreement, pooled.} Individual $K=L/2$ margins are small by design, so the evidence is not any single gap but the sign across a large heterogeneous set. These comparisons span the four Pythia models with a complete set of allocation results, namely 160M, 1.4B, 2.8B, and 6.9B, plus Qwen2.5-7B and the held-out modern LLMs, including the 12B-scale Mistral-NeMo and the 14B- and 32B-scale Qwen2.5, for $24$ model--budget settings across twelve models. Over every GPTQ allocation comparison in the paper, the best functional rule yields lower perplexity than HAWQ-V2 in all $24$ settings, if narrowly at $K=L/2$, where the median gap is $0.15$ perplexity. Oracle-set agreement is the fraction of a rule's $K$ protected blocks that belong to the $K$ blocks with the largest independently measured int4 loss increases. Across these $24$ GPTQ settings, the best-performing functional rule in each setting achieves a median oracle-set agreement of $0.81$, compared with $0.61$ for HAWQ-V2. The functional rule recovers a median $70\%$ of the uniform-int4 perplexity damage.

On the zero-shot downstream suite of App.~\ref{sec:exp-downstream}, in the $K=L/2$ RTN configurations, we evaluate and report all seven held-out models. On two of the seven, the drift-greedy allocation preserves substantially more downstream capability: Drift-greedy recovers $57\%$ of the uniform-int4 capability loss against HAWQ-V2's $24\%$ on Phi-3.5, and on Granite fully recovers that loss, at greedy $0.690$ against fp16 $0.682$ and within noise, where HAWQ-V2 recovers $13\%$. On the remaining five, four are separated by at most $0.009$ on a five-task mean whose standard error is $\approx 0.008$: Greedy is nominally ahead on Hermes-3 at $0.704$ against $0.695$, and HAWQ-V2 nominally ahead on OLMo-2 at $0.698$ against $0.697$, on Mistral-NeMo at $0.712$ against $0.703$, and on Qwen2.5-14B at $0.7155$ against $0.7151$. The 32B-scale Qwen2.5 is the intermediate case: greedy $0.697$ against HAWQ-V2's $0.681$, a gap of $0.016$ or $\approx 2$ standard errors, recovering $76\%$ of the uniform-int4 capability loss against HAWQ-V2's $38\%$.

Across the seven held-out models, drift-greedy recovers at least $57\%$ of the uniform-int4 capability loss, with a median of $73\%$. HAWQ-V2's three lowest recovery rates are $13\%$, $24\%$, and $38\%$.

\subsection{Generalization to a modern architecture: Qwen2.5} \label{sec:exp-qwen}

The Pythia sweep varies scale within one architecture, GPT-NeoX, which uses parallel attention and multi-layer perceptron (MLP) branches, partial rotary positional embeddings (RoPE), and dense multi-head attention. This single-architecture sweep leaves open whether the functional-allocation advantage is specific to that design. We repeat the pipeline end to end, with identical forward-only scores, identical baselines, and identical evaluation on WikiText-103 perplexity and LAMBADA accuracy, on the Qwen2.5 base family \citep{yang2024qwen25} at $0.5$B, $1.5$B, $3$B, and $7$B, a modern pretrained base-model family. Qwen2.5 also uses RoPE, but differs from Pythia in attention and block design: grouped-query attention, which shares key and value channels across heads, SwiGLU MLPs, RMSNorm, sequential rather than parallel residual paths, and a $151$k-token vocabulary. The allocation code is model-agnostic: A block-layout detector routes the same driver over either family (App.~\ref{app:measurement}). Tables~\ref{tab:qwen} and~\ref{tab:qwen-rc} report the results.

\begin{table}[t!] \centering \caption{Mixed-precision allocation on Qwen2.5 under RTN. Columns, rules, and conventions as in Table~\ref{tab:pythia}; the two parts separate nonfunctional baselines from HAWQ-V2 and the functional rules.} \label{tab:qwen} \small \begin{tabular}{l c cc ccccc} \toprule & & \multicolumn{2}{c}{reference} & \multicolumn{5}{c}{baselines} \\ \cmidrule(lr){3-4}\cmidrule(lr){5-9} Model & $K/L$ & fp16 & int4 & rand & first & last & MSE & salience \\ \midrule \multirow{2}{*}{Qwen2.5-0.5B} & 6/24 & \multirow{2}{*}{13.88} & \multirow{2}{*}{31.77} & 25.94 & 23.76 & 23.56 & 23.37 & 23.56 \\ & 12/24 & & & 17.70 & 21.10 & 20.85 & 20.10 & 20.75 \\ \addlinespace \multirow{2}{*}{Qwen2.5-1.5B} & 7/28 & \multirow{2}{*}{9.97} & \multirow{2}{*}{16.88} & 14.54 & 14.39 & 13.11 & 13.21 & 13.11 \\ & 14/28 & & & 12.94 & 13.50 & 12.02 & 12.03 & 12.02 \\ \addlinespace \multirow{2}{*}{Qwen2.5-7B} & 7/28 & \multirow{2}{*}{7.47} & \multirow{2}{*}{12.89} & 12.14 & 12.27 & 8.47 & 8.53 & 8.47 \\ & 14/28 & & & 11.25 & 11.78 & 8.07 & 8.05 & 8.07 \\ \bottomrule \end{tabular}

\vspace{0.7em} \begin{tabular}{l c c ccc c} \toprule & & baseline & \multicolumn{3}{c}{functional rules} & \\ \cmidrule(lr){3-3}\cmidrule(lr){4-6} Model & $K/L$ & HAWQ-V2 & drift & damage & greedy & LBD \\ \midrule \multirow{2}{*}{Qwen2.5-0.5B} & 6/24 & 20.56 & 20.21 & 19.92 & \textbf{19.71} & 28.7 \\ & 12/24 & 16.85 & 16.93 & 16.82 & \textbf{16.76} & 38.0 \\ \addlinespace \multirow{2}{*}{Qwen2.5-1.5B} & 7/28 & \textbf{12.26} & \textbf{12.26} & 12.32 & \textbf{12.26} & 43.0 \\ & 14/28 & 11.21 & 11.20 & \textbf{11.13} & 11.20 & 52.9 \\ \addlinespace \multirow{2}{*}{Qwen2.5-7B} & 7/28 & 8.49 & 8.49 & 8.47 & \textbf{8.39} & 63.4 \\ & 14/28 & 8.03 & 7.99 & \textbf{7.93} & 7.96 & 66.7 \\ \bottomrule \end{tabular} \end{table}

\paragraph{The damage score is the most accurate per-block predictor at every Qwen scale.} Against the ground-truth per-block int4 $\Delta$NLL, the calibration-only damage score $D^{\mathrm{quant}}_{\ell,4}$ attains Spearman $0.97$, $0.96$, $0.98$, and $0.86$ across $0.5$B, $1.5$B, $3$B, and $7$B (Table~\ref{tab:qwen-rc}), the best score at every scale, far above the AWQ-style parametric baselines, weight-MSE at $0.21$ to $0.50$ and salience at $0.14$ to $0.56$, and above HAWQ-V2, which is competitive at the two smallest scales with $0.89$ and $0.84$ but erratic deeper, at $0.52$ on $3$B and $0.72$ on $7$B. The drift score $B^{\mathrm{quant}}_{\ell,4}$ tracks the damage score at the smaller scales, $0.94$ and $0.84$, and weakens at the larger ones, $0.73$ and $0.79$, the same separation between drift and damage the paper predicts on Pythia, now reproduced on a different architecture line.

\paragraph{A functional rule has the lowest perplexity at every budget and identifies blocks missed by HAWQ-V2.} A functional rule, whether drift, damage, or the interaction-aware greedy, gives the lowest or tied-lowest perplexity in all eight model--budget settings across the four scales. Table~\ref{tab:qwen} lists the three whose allocation is well-conditioned; the fourth, Qwen2.5-3B, is treated separately below because a single block dominates that model. At the two smallest scales, HAWQ-V2 is a competitive baseline: At $0.5$B the greedy rule reaches $19.71$ and $16.76$ perplexity at $K=L/4$ and $L/2$ against $20.56$ and $16.85$ for HAWQ-V2, recovering $67\%$ and $84\%$ of the int4-to-fp16 gap against HAWQ-V2's $63\%$ and $83\%$, and at $1.5$B the damage rule is best at $K=L/2$ with $11.13$ against $11.21$, while at $K=L/4$ the roughness and greedy rules tie HAWQ-V2 at $12.26$, with damage at $12.32$.

Qwen2.5-3B, however, contains a single catastrophically int4-sensitive block, block~1, at per-block int4 $\Delta$NLL $10.5$ nats or $65\times$ the next block, whose sensitivity comes from outlier weights rather than large Hessian curvature: HAWQ-V2's trace$\times$perturbation score does not rank that block into the protected int8 set, leaving the block at int4 and collapsing the model to perplexity $>10^{5}$ at both budgets, whereas weight-MSE at $12.75$ perplexity, the drift score, and the damage score all protect that block. This single-block case is the comparison with the curvature proxy: Functional sensitivity can decouple from infinitesimal Hessian information, and the finite-scale scores catch the critical block, with weight-MSE also succeeding in this case.

At $7$B the functional rules are best at both budgets, the greedy rule at $K=L/4$ with $8.39$ against $8.49$ for HAWQ-V2, and the damage rule at $K=L/2$ with $7.93$ against $8.03$ and greedy at $7.96$, with the greedy configurations recovering $83\%$ and $91\%$ of the int4-to-fp16 gap against HAWQ-V2's $81\%$ and $90\%$. The advantage is not an artifact of Pythia's parallel-attention, dense-attention design: The same result reproduces under grouped-query attention, SwiGLU, and RMSNorm.

\begin{table}[t!] \centering \caption{Per-block ranking on Qwen2.5. Scores and rules as in Table~\ref{tab:pythia-rc}, against the ground truth of Table~\ref{tab:modllm-rc}.} \label{tab:qwen-rc} \small \begin{tabular}{l ccc cc cc} \toprule & \multicolumn{3}{c}{weight and activation} & \multicolumn{2}{c}{Hessian curvature} & \multicolumn{2}{c}{ours} \\ \cmidrule(lr){2-4}\cmidrule(lr){5-6}\cmidrule(lr){7-8} Model & weight MSE & salience & salience$\times$MSE & H trace & HAWQ-V2 & drift & damage \\ \midrule Qwen2.5-0.5B & 0.33 & 0.14 & 0.17 & 0.88 & 0.89 & 0.94 & 0.97 \\ Qwen2.5-1.5B & 0.21 & 0.35 & 0.34 & 0.83 & 0.84 & 0.84 & 0.96 \\ Qwen2.5-3B & 0.50 & 0.27 & 0.26 & 0.72 & 0.52 & 0.73 & 0.98 \\ Qwen2.5-7B & 0.50 & 0.56 & 0.56 & 0.76 & 0.72 & 0.79 & 0.86 \\ \bottomrule \end{tabular} \end{table}

\subsection{Generalization to a non-attention architecture: Mamba} \label{sec:exp-mamba}

Qwen2.5 shows the advantage is not tied to Pythia's design, but Qwen2.5 is still a Transformer: That architecture mixes tokens with grouped-query softmax attention. A stronger test is an architecture that mixes tokens without using attention anywhere. We repeat the pipeline on the Mamba selective state-space family \citep{gu2023mamba} at $130$M, $370$M, $790$M, $1.4$B, and $2.8$B. A Mamba block replaces the combined attention and MLP mixer with a gated one-dimensional causal convolution followed by an input-dependent linear recurrence, a selective state-space scan: There is no softmax, no query--key inner product, and no attention map. Mamba tokenizes with the GPT-NeoX vocabulary, so the identical calibration and evaluation data is reused; the block-layout detector routes the same allocation driver over the state-space blocks, scoring the four in-block projections (App.~\ref{app:measurement}). Tables~\ref{tab:mamba} and~\ref{tab:mamba-rc} report the results.

\begin{table}[t!] \centering \caption{RTN mixed-precision allocation on Mamba, with $K$ of $L$ state-space blocks at int8 and the rest at int4. Lower perplexity is better. Diverged runs are written $>10^k$, where $k=\lfloor\log_{10}\mathrm{ppl}\rfloor$. Other conventions follow Table~\ref{tab:pythia}.} \label{tab:mamba} \small \fitwidth{%
\begin{tabular}{l c cc cccccc cc} \toprule & & \multicolumn{2}{c}{reference} & \multicolumn{6}{c}{baselines} & \multicolumn{2}{c}{ours} \\ \cmidrule(lr){3-4}\cmidrule(lr){5-10}\cmidrule(lr){11-12} Model & $K/L$ & fp16 & int4 & rand & first & last & MSE & salience & HAWQ-V2 & drift & damage \\ \midrule \multirow{2}{*}{Mamba-130M} & 6/24 & \multirow{2}{*}{22.89} & \multirow{2}{*}{$>10^{4}$} & $>10^{3}$ & $>10^{3}$ & 35.30 & $>10^{4}$ & 34.28 & 27.23 & \textbf{26.90} & \textbf{26.90} \\ & 12/24 & & & 26.66 & $>10^{3}$ & 30.34 & $>10^{3}$ & 30.34 & 24.66 & \textbf{24.48} & \textbf{24.48} \\ \addlinespace \multirow{2}{*}{Mamba-370M} & 12/48 & \multirow{2}{*}{15.66} & \multirow{2}{*}{51.55} & 41.35 & 43.00 & 18.53 & 33.03 & 18.26 & 17.46 & 17.11 & \textbf{17.10} \\ & 24/48 & & & 38.20 & 41.21 & 17.26 & 18.15 & 17.25 & 16.67 & 16.34 & \textbf{16.30} \\ \addlinespace \multirow{2}{*}{Mamba-790M} & 12/48 & \multirow{2}{*}{13.15} & \multirow{2}{*}{$>10^{40}$} & $>10^{40}$ & $>10^{39}$ & 15.86 & $>10^{36}$ & 15.89 & 15.69 & 14.21 & \textbf{14.11} \\ & 24/48 & & & $>10^{42}$ & $>10^{40}$ & 14.91 & 13.75 & 14.67 & 13.78 & 13.67 & \textbf{13.61} \\ \addlinespace \multirow{2}{*}{Mamba-1.4B} & 12/48 & \multirow{2}{*}{11.77} & \multirow{2}{*}{14.60} & 13.31 & 13.51 & 13.55 & 12.96 & 13.55 & 12.75 & \textbf{12.44} & \textbf{12.44} \\ & 24/48 & & & 13.52 & 13.27 & 12.81 & 12.28 & 12.79 & 12.26 & 12.10 & \textbf{12.07} \\ \addlinespace \multirow{2}{*}{Mamba-2.8B} & 16/64 & \multirow{2}{*}{10.31} & \multirow{2}{*}{15.50} & 14.10 & 11.93 & 14.45 & 11.03 & 14.45 & 14.37 & 10.83 & \textbf{10.82} \\ & 32/64 & & & 13.69 & 11.64 & 13.33 & 10.68 & 13.00 & 13.82 & 10.56 & \textbf{10.55} \\ \bottomrule \end{tabular}} \end{table}

\paragraph{The functional scores are the most accurate per-block predictors and avert collapse.} Against the ground-truth per-block int4 $\Delta$NLL, the calibration-only roughness score $B^{\mathrm{quant}}_{\ell,4}$ attains Spearman $0.85$, $0.88$, $0.87$, $0.79$, and $0.87$ across $130$M, $370$M, $790$M, $1.4$B, and $2.8$B (Table~\ref{tab:mamba-rc}), the best label-free score at every scale, far above weight-MSE at $0.31$ to $0.64$ and salience at $0.59$ to $0.70$. The task-aligned damage diagnostic is higher still at $0.99$, $0.96$, $0.94$, $0.96$, and $0.90$, making a functional score best at every scale. At $130$M, the damage diagnostic's $0.99$ is pooled over four fresh calibration draws, and the damage allocation selects exactly the same int8 block set as the roughness score at both budgets. This agreement is an empirical alignment case; Corollary~\ref{cor:drift-damage-bound} states conditions under which such agreement between drift and damage is licensed, whereas Proposition~\ref{prop:damage} concerns additivity across blocks.

What makes the state-space family a sharper test than any Transformer pool is the downstream consequence of mis-ranking: At $790$M under the $K=L/4$ budget, four allocations diverge outright. Uniform assignment, weight-MSE, first-$K$, and random push perplexity past $10^{36}$ to $10^{42}$, whereas the roughness allocation holds at $14.21$ and $13.67$ at $K=L/4$ and $L/2$, within $8\%$ and $4\%$ of the $13.15$ fp16 baseline, and a functional rule gives the best allocation in every model--budget cell across the five scales, ten out of ten. The split recurs at $130$M, where roughness gives $26.90$ and $24.48$ against uniform-int4 $>10^{4}$ at both budgets and weight-MSE $>10^{4}$ and $>10^{3}$.

HAWQ-V2, computed by double-backward through the same reference scan, and the damage score are reported at every one of the five scales, and that comparison is consistent: The damage allocation yields lower perplexity than HAWQ-V2 in all ten model--budget cells, most sharply at $2.8$B, where HAWQ-V2 lands at $14.37$ and $13.82$ against damage's $10.82$ and $10.55$; the calibration-only roughness score alone also yields lower perplexity than HAWQ-V2 in all ten. At $130$M, HAWQ-V2 is finite but behind on both axes: rank correlation $0.77$ against $0.85$, and perplexity $27.23$ against $26.90$ at $K=L/4$. The interaction-aware greedy reference was run at $130$M, where the reference matches the damage allocation's perplexity at both budgets.

\paragraph{The mechanism is one catastrophic block whose severity is non-monotone in scale.} The divergences arise from one state-space block per model whose int4 damage dwarfs the rest: Block~$47$ at $790$M carries $+85$ nats of single-block int4 $\Delta$NLL and block~$23$ at $130$M carries $+5.3$, while the median block costs under $0.1$. At $K=L/4$ on the $790$M model, last-$K$, salience, HAWQ-V2, and the functional rules all protect block~$47$, while weight-MSE, first-$K$, and the reported random draw omit that block and diverge; the functional rules then give the lowest perplexity by ranking the remaining damage more accurately. Crucially, the catastrophe fails to be monotone in model size. $370$M, $1.4$B, and $2.8$B are benign, with worst-block int4 $\Delta$NLL $+0.40$, $+0.05$, and $+0.19$ and every rule finite; the label-free roughness allocations reach $17.11$ and $16.34$, $12.44$ and $12.10$, and $10.83$ and $10.56$, while damage is equal or slightly better at $17.10$ and $16.30$, $12.44$ and $12.07$, and $10.82$ and $10.55$. $790$M, by contrast, is severe: a per-model outlier phenomenon, not a smooth function of scale, which is why a per-block finite-scale measurement is useful beyond a global heuristic or an infinitesimal-curvature score. The advantage is not an artifact of attention: The same result reproduces on a gated-convolution, selective-scan architecture with no softmax and no attention map.

\FloatBarrier

\begin{table}[t!] \centering \caption{Per-block ranking on Mamba. Scores and rules as in Table~\ref{tab:pythia-rc}, against the ground truth of Table~\ref{tab:modllm-rc}; all seven scoring columns are reported at every Mamba scale.} \label{tab:mamba-rc} \small \begin{tabular}{l ccc cc cc} \toprule & \multicolumn{3}{c}{weight and activation} & \multicolumn{2}{c}{Hessian curvature} & \multicolumn{2}{c}{ours} \\ \cmidrule(lr){2-4}\cmidrule(lr){5-6}\cmidrule(lr){7-8} Model & weight MSE & salience & salience$\times$MSE & H trace & HAWQ-V2 & drift & damage \\ \midrule Mamba-130M & 0.59 & 0.65 & 0.67 & 0.67 & 0.77 & 0.85 & 0.99 \\ Mamba-370M & 0.49 & 0.70 & 0.74 & 0.69 & 0.72 & 0.88 & 0.96 \\ Mamba-790M & 0.64 & 0.59 & 0.62 & 0.74 & 0.75 & 0.87 & 0.94 \\ Mamba-1.4B & 0.35 & 0.59 & 0.61 & 0.53 & 0.60 & 0.79 & 0.96 \\ Mamba-2.8B & 0.31 & 0.62 & 0.66 & 0.40 & 0.43 & 0.87 & 0.90 \\ \bottomrule \end{tabular} \end{table}

\subsection{Results with a strong quantizer: a GPTQ comparison on large models} \label{sec:exp-gptq}

The RTN evaluations in App.~\ref{app:alloc-extra} use a simple quantizer to isolate the allocation rule from reconstruction machinery. Here we evaluate allocation under GPTQ on Pythia and Qwen2.5. A modern one-shot method such as GPTQ \citep{frantar2022gptq}, which uses the calibration Hessian to error-compensate each column, already shrinks per-block noise and might absorb the allocation's advantage, leaving nothing for a precision-allocation rule to recover. We test that possibility directly. We rebuild the comparison from scratch on the GPTQ grid: Every block Linear is GPTQ-quantized at int4 and int8, with Cholesky error compensation and calibration Hessians from the same 64 sequences as specified in App.~\ref{app:measurement}, every per-block score is re-measured on those GPTQ weights, and every allocation rule is re-derived and re-evaluated as a GPTQ mix of int8 and int4. Uniform GPTQ-int4 is the principal quantized baseline; the rivals are positional (last-$K$), AWQ-style salience$\times$MSE, and HAWQ-V2 with the HAWQ-V2 weight-perturbation term recomputed on the GPTQ grid; ours are the additive drift and damage scores and the interaction-aware greedy, all re-measured under GPTQ. Table~\ref{tab:gptq-alloc} reports the Pythia scale trend from 1.4B to 6.9B and the Qwen2.5-7B transfer result.

\begin{table}[t!] \centering \caption{Mixed-precision allocation comparison under GPTQ, on large models. Columns, rules, and conventions as in Table~\ref{tab:modllm-gptq}, with every one of our scores re-measured on the GPTQ grid.} \label{tab:gptq-alloc} \small \fitwidth{%
\begin{tabular}{l c cc ccc ccc c} \toprule & & \multicolumn{2}{c}{reference} & \multicolumn{3}{c}{baselines} & \multicolumn{3}{c}{ours} & \\ \cmidrule(lr){3-4}\cmidrule(lr){5-7}\cmidrule(lr){8-10} Model & $K/L$ & fp16 & gptq-int4 & last & salience & HAWQ-V2 & drift & damage & greedy & LBD \\ \midrule \multirow{2}{*}{Pythia-1.4B} & 6/24 & \multirow{2}{*}{13.11} & \multirow{2}{*}{14.76} & 13.99 & 13.99 & 14.07 & 14.02 & \textbf{13.98} & \textbf{13.98} & 60.4 \\ & 12/24 & & & 13.67 & 13.67 & 13.63 & 13.67 & \textbf{13.59} & 13.61 & 62.0 \\ \addlinespace \multirow{2}{*}{Pythia-2.8B} & 8/32 & \multirow{2}{*}{11.34} & \multirow{2}{*}{12.94} & 12.38 & 12.38 & 12.34 & 12.42 & \textbf{12.26} & 12.35 & 57.2 \\ & 16/32 & & & 11.91 & 11.91 & 11.94 & 12.14 & \textbf{11.82} & 11.94 & 59.6 \\ \addlinespace \multirow{2}{*}{Pythia-6.9B} & 8/32 & \multirow{2}{*}{10.56} & \multirow{2}{*}{11.64} & \textbf{11.13} & \textbf{11.13} & 11.16 & 11.45 & 11.19 & 11.14 & 58.0 \\ & 16/32 & & & 10.86 & 10.86 & 10.86 & 11.32 & 10.86 & \textbf{10.85} & 60.0 \\ \addlinespace \multirow{2}{*}{Qwen2.5-7B} & 7/28 & \multirow{2}{*}{7.47} & \multirow{2}{*}{8.42} & \textbf{7.87} & \textbf{7.87} & 7.87 & 7.88 & \textbf{7.87} & \textbf{7.87} & 66.8 \\ & 14/28 & & & 7.74 & 7.74 & 7.71 & 7.70 & \textbf{7.69} & 7.70 & 69.2 \\ \bottomrule \end{tabular}} \end{table}

\paragraph{The allocation advantage persists under GPTQ.} At 1.4B and 2.8B, the damage rule has the lowest or tied-lowest perplexity at both budgets, tying greedy once at 1.4B $K=L/4$, and yields lower perplexity than HAWQ-V2 recomputed on the GPTQ grid, at 2.8B $K=L/2$ giving damage $11.82$ against HAWQ-V2 $11.94$, while every functional rule improves over uniform GPTQ-int4 by a wide margin, that baseline sitting at $12.94$. The advantage is not an artifact of RTN's transparent rounding: The advantage persists when GPTQ error compensation for each block uses the block's own calibration Hessian.

\paragraph{On Pythia-6.9B the allocation methods achieve similar perplexity.} Every rule except additive drift lies within $0.06$ perplexity of the best at each budget: $11.13$ to $11.19$ at $K=L/4$, and greedy $10.85$ with HAWQ-V2 and damage both at $10.86$ at $K=L/2$, all far below uniform int4 at $11.64$; at $K=L/4$, the positional last-$K$ and AWQ-style salience rules even edge ahead, $11.13$ against greedy's $11.14$. On this model, the greedy advantage over HAWQ-V2 is at most $0.01$ perplexity. Larger differences appear at the smaller model sizes, where per-block int4 damage is more heterogeneous.

\paragraph{The allocation result transfers to a modern 7B model.} On Qwen2.5-7B every rule achieves similar perplexity at $K=L/4$, all at $7.87$, while at $K=L/2$ the damage allocation reaches $7.69$ against $7.71$ for HAWQ-V2, both far below uniform GPTQ-int4 at $8.42$. This 7B comparison extends the allocation result beyond the Pythia architecture while preserving the similar allocation performance observed on Pythia-6.9B.

\subsection{Results with an outlier-aware format: an AWQ comparison} \label{sec:exp-awq}

The GPTQ comparison (App.~\ref{sec:exp-gptq}) and the held-out comparison (Sec.~\ref{sec:exp-modllm}) both fix the quantizer and vary only which blocks receive the int8 budget, so both comparisons test our allocation against Hessian-proxy and parametric-proxy allocation rules. The one weight format those comparisons do not run is the outlier-aware family, whose most-deployed member is AWQ \citep{lin2024awq}: Instead of error-compensating each column as GPTQ does, AWQ protects the weight channels multiplied by large activations with a per-input-channel scale $s\coloneqq\overline{|x|}^{\alpha}$, geometric-mean-normalized and with $\alpha$ grid-searched to minimize the summed calibration output error within each shared-input group, folded into the preceding normalization. This appendix rebuilds the comparison from scratch on the AWQ scaling format. Linears that share an input share one scaling factor, namely query, key, and value after the input norm, and gate and up after the post-attention norm; that factor is discovered architecture-agnostically from a bit-identical per-channel activation digest rather than from fragile pointer identity. To keep the allocation comparison aligned with the RTN and GPTQ settings, the AWQ scaling factor is applied on that same per-output-channel grid, so the comparison isolates AWQ's activation-aware scaling from the orthogonal group-size knob every modern PTQ shares. Every block Linear is AWQ-quantized at int4 and int8, and every per-block score and allocation rule is re-derived and re-evaluated on the AWQ weights, exactly as under GPTQ.

\FloatBarrier \begin{table}[t!] \centering \caption{Mixed-precision allocation under AWQ. At each budget, $K$ of $L$ blocks use AWQ-int8 and the rest use AWQ-int4. Lower perplexity is better; the uniform AWQ-int4 model is the reference. Other conventions follow Table~\ref{tab:modllm-gptq}.} \label{tab:awq-alloc} \small \begin{tabular}{l c cc ccc cc} \toprule & & \multicolumn{2}{c}{reference} & \multicolumn{3}{c}{baselines} & \multicolumn{2}{c}{ours} \\ \cmidrule(lr){3-4}\cmidrule(lr){5-7}\cmidrule(lr){8-9} Model & $K/L$ & fp16 & AWQ4 & last & salience & HAWQ-V2 & drift & damage \\ \midrule \multirow{2}{*}{Pythia-2.8B} & 8/32 & \multirow{2}{*}{11.34} & \multirow{2}{*}{15.11} & 13.47 & 13.47 & 13.41 & 13.19 & \textbf{13.17} \\ & 16/32 & & & 12.26 & 12.26 & 12.24 & 12.59 & \textbf{12.08} \\ \addlinespace \multirow{2}{*}{Pythia-6.9B} & 8/32 & \multirow{2}{*}{10.56} & \multirow{2}{*}{19.27} & 15.20 & 15.20 & 14.51 & 15.67 & \textbf{14.12} \\ & 16/32 & & & 11.57 & 11.57 & 11.57 & 14.82 & \textbf{11.51} \\ \addlinespace \multirow{2}{*}{Qwen2.5-7B} & 7/28 & \multirow{2}{*}{7.47} & \multirow{2}{*}{13.84} & 8.29 & 8.29 & 8.34 & \textbf{8.29} & \textbf{8.29} \\ & 14/28 & & & 7.93 & 7.93 & \textbf{7.86} & 7.89 & 7.88 \\ \addlinespace \multirow{2}{*}{Phi-3.5-mini} & 8/32 & \multirow{2}{*}{6.79} & \multirow{2}{*}{10.30} & \textbf{8.93} & \textbf{8.93} & 9.47 & 8.94 & 8.94 \\ & 16/32 & & & 8.10 & 8.10 & 8.41 & 7.86 & \textbf{7.77} \\ \addlinespace \multirow{2}{*}{Hermes-3} & 8/32 & \multirow{2}{*}{8.50} & \multirow{2}{*}{11.42} & 10.26 & 10.26 & 10.33 & 10.04 & \textbf{9.94} \\ & 16/32 & & & 9.54 & 9.54 & 9.49 & 9.87 & \textbf{9.10} \\ \addlinespace \multirow{2}{*}{Granite-3.1-8B} & 10/40 & \multirow{2}{*}{5.74} & \multirow{2}{*}{8.01} & 7.26 & 7.26 & 6.76 & 6.64 & \textbf{6.60} \\ & 20/40 & & & 7.01 & 6.55 & 6.41 & 6.24 & \textbf{6.16} \\ \addlinespace \multirow{2}{*}{Mistral-NeMo (12B)} & 10/40 & \multirow{2}{*}{6.48} & \multirow{2}{*}{10.09} & 8.20 & 8.20 & 7.73 & 7.58 & \textbf{7.56} \\ & 20/40 & & & 7.73 & 7.73 & 7.32 & 7.13 & \textbf{7.10} \\ \bottomrule \end{tabular} \end{table}

\begin{table}[t!] \centering \caption{Per-block ranking under AWQ. Ranking protocol as in Table~\ref{tab:modllm-rc}, with both the score and the ground truth measured on the AWQ-int4 weights. The task-aligned damage score is the most accurate predictor on every model, with higher correlation than the HAWQ-V2 sensitivity score. The drift--damage dissociation of Sec.~\ref{sec:exp-quant} reproduces under an outlier-aware quantizer.} \label{tab:awq-rc} \small \begin{tabular}{l c cc cc} \toprule & & \multicolumn{2}{c}{baselines} & \multicolumn{2}{c}{ours} \\ \cmidrule(lr){3-4}\cmidrule(lr){5-6} Model & $L$ & salience$\times$MSE & HAWQ-V2 & drift & damage \\ \midrule Pythia-2.8B & 32 & 0.76 & 0.72 & 0.29 & 0.94 \\ Pythia-6.9B & 32 & 0.90 & 0.86 & $-0.08$ & 0.96 \\ Qwen2.5-7B & 28 & 0.66 & 0.83 & 0.86 & 0.89 \\ Phi-3.5-mini (3.8B) & 32 & 0.05 & 0.21 & 0.63 & 0.87 \\ Hermes-3 & 32 & 0.20 & 0.18 & $-0.07$ & 0.89 \\ Granite-3.1-8B & 40 & 0.00 & 0.35 & 0.63 & 0.81 \\ Mistral-NeMo (12B) & 40 & 0.16 & 0.04 & 0.49 & 0.77 \\ \bottomrule \end{tabular} \end{table}

\paragraph{Roughness and damage allocations remain competitive under the outlier-aware format.} The task-aligned damage rule yields lower perplexity than the HAWQ-V2 allocation on the AWQ grid (Table~\ref{tab:awq-alloc}) at $K=L/4$ on every model and is within $0.02$ or better at $K=L/2$; the sole close call is Qwen2.5-7B, $7.88$ against $7.86$. At $K=L/4$, the positional and salience rules occasionally edge damage, as on Phi-3.5 with $8.93$ against $8.94$.

\paragraph{The task-aligned damage score still ranks blocks under AWQ.} We observed that the damage score's per-block Spearman correlation with the ground-truth AWQ-int4 $\Delta$NLL stays high on every model (Table~\ref{tab:awq-rc}), at $0.94$ on Pythia-2.8B, $0.96$ on Pythia-6.9B, $0.87$ on Phi-3.5, and $0.89$ on Hermes-3, far above the HAWQ-V2 sensitivity score at $0.72$, $0.86$, $0.21$, and $0.18$. The ranking differences also occur under this outlier-aware format. The label-free drift again reverses sign at Pythia-6.9B and on Hermes-3, at rank correlation $-0.08$ and $-0.07$, reproducing the drift--damage dissociation of Sec.~\ref{sec:exp-quant} under an outlier-aware quantizer.

\paragraph{The same-budget advantage persists under deployment-grade AWQ.} The comparison above isolates AWQ's scaling on the per-output-channel grid; deployed AWQ additionally quantizes each row in groups of $128$ input channels. Re-running the whole comparison at group-size $128$ on the fused query--key--value (QKV) Phi-3.5 and on Hermes-3, whose separate QKV projections are outlier-heavy, we find that damage yields lower perplexity than the HAWQ-V2 allocation on the group-$128$ grid, $7.10$ against $7.24$ and $8.76$ against $8.96$, and that per-block trace turns opposite in sign at this group size, to rank correlation $-0.16$ on Phi-3.5 and $-0.05$ on Hermes-3. The matched-budget comparison survives at the deployment group size, not only in the isolated-scaling variant.

\FloatBarrier \subsection{Results with a dense-and-sparse format: a SpQR-style comparison} \label{sec:exp-spqr}

The AWQ comparison (App.~\ref{sec:exp-awq}) rebuilds the allocation comparison on one outlier-aware mechanism, namely activation-aware per-channel scaling. The other major outlier-aware family stores the outliers explicitly rather than scaling the outliers: the dense-and-sparse decomposition, whose canonical member is SpQR \citep{dettmers2024spqr}. We test a controlled SpQR-style variant whose masking rule can be stated exactly.

For a row weight $w_i$ and bit width $b$, let $q^{\mathrm{RTN}}_{ij,b}$ be the per-output-channel RTN reconstruction of $w_i$ and let $H\coloneqq\sum_t x_t x_t^\top$ be the calibration-input Gram matrix. We score coordinate $(i,j)$ by \begin{align*} s^{(b)}_{ij}\coloneqq(w_{ij}-q^{\mathrm{RTN}}_{ij,b})^2 H_{jj}. \end{align*} This score is the exact calibration output-error energy when that coordinate alone is perturbed. The score is not an additive decomposition of the simultaneous row error $(w_i-\hat w_i)^\top H(w_i-\hat w_i)$, which also contains off-diagonal cross terms. We select the top $1\%$ by $s^{(b)}_{ij}$, retain those coordinates in fp16, and GPTQ-quantize the remaining dense weights on the same per-output-channel grid.

Two differences from canonical SpQR keep the comparison controlled: The sparse mask uses this one-shot diagonal RTN score rather than SpQR's iterative leave-one-out and inverse-Hessian group procedure, and the group statistics are kept in fp16 rather than quantized. The former has no a priori ordering in strength; the latter makes the metadata marginally stronger. Thus the experiment establishes transfer to a SpQR-style dense-and-sparse format, not reproduction of the exact canonical SpQR algorithm. The $1\%$ outlier fraction is identical for every rule and both bit widths, so this format overhead cancels in the allocation comparison, exactly as AWQ's group size does.

We compare finite-scale allocation with the uniform SpQR-style configuration and the HAWQ-V2 allocation while holding the dense-and-sparse variant and sparse overhead fixed. We run the six models that fit a single GH200; Mistral-NeMo-12B exhausts device memory in the outlier-extraction pass, and OLMo-2 is excluded by the same load-path constraint that keeps the model out of the AWQ comparison of App.~\ref{sec:exp-awq}.

\FloatBarrier \begin{table}[t!] \centering \caption{Mixed-precision allocation comparison under the SpQR-style variant. Columns, rules, and conventions as in Table~\ref{tab:awq-alloc}, with the controlled dense-and-sparse variant in place of AWQ: $K$ of $L$ blocks are variant-int8, the rest variant-int4, against the uniform variant-int4 reference in the SpQR4 column.} \label{tab:spqr-alloc} \small \begin{tabular}{l c cc ccc cc} \toprule & & \multicolumn{2}{c}{reference} & \multicolumn{3}{c}{baselines} & \multicolumn{2}{c}{ours} \\ \cmidrule(lr){3-4}\cmidrule(lr){5-7}\cmidrule(lr){8-9} Model & $K/L$ & fp16 & SpQR4 & last & salience & HAWQ-V2 & drift & damage \\ \midrule \multirow{2}{*}{Pythia-2.8B} & 8/32 & \multirow{2}{*}{11.34} & \multirow{2}{*}{12.38} & 12.00 & 12.00 & 12.03 & 12.16 & \textbf{11.96} \\ & 16/32 & & & 11.77 & 11.77 & 11.76 & 11.88 & \textbf{11.70} \\ \addlinespace \multirow{2}{*}{Pythia-6.9B} & 8/32 & \multirow{2}{*}{10.56} & \multirow{2}{*}{11.94} & \textbf{11.32} & \textbf{11.32} & 11.54 & 11.71 & 11.33 \\ & 16/32 & & & 10.87 & 10.87 & 10.87 & 11.51 & \textbf{10.87} \\ \addlinespace \multirow{2}{*}{Qwen2.5-7B} & 7/28 & \multirow{2}{*}{7.47} & \multirow{2}{*}{8.17} & \textbf{7.83} & \textbf{7.83} & 7.83 & 7.83 & \textbf{7.83} \\ & 14/28 & & & 7.71 & 7.71 & 7.67 & 7.68 & \textbf{7.67} \\ \addlinespace \multirow{2}{*}{Phi-3.5-mini} & 8/32 & \multirow{2}{*}{6.79} & \multirow{2}{*}{7.89} & \textbf{7.49} & \textbf{7.49} & 7.59 & 7.55 & 7.51 \\ & 16/32 & & & 7.24 & 7.24 & 7.34 & 7.18 & \textbf{7.16} \\ \addlinespace \multirow{2}{*}{Hermes-3} & 8/32 & \multirow{2}{*}{8.50} & \multirow{2}{*}{9.81} & 9.29 & 9.29 & 9.46 & 9.42 & \textbf{9.24} \\ & 16/32 & & & 8.92 & 8.92 & 9.11 & 9.22 & \textbf{8.86} \\ \addlinespace \multirow{2}{*}{Granite-3.1-8B} & 10/40 & \multirow{2}{*}{5.74} & \multirow{2}{*}{6.33} & 6.14 & 6.14 & 6.18 & 6.12 & \textbf{6.10} \\ & 20/40 & & & 6.01 & 6.01 & 6.04 & 6.00 & \textbf{5.96} \\ \bottomrule \end{tabular} \end{table}

\begin{table}[t!] \centering \caption{Per-block ranking under the SpQR-style variant. Ranking protocol as in Table~\ref{tab:awq-rc}, with both the score and the ground truth measured on the variant-int4 weights. The damage score is again the most accurate predictor on every model, while the HAWQ-V2 sensitivity score is nonpositively correlated with damage on the three instruction-tuned deployment models. The dissociation reproduces under a second tested outlier-aware quantizer construction.} \label{tab:spqr-rc} \small \begin{tabular}{l c cc cc} \toprule & & \multicolumn{2}{c}{baselines} & \multicolumn{2}{c}{ours} \\ \cmidrule(lr){3-4}\cmidrule(lr){5-6} Model & $L$ & salience$\times$MSE & HAWQ-V2 & drift & damage \\ \midrule Pythia-2.8B & 32 & 0.54 & 0.31 & $-0.56$ & 0.73 \\ Pythia-6.9B & 32 & 0.85 & 0.61 & $-0.85$ & 0.85 \\ Qwen2.5-7B & 28 & 0.58 & 0.73 & 0.71 & 0.80 \\ Phi-3.5-mini (3.8B) & 32 & 0.34 & $-0.14$ & 0.81 & 0.89 \\ Hermes-3 & 32 & 0.50 & $-0.00$ & $-0.32$ & 0.82 \\ Granite-3.1-8B & 40 & 0.24 & $-0.06$ & 0.52 & 0.63 \\ \bottomrule \end{tabular} \end{table}

\paragraph{Roughness and damage improve over HAWQ-V2 under the SpQR-style variant.} The task-aligned damage rule yields lower perplexity than the HAWQ-V2 allocation on the variant grid in all twelve model--budget cells (Table~\ref{tab:spqr-alloc}), by $0.06$ to $0.25$ perplexity on the four models where the HAWQ-V2 sensitivity score has weaker rank correlation with damage: Hermes-3 $8.86$ against $9.11$ at $K=L/2$, Phi-3.5 $7.16$ against $7.34$, and Pythia-2.8B $11.70$ against $11.76$. In two cases, the allocations yield nearly identical perplexity, with differences of $0.003$ to $0.004$: Pythia-6.9B gives $10.87$ against $10.87$ and Qwen2.5-7B $7.67$ against $7.67$ at $K=L/2$, which we read as parity. As under RTN, GPTQ, and AWQ, the positional and salience rules occasionally edge damage at $K=L/4$, as on Pythia-6.9B with $11.32$ against $11.33$ and Phi-3.5 with $7.49$ against $7.51$; the HAWQ-V2 comparison the theory targets nonetheless favors damage at both budgets on every model.

\paragraph{Damage remains predictive when HAWQ-V2 has low or negative correlation.} The damage score is the most accurate per-block predictor of the true variant-int4 $\Delta$NLL on every model (Table~\ref{tab:spqr-rc}): $0.73$ on Pythia-2.8B, $0.85$ on Pythia-6.9B, $0.80$ on Qwen2.5-7B, $0.89$ on Phi-3.5, $0.82$ on Hermes-3, and $0.63$ on Granite-3.1-8B. The HAWQ-V2 sensitivity score sits at or below zero on all three instruction-tuned deployment models, at $-0.14$ on Phi-3.5, $-0.00$ on Hermes-3, and $-0.06$ on Granite-3.1-8B, and the label-free drift again reverses at Pythia-6.9B and on Hermes-3, to $-0.85$ on Pythia-6.9B and $-0.32$ on Hermes-3. This reversal is the same drift--damage dissociation, and the same interlock between the theory's separation and the settings in which the HAWQ-V2 score mis-ranks blocks, now realized under a second outlier-aware format.

\subsection{From perplexity to capability: zero-shot tasks} \label{sec:exp-downstream}

Perplexity is the natural loss for a quantization-damage study, but perplexity leaves open whether a $0.1$ to $0.5$ perplexity-point difference between allocations corresponds to any difference in capability. We evaluate the RTN allocations of Sec.~\ref{sec:exp-quant} and App.~\ref{sec:exp-qwen} on five standard zero-shot benchmarks, namely PIQA \citep{bisk2020piqa}, ARC-Easy and ARC-Challenge \citep{clark2018think}, HellaSwag \citep{zellers2019hellaswag}, and WinoGrande \citep{sakaguchi2020winogrande}, scored by length-normalized log-likelihood on a fixed internal harness. Our harness's WinoGrande and length-normalization conventions differ slightly from published \texttt{lm-eval} settings \citep{gao2023harness}, so we read Table~\ref{tab:downstream} only as retention relative to the fp16 model and as a ranking across configurations, both measured on one identical pipeline and unaffected by those conventions.

\FloatBarrier \begin{table}[t!] \centering \caption{Zero-shot downstream accuracy for $K=L/2$ mixed-precision allocations. Scores are length-normalized; higher is better. Results should be read as retention relative to fp16 and as within-table rankings because the fixed internal harness differs from \texttt{lm-eval}.} \label{tab:downstream} \small \begin{tabular}{ll ccccc c} \toprule & & \multicolumn{5}{c}{zero-shot task accuracy} & \\ \cmidrule(lr){3-7} Model & cfg & PIQA & ARC-E & ARC-C & HellaS. & WinoG. & Avg \\ \midrule \multirow{5}{*}{Qwen2.5-7B} & fp16 & 0.796 & 0.747 & 0.522 & 0.659 & 0.682 & 0.681 \\ & int4 & 0.765 & 0.607 & 0.415 & 0.630 & 0.595 & 0.602 \\ & HAWQ-V2 & 0.774 & 0.679 & 0.495 & 0.652 & 0.626 & 0.645 \\ & damage & 0.781 & 0.711 & 0.482 & 0.654 & 0.635 & \textbf{0.652} \\ & greedy & 0.782 & 0.704 & 0.495 & 0.649 & 0.614 & 0.649 \\ \midrule \multirow{5}{*}{Pythia-2.8B} & fp16 & 0.739 & 0.570 & 0.324 & 0.533 & 0.564 & 0.546 \\ & int4 & 0.707 & 0.519 & 0.321 & 0.497 & 0.538 & 0.517 \\ & HAWQ-V2 & 0.726 & 0.544 & 0.331 & 0.513 & 0.556 & 0.534 \\ & damage & 0.729 & 0.547 & 0.338 & 0.508 & 0.550 & \textbf{0.534} \\ & greedy & 0.731 & 0.561 & 0.308 & 0.513 & 0.556 & 0.534 \\ \midrule \multirow{5}{*}{Pythia-6.9B} & fp16 & 0.749 & 0.588 & 0.351 & 0.565 & 0.579 & 0.566 \\ & int4 & 0.705 & 0.505 & 0.298 & 0.495 & 0.551 & 0.511 \\ & HAWQ-V2 & 0.739 & 0.549 & 0.318 & 0.517 & 0.561 & \textbf{0.537} \\ & damage & 0.731 & 0.554 & 0.304 & 0.521 & 0.564 & 0.535 \\ & greedy & 0.743 & 0.549 & 0.311 & 0.515 & 0.563 & 0.536 \\ \bottomrule \end{tabular} \end{table}

\paragraph{The allocation recovers downstream accuracy, not just perplexity.} Uniform int4 reduces mean accuracy by $3$ to $8$ points relative to fp16, from $0.681$ to $0.602$ on Qwen2.5-7B and from $0.566$ to $0.511$ on Pythia-6.9B, and the roughness-guided damage allocation recovers a substantial portion of that loss at the same $L/2$ int8 budget, reaching $0.652$ on Qwen and $0.535$ on Pythia-6.9B.

\paragraph{The damage allocation achieves comparable or better downstream accuracy than HAWQ-V2.} On Qwen2.5-7B the damage allocation is the best quantized configuration on average, at $0.652$ against $0.645$ for HAWQ-V2 and $0.602$ for uniform int4, driven by ARC-Easy at $0.711$ against $0.679$; on Pythia-2.8B damage ties HAWQ-V2 at $0.534$; on Pythia-6.9B the three allocations sit within $0.2$ points, HAWQ-V2 at $0.537$, greedy at $0.536$, and damage at $0.535$.

\paragraph{The downstream ranking mirrors the perplexity ranking.} At every scale, the configuration that minimizes WikiText perplexity is within noise of the one that maximizes downstream accuracy, so the perplexity comparison is not measuring a capability-irrelevant artifact.

\subsection{ImageNet-scale post-training allocation: roughness-guided mixed precision} \label{sec:exp-inq}

The RSC training intervention of Sec.~\ref{sec:exp-vision} is evaluated at CIFAR scale. Here the separate post-training, forward-only mixed-precision procedure tests the allocation principle on pretrained ImageNet-scale models. On a pretrained RSB-A1 ResNet-50 \citep{wightman2021rsb}, a competitive public ImageNet training recipe, at ImageNetV2 (V2) top-1 $69.0\%$ under the recommended RSB-A1 evaluation transform, we score each of the 16 bottleneck blocks $l$ by the quantization-noise roughness $B^{\mathrm{quant}}_l$ of Sec.~\ref{sec:exp-quant}, the root-mean symmetrized KL between the full-precision output softmax and the output with block $l$'s convolutions, and nothing else, quantized to int4, over a fixed 1024-image V2 calibration batch, a functional score that needs no labels. We then allocate mixed int8 and int4 precision under a fixed budget of $K$ blocks int8 and $16-K$ int4, evaluating on V2 and an ImageNet-C (IN-C) subset of five corruption families $\times$ two severities.

\begin{table}[t!] \centering \caption{Post-training mixed-precision allocation on two pretrained ResNet-50 checkpoints. At each budget, $K$ of 16 bottleneck blocks use int8 and the rest use int4. Roughness is the label-free output-KL score; the oracle ranks blocks by measured $\Delta$-accuracy. Values are V2 and IN-C top-1 (\%).} \label{tab:inq} \small \begin{tabular}{lcccc} \toprule & \multicolumn{2}{c}{$K=4$ int8} & \multicolumn{2}{c}{$K=8$ int8} \\ \cmidrule(lr){2-3}\cmidrule(lr){4-5} Allocation & V2 & IN-C & V2 & IN-C \\ \midrule \multicolumn{5}{l}{a1\_in1k: fp16 69.0, uniform int8 68.5, int4 0.4} \\ roughness $B^{\mathrm{quant}}$ (ours) & 57.9 & 17.1 & 65.6 & 29.6 \\ HAWQ-V2 sensitivity & 56.7 & 15.6 & 65.6 & 29.6 \\ salience$\times$MSE (AWQ-like) & 42.4 & 12.1 & 50.2 & 17.1 \\ weight magnitude & 0.9 & 0.3 & 2.3 & 1.0 \\ oracle ($\Delta$acc) & 57.9 & 17.1 & 65.3 & 30.6 \\ \midrule \multicolumn{5}{l}{tv2\_in1k: fp16 67.9, uniform int8 68.1, int4 35.5} \\ roughness $B^{\mathrm{quant}}$ (ours) & 63.9 & 33.1 & 66.3 & 38.1 \\ HAWQ-V2 sensitivity & 62.8 & 33.3 & 66.4 & 40.7 \\ salience$\times$MSE (AWQ-like) & 58.4 & 25.8 & 64.2 & 35.6 \\ weight magnitude & 37.7 & 10.1 & 39.7 & 10.7 \\ oracle ($\Delta$acc) & 63.7 & 33.2 & 66.3 & 38.0 \\ \bottomrule \end{tabular} \end{table}

\paragraph{Roughness ranks damage; weight magnitude ranks damage backwards.} We observed that the per-block roughness predicts each block's true int4 damage, the V2 accuracy drop caused by quantizing that block, at Spearman $\rho_s = 0.88$, above the strengthened HAWQ-V2 Hessian-trace proxy \citep{dong2020hawqv2} at $0.85$, a seeded 8-probe Hutchinson trace that, unlike roughness, needs labels, and above the AWQ-like salience$\times$MSE score \citep{lin2024awq} at $0.37$, and opposite in sign to weight-magnitude relative MSE at $-0.24$. The mechanism is concrete: The first bottleneck block \texttt{layer1.0} is catastrophically quantization-sensitive, since quantizing that block alone to int4 costs $64$ points of V2 top-1, yet carries near-median weight magnitude, so a magnitude-guided allocation leaves the block at int4.

\paragraph{Roughness allocation matches the damage oracle at $K=4$.} Uniform int4 quantization is catastrophic here, dropping V2 from $69.0\%$ to $0.4\%$, so the allocation, meaning which blocks to spend the int8 budget on, is the whole problem. Table~\ref{tab:inq} gives the result. At $K=4$, protecting the four highest-roughness blocks recovers $57.9\%$ V2 top-1, exactly matching the ground-truth $\Delta$-accuracy oracle and above the strengthened HAWQ-V2 allocation's $56.7\%$ at an identical $4.35$ average bits. ResNet-50 blocks span roughly $200\times$ in size, so a $K$-block budget is not a fixed bit width; we record each configuration's realized average bit width, and the highlighted roughness and HAWQ configurations are exactly bit-matched here. Magnitude-, random-, and last-$K$ allocations, by contrast, collapse below $1\%$, protecting the wrong blocks. At $K=8$, the allocation problem is easier and HAWQ-V2 catches up: Roughness recovers $65.6\%$, within $0.3$ points of the oracle's $65.3\%$ and level with HAWQ-V2's $65.6\%$, with IN-C robustness likewise level at $29.6\%$ against HAWQ-V2's $29.6\%$ and magnitude's $1.0\%$. Roughness's advantage is concentrated at $K=4$, where the ranking must be most accurate; the roughness score's damage rank correlation of $0.88$ stays above the label- and Hessian-based HAWQ-V2's $0.85$ on this int4-fragile checkpoint. A forward-only, label-free functional roughness score allocates precision at least as well as parametric weight and Hessian scores on a public ImageNet checkpoint, with no training.

\paragraph{Replication across quantization regimes.} A torchvision-v2 ResNet-50 at V2 $67.9\%$, a recipe unrelated to RSB and far more int4-robust, with uniform int4 $35.5\%$ against RSB-A1's $0.4\%$, gives the same ranking result: The label-free roughness score ranks int4 damage on par with the strengthened, label- and Hessian-based HAWQ-V2, at $\rho_s = 0.68$ against $0.70$ and within measurement noise, both far above weight magnitude, which ranks that damage backwards at $-0.33$; roughness allocation still near-matches the damage oracle at $K=4$ and matches the oracle at $K=8$, giving $63.9\%$ and $66.3\%$ V2 against the oracle's $63.7\%$ and $66.3\%$ and HAWQ-V2's $62.8\%$ and $66.4\%$ in the lower part of Table~\ref{tab:inq}. The checkpoints differ substantially in quantization sensitivity: RSB-A1 collapses under uniform int4, while the torchvision-v2 model retains substantially more accuracy. Together, the two checkpoints establish the forward-only, label-free roughness score as competitive with a strong label-based HAWQ-V2: ahead on the int4-fragile checkpoint, where the allocation problem is hardest, and level with HAWQ-V2 on the int4-robust one, while weight magnitude is anti-correlated with damage in both.

\section{Resolving the allocation by scale on the residual stream} \label{sec:exp-spectrum}

The residual-stream damage curve and the empirical ranking results for that curve are reported in Sec.~\ref{sec:exp-quant} and Table~\ref{tab:spectrum-corr}. This section records the complementary local expansion and detailed interpretation of that scale-resolved result.

Because the finest dyadic scales fall below the bf16 residual-stream unit of least precision, we measure $D_i(r)$ in fp32 on $24$ to $32$ calibration windows for Pythia-160M to 6.9B and Qwen2.5 at $1.5$B and $7$B, and, with the method frozen, for three held-out modern deployment LLMs, namely Phi-3.5-mini, Granite-3.1-8B, and Hermes-3-Llama-3.1-8B \citep{teknium2024hermes3}, with the last denoted Hermes-3 elsewhere in the paper, reusing the ground-truth per-block int4 damage of Sec.~\ref{sec:exp-quant} as the target.

This damage curve is a signed order-$1$ finite-difference curve for the loss functional $g=\mathrm{NLL}$ along the quantization direction. Put $G_i(r)\coloneqq\E_t[\mathrm{NLL}_t(\mathrm{res}_i+r e_i)]$, so that $D_i(r)=G_i(r)-G_i(0)$. Whenever $G_i$ is twice differentiable at zero, \begin{align*} D_i(r)=rG_i'(0)+\tfrac12 r^2G_i''(0)+o(r^2). \end{align*} Consequently $D_i(r)/r\to G_i'(0)$, the directional gradient term; the local Hessian controls the quadratic coefficient $G_i''(0)$, which can be isolated by subtracting the linear term or by taking a symmetric second difference. The parameter-space loss-Hessian trace and HAWQ-V2 sensitivity score used below are separate local-curvature proxies, not literal $r\to0$ limits of this residual-space first-difference curve.

\paragraph{The measured radius range is approximately quadratic-dominated.} The median base-2 scaling exponent of $D_i(r)$ on $[1/2,1]$ is $\nu\approx 2.0$ across the seven models. By the Taylor expansion above, $\nu\approx2$ is compatible with a small directional linear term and a dominant quadratic curvature term over this range; the scaling exponent alone is not evidence of nonsmoothness or a failure of local Hessian analysis.

\paragraph{The quantization scale is the most informative probe.} We observed that the rank correlation between $D_i(r)$ and the true int4 damage (Table~\ref{tab:spectrum-corr}) is highest at $r=1$ and declines as the probe shrinks over the measured grid: Averaged over the seven models, the correlation falls from $0.92$ at the quantization scale to $0.53$ at $r=2^{-4}$. This decline is the finite-scale observation: The predictive content of the damage curve concentrates at the finite scale the quantizer actually occupies, with weaker prediction at the smallest measured radius.

\paragraph{The infinitesimal Hessian scalars sit below the finite-scale reading.} The advantage at the quantization scale extends beyond comparisons among radii of the same damage curve: The separate parameter-space baselines, the loss-Hessian trace and HAWQ-V2, also fall below the quantization-scale reading, at mean $0.59$ and $0.70$ across the seven models against $0.92$ at $r=1$, though the gap is model-dependent and narrows on the larger models, where Qwen2.5-1.5B's HAWQ-V2 reaches $0.84$. Together with the fixed-probe radius sweep, this comparison supports scale mismatch as one contributor to the finite-scale damage allocation's advantage over HAWQ-V2 at equal bit budgets: The scale decomposition locates the mechanism rather than proposing a new allocator, since $D_i(1)$ already coincides with the damage score.

\paragraph{Damage remains predictive where drift reverses.} The finite-scale damage curve does not reverse at the larger Pythia model sizes: The damage curve's rank correlation with the true int4 damage stays above $0.81$ on every model, including the 2.8B and 6.9B scales where the a priori drift score reverses to $-0.09$ and $-0.45$ (Sec.~\ref{sec:exp-quant}). Reading the task-aligned damage curve at the finite quantization scale is stable exactly where the finer-radius damage measurements and a priori drift summaries are not.

\paragraph{The finite-scale-to-Hessian-proxy relation sharpens on held-out modern LLMs.} Applying the identical fp32 instrument to three held-out modern deployment LLMs never used to design the method, namely Phi-3.5-mini, Granite-3.1-8B, and Hermes-3 in the lower part of Table~\ref{tab:spectrum-corr}, reproduces the pattern and widens the gap further. The finite-scale $r=1$ reading averages $0.87$ across the three, as predictive of int4 damage as on the Pythia family, while the two infinitesimal Hessian scalars fall to a mean of $0.15$ for the loss-Hessian trace and $0.22$ for HAWQ-V2. The finite-scale-to-Hessian-proxy gap is $0.65$ here, against $\approx 0.22$ across the seven models in the upper part of that table. The residual-stream damage spectrum locates int4 damage on models whose Hessian proxies carry almost no rank information, the regime in which the finite-scale allocation of Sec.~\ref{sec:exp-modllm} has the largest observed advantage over HAWQ-V2.

\begin{figure}[t!] \centering \includegraphics[width=\linewidth]{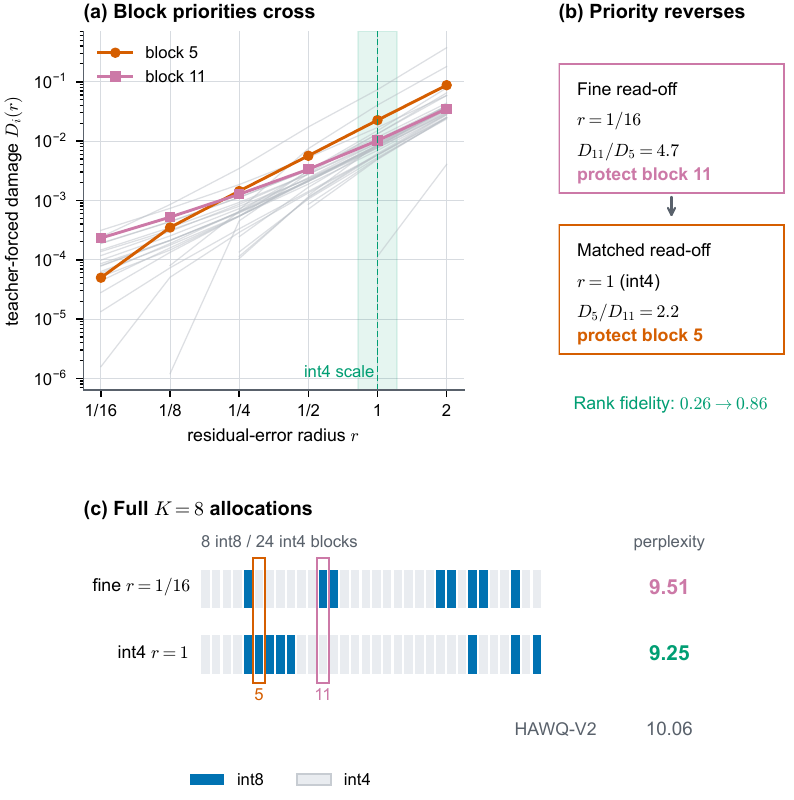} \caption{A held-out Phi-3.5-mini-instruct example of scale-matched allocation. (a) Blocks 5 and 11 reverse order between $r=1/16$ and the int4 error radius $r=1$. (b) Rank correlation with independently measured int4 damage rises from $0.26$ to $0.86$ at the matched radius. (c) With $K=8$ int8 blocks, the fine- and matched-radius allocations yield perplexities $9.51$ and $9.25$, respectively, against $10.06$ for HAWQ-V2.} \label{fig:qual-scale-crossing} \end{figure}

\begin{figure}[t!] \centering \includegraphics[width=\linewidth]{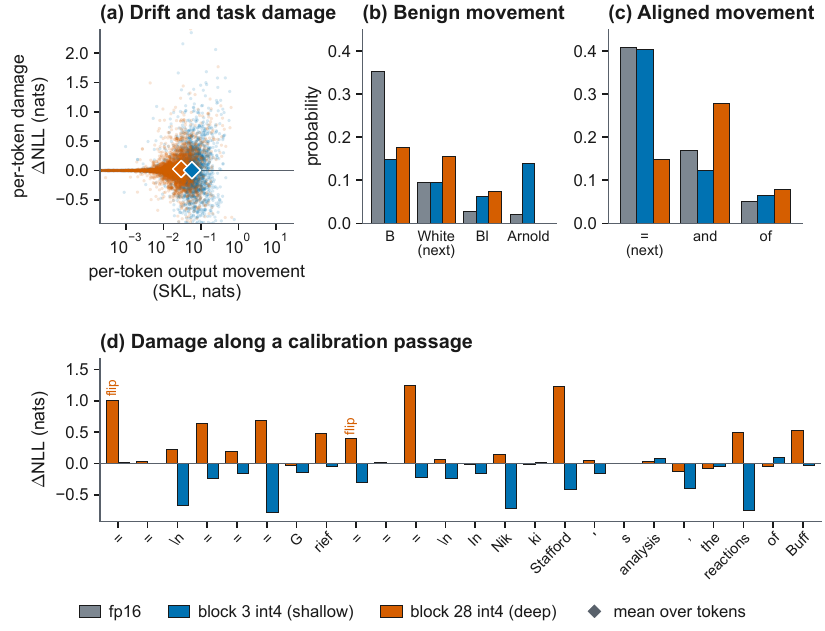} \caption{Token-level drift--damage dissociation for Pythia-6.9B under single-block int4 RTN. (a) A shallow block moves the output distribution farther but causes less next-token NLL damage than a deep block, where SKL denotes per-token symmetric KL, $\dskl^2$, in nats. Diamonds mark the means over tokens, SKL $0.057$ and $\Delta$NLL $0.0041$ nats for block 3 and SKL $0.030$ and $\Delta$NLL $0.0258$ nats for block 28. (b) and (c) Example next-token distributions show large task-benign movement and smaller task-aligned movement, respectively. (d) Damage concentrates on particular tokens for the deep block.} \label{fig:qual-token-dissociation} \end{figure}

The endpoint-ordering change in Figure~\ref{fig:vision-spectra} matters operationally only if the ordering change alters a deployment decision. Figure~\ref{fig:qual-scale-crossing} makes that link explicit on a held-out instruction-tuned model: Two blocks exchange priority before the radius reaches the actual int4 perturbation scale, and the resulting equal-budget allocations protect different blocks. This block-ranking reversal is the event summarized across models in Table~\ref{tab:spectrum-corr} and the concrete reason that the read-off in \eqref{eq:readoff} must be matched to the failure scale.

 \FloatBarrier

 \clearpage

\section{Extended discussion} \label{app:discussion-extra}

\paragraph{Limitations.} \label{app:limitations} The experimental scope is broad but not exhaustive: The pools, models, and formats we run are the ones listed in App.~\ref{app:setup-details}, namely CIFAR-scale training pools and ImageNet-1k in vision, the Pythia, Qwen2.5, and Mamba families in language, and uniform, GPTQ, AWQ, and SpQR-style quantization, rather than every dataset and every model. Results outside that scope, in particular at larger-scale pretraining and under formats and architectures we did not run, remain to be measured.

\paragraph{Outlook.} The measurement layer is generic: Scalar and vector-valued responses support literal finite-difference spectra, while output divergences support analogous diagnostic curves without automatically defining classical (quasi-)seminorms. Natural extensions include task-carrying directions that directly measure the sensitivity tracked indirectly by clean accuracy, state-noise spectra for Mamba and other state-space models, LoRA-subspace roughness for fine-tuning stability, and roughness-matched distillation. The construction also applies to loss roughness along an optimizer's update direction at the optimizer's actual step size: In Definition~\ref{def:spectrum}, set $g$ to the loss, $z$ to the parameters, and use the direction family supplied by the optimizer.

\paragraph{The allocation comparisons at matched format and budget.} The most direct evidence compares allocation rules at a shared format and bit budget. Apps.~\ref{sec:exp-awq} and~\ref{sec:exp-spqr} rebuild that comparison on the outlier-aware AWQ scaling format \citep{lin2024awq} and a controlled SpQR-style dense-and-sparse variant \citep{dettmers2024spqr}; at matched $K$, the damage rule is compared directly with HAWQ-V2. Interaction-aware greedy re-measurement removes the additive modeling step, and the reported greedy performance is empirical.

\paragraph{Datasets and checkpoints used.} Every dataset used is public: CIFAR-10 and CIFAR-100 \citep{krizhevsky2009learning} with the corresponding corruption sets \citep{hendrycks2019benchmarking}, ImageNet-1k \citep{deng2009imagenet} with IN-C \citep{hendrycks2019benchmarking} and V2 \citep{recht2019imagenet}, WikiText-103 \citep{merity2017pointer}, LAMBADA \citep{paperno2016lambada}, and the Pile \citep{gao2020pile} through the public Pythia releases. Every pretrained model is an off-the-shelf public checkpoint obtained from the Hugging Face Hub or the \texttt{timm} registry \citep{rw2019timm}; no proprietary data or models are used.

\paragraph{Motivation for spectra and source-specific interventions.} Our observations across the testbeds reported here support the finite-scale design principle: Scalar proxies compress global, average, or infinitesimal behavior, while the failures we measured are finite-scale events, namely corruptions at severity scales and quantization noise at the magnitudes realized by a specified quantizer and bit width. Margin finite differences and source-specific quantization scores make the diagnosis actionable: RSC regularizes margin roughness along selected nuisance directions and scales, while the allocation rule spends precision where finite-scale task damage is high. In the convolutional CIFAR pools, RSC reduces the reported margin roughness and improves noise robustness; the ViT pool establishes intervention transfer without a reduction in the reported diagnostic. The auxiliary activation-crossing term in App.~\ref{app:crossing-penalty} provides no additional robustness benefit in the evaluated auxiliary-penalty settings. Clean costs and the limits of the source interpretation are reported alongside these outcomes. This evidence supports scale and source alignment. The classical Besov connection follows through the isotropic and mixed-direction population equivalences, the directionally weighted RSC energy, and the uniform control of data-dependent moments (Theorem~\ref{thm:besov-equiv}, Corollary~\ref{cor:mixed-besov}, Propositions~\ref{prop:directional-energy} and~\ref{prop:uniform-besov}).

\paragraph{Relation to scalar diagnostics.} Finite differences recover Lipschitz sensitivity through worst-case first-order quotients and directional Hessian curvature through second-order limits for smooth functions (Remark~\ref{rem:scalar-endpoints}, App.~\ref{app:scalar-endpoints}). These identities fix the same function and representation. The experiments also compare finite-scale measurements with scalar baselines in other representations, including parameter-space loss Hessians. These experimental comparisons support using the measurements for intervention and allocation, but do not identify every baseline with a limit of the deployed curve. Hessian and Lipschitz analyses remain useful for infinitesimal analysis and worst-case certification. Finite-scale probes use forward passes only. On one GH200, the full CIFAR-10 ResNet-18 diagnostic and robustness suite takes $45$ seconds; per-block LLM measurement and allocation evaluation take about $14$ minutes on Pythia-2.8B and under one hour on Pythia-6.9B (App.~\ref{app:measurement}).

\end{document}